\documentclass{article}

\PassOptionsToPackage{numbers, sort, compress}{natbib}

\usepackage[final]{arxiv}

\author{
Dweep Trivedi\thanks{Contributed equally.} \ \thanks{Work partially done as a visiting scholar at USC.} \hskip2em Jesse Zhang\footnotemark[1] \ \textsuperscript{1} \hskip2em Shao-Hua Sun\textsuperscript{1} \hskip2em Joseph J. Lim\thanks{AI Advisor at NAVER AI Lab.} \ \textsuperscript{1} \\
\textsuperscript{1}University of Southern California \\
{\small \texttt{\{dtrivedi, jessez, shaohuas, limjj\}@usc.edu}}
}

\usepackage[utf8]{inputenc} %
\usepackage[T1]{fontenc}    %
\usepackage[colorlinks=True, citecolor=black, linkcolor=red]{hyperref} 
\usepackage{url}            %
\usepackage{booktabs}       %
\usepackage{amsfonts}       %
\usepackage{nicefrac}       %
\usepackage{microtype}      %

\usepackage{graphicx}
\usepackage{caption}
\usepackage{subcaption}
\usepackage{amsmath}
\usepackage{amssymb}
\usepackage{caption}
\usepackage{courier}
\usepackage{dsfont}
\usepackage{mathtools}
\usepackage{dsfont}
\usepackage{multirow}
\usepackage{tabularx}
\usepackage{arydshln}
\usepackage{pifont}
\usepackage{algorithm}
\usepackage{setspace}
\usepackage{wrapfig}
\usepackage{lipsum}
\usepackage{mdframed}
\usepackage{framed}
\usepackage{tcolorbox}
\usepackage[formats]{listings} %
\usepackage{lscape} %
\mdfsetup{innertopmargin=1pt, frametitlealignment=\center}
\usepackage[nottoc]{tocbibind}
\usepackage{minitoc}
\usepackage{bbm}

\usepackage{enumitem}
\setlist[itemize]{leftmargin=5mm}

\newcommand{\mytitle}{\title{
Learning to Synthesize Programs as Interpretable and Generalizable Policies
}}

\newcommand{\sunnote}[1]{}

\newcommand{\Skip}[1]{}

\newcommand{\vspacesection}[1]{\vspace{-0.0cm}
\section{#1}
\vspace{-0.0cm}}
\newcommand{\vspacesubsection}[1]{\vspace{-0.0cm}
\subsection{#1}
\vspace{-0.0cm}}
\newcommand{\vspacesubsubsection}[1]{\vspace{-0.0cm}
\subsubsection{#1}
\vspace{-0.0cm}}

\newcommand{\method}[1]{LEAPS}

\newcommand{\ie}{\textit{i}.\textit{e}.\ }
\newcommand{\eg}{\textit{e}.\textit{g}.\ }

\newcommand{\myfig}[1]{Figure~\ref{#1}}
\newcommand{\mytable}[1]{Table~\ref{#1}}
\newcommand{\myeq}[1]{Eq.~\ref{#1}}
\newcommand{\mysecref}[1]{Section~\ref{#1}}

\newcommand{\dotieconcat}[2]{%
  \text{\raisebox{.8ex}{$\smallfrown$}}%
}

\makeatletter
\newcommand\dslfontsize{\@setfontsize\dslfontsize\@viipt\@viiipt}
\makeatother

\newcommand\SmallCaption[1]{%
  \captionsetup{font=scriptsize}%
  \caption{#1}}

\usepackage{color}
\usepackage{xcolor}
\definecolor{codegray}{rgb}{0.5,0.5,0.5}
\lstdefinelanguage{DSL}{
  sensitive = true,
  keywords={DEF, WHILE, IF},
  otherkeywords={%
    >, <, ==
  },
  keywords = [2]{noMarkersPresent, leftIsClear, rightIsClear, frontIsClear, markersPresent},
  keywords = [3]{move, turnLeft, turnRight, pickMarker, putMarker},
  numbersep=8pt,
  tabsize=1,
  showstringspaces=false,
  breaklines=true,
  frame=top,
  comment=[l]{//},
  morecomment=[s]{/*}{*/},
  commentstyle=\color{purple}\ttfamily,
  stringstyle=\color{red}\ttfamily,
  belowskip=0.5pt,
  aboveskip=0.2pt,
  morestring=[b]',
  morestring=[b]",
  keywordstyle=\color{blue},
  keywordstyle=[2]\color{red},%
  keywordstyle=[3]\color{codegray},%
  numberstyle=\tiny\color{codegray},
  basicstyle=\ttfamily\tiny,
  }
\lstloadaspects{formats}
\begin{document}
\doparttoc %
\faketableofcontents %

\mytitle
\maketitle

\begin{abstract}
Recently, deep reinforcement learning (DRL) methods have achieved impressive performance on tasks in a variety of domains. However, neural network policies produced with DRL methods are not human-interpretable and often have difficulty generalizing to novel scenarios. To address these issues, prior works explore learning programmatic policies that are more interpretable and structured for generalization. Yet, these works either employ limited policy representations (e.g. decision trees, state machines, or predefined program templates) or require stronger supervision (e.g. input/output state pairs or expert demonstrations). We present a framework that instead learns to synthesize a program, which details the procedure to solve a task in a flexible and expressive manner, solely from reward signals. To alleviate the difficulty of learning to compose programs to induce the desired agent behavior from scratch, we propose to first learn a program embedding space that continuously parameterizes diverse behaviors in an unsupervised manner and then search over the learned program embedding space to yield a program that maximizes the return for a given task. Experimental results demonstrate that the proposed framework not only learns to reliably synthesize task-solving programs but also outperforms DRL and program synthesis baselines while producing interpretable and more generalizable policies. We also justify the necessity of the proposed two-stage learning scheme as well as analyze various methods for learning the program embedding. Website at \url{https://clvrai.com/leaps}.

\end{abstract}

\vspacesection{Introduction}
\label{sec:intro}

\sunnote{DRL: (1) not interpretable (2) cant generalize}

Recently, deep reinforcement learning (DRL) methods have demonstrated encouraging performance on a variety of domains such as outperforming humans in complex games~\cite{mnih2015humanlevel, silver2016mastering, silver2018general, vinyals2019grandmaster} or controlling robots~\cite{campeau2019kinova, gu2017deep, openai2018learning, hafner2019dream, yamada2020motion, zhang2021policy, lee2019follow}. Despite the recent progress in the field, acquiring complex skills through trial and error
still remains challenging and these neural network policies often have difficulty generalizing to novel scenarios. Moreover, such policies are not interpretable to humans and therefore are difficult to debug
when these challenges arise.

\sunnote{PRL prior works: (1) limited representations (2) strong supervision}

To address these issues, a growing body of work aims to learn programmatic policies that are structured in more interpretable and generalizable representations such as decision trees~\cite{bastani2018verifiable}, state-machines~\cite{Inala2020Synthesizing}, 
and programs described by domain-specific programming languages~\cite{verma2018programmatically, verma2019imitation}.
Yet, the programmatic representations employed in these works are often limited in expressiveness due to constraints on the policy spaces. 
For example, decision tree policies are incapable of na\"{i}vely generating repetitive behaviors, state machine policies used in~\cite{Inala2020Synthesizing} are computationally complex to scale to policies representing 
diverse behaviors, 
and the programs of~\citep{verma2018programmatically, verma2019imitation} are constrained to a set of predefined program templates.
On the other hand, 
program synthesis works 
that aim to represent desired behaviors using flexible domain-specific programs often require extra supervision such as input/output pairs~\citep{devlin2017robustfill, bunel2018leveraging, chen2018executionguided, shin2018improving, lazaro2018beyond} or expert demonstrations~\citep{sun2018neural, plans},
which can be difficult to obtain.

\sunnote{our problem: programs from reward}
In this paper, we present a framework to instead synthesize human-readable programs in an expressive representation, solely from rewards, to solve tasks described by Markov Decision Processes (MDPs).
Specifically, we represent a policy using a program 
composed of control flows (\eg if/else and loops) and an agent's perceptions and actions.
Our programs can flexibly compose behaviors through perception-conditioned loops 
and nested conditional statements.
However, composing individual program tokens (\eg \texttt{if}, \texttt{while}, \texttt{move()}) in a trial-and-error fashion to synthesize programs 
that can solve given MDPs can be extremely difficult and inefficient.

\sunnote{our method: the intuition of the proposed two stage scheme}

To address this problem, 
we propose to first learn a 
latent program embedding space 
where 
nearby latent programs correspond to
similar behaviors and allows for smooth interpolation,
together with a program decoder that can decode 
a latent program to a program consisting of a sequence of program tokens.
Then, when a task is given, this embedding space allows us to iteratively search over candidate latent programs to find a program that induces desired behavior to maximize the reward.
Specifically,
this embedding space is learned through reconstruction of randomly generated programs and the behaviors they induce in the environment in an unsupervised manner.
Once learned, the embedding space can be reused to solve different tasks without retraining.

\sunnote{experiments and results}

To evaluate the proposed framework, we consider the Karel domain~\citep{pattis1981karel},
featuring an agent navigating through a gridworld and interacting with objects 
to solve 
tasks such as stacking and navigation. 
The experimental results demonstrate that the proposed framework not only learns to reliably synthesize task-solving programs but also outperforms program synthesis and deep RL baselines.
In addition, we justify the necessity of the proposed two-stage learning scheme as well as conduct an extensive analysis comparing various approaches for learning the latent program embedding spaces. 
Finally, we perform experiments which highlight that the programs
produced by our proposed framework can both generalize to larger state 
spaces and unseen state configurations 
as well as be interpreted and edited by humans to improve
their task performance.

\vspacesection{Related Work} \label{sec:related_work}
\Skip{
\textbf{Deep reinforcement learning.} 
Recently, the success of DRL has led to the development of
methods that address a variety of tasks such as playing games or 
controlling robots~\citep{mnih2015humanlevel, silver2016mastering, silver2018general, vinyals2019grandmaster, 
campeau2019kinova, kalashnikov18a, gu2017deep, openai2018learning, kalashnikov2018qtopt, hafner2019dream}.
However, neural networks learned by DRL methods  
are not interpretable, making it difficult to diagnose and address task failures. 
Furthermore, DRL methods can struggle to acquire long horizon skills or generalize to new settings. 
Our work aims to address these issues by instead learning to synthesize programmatic policies.
}

\textbf{Neural program induction and synthesis.} 
Program induction methods~\cite{lazaro2018beyond, xu2018neural,
devlin2017neural, neelakantan2015neural, graves2014neural, kaiser2015neural,
gaunt2017differentiable,reed2016neural, cai2017making, xiao2018improving, Penkov2017, yan2020neural, li2020strong, huang2019neural} 
aim to implicitly induce
the underlying programs to mimic the behaviors demonstrated in given task
specifications such as input/output pairs or expert demonstrations. 
On the other hand, program synthesis
methods~\cite{ 
devlin2017robustfill, bunel2018leveraging, chen2018executionguided,  shin2018improving, sun2018neural,
bosnjak17a, parisotto2016neuro, NIPS2018_8107, liu2018learning,
lin2018nl2bash, liao2019synthesizing, ellis2019write, ellis2020dreamcoder, balog2016deepcoder, gecco_program_synthesis_paper, abolafia2018neural, hong2020latent, silver2020few, nsd, openai_codex, alet2021large, chen2021latent, austin2021program, hong2021latent, wong2021leveraging, chen2021spreadsheetcoder, nye2021representing}
explicitly synthesize the underlying programs and execute the programs to
perform the tasks from task specifications such input/output pairs, demonstrations, language instructions.
In contrast, we aim to learn to synthesize programs solely from reward described by an MDP without other task specifications.
Similarly to us, a two-stage synthesis method is proposed in~\citep{gecco_program_synthesis_paper}. 
Yet, the task is to 
match truth tables for given test programs rather than solve MDPs.
Their first stage requires the entire ground-truth table for each program synthesized during training, 
which is infeasible to apply to our problem setup (\ie synthesizing imperative programs for solving MDPs).

\textbf{Learning programmatic policies.} 
Prior works have also addressed the
problem of learning programmatic policies~\citep{choi2005learning, distill, landajuela21a}.
\citet{bastani2018verifiable} learns a decision tree as a programmatic policy for pong and cartpole
environments by imitating an oracle neural policy. 
However, decision trees are incapable of representing repeating behaviors
on their own. \citet{silver2020few} addresses this by including a loop-style token for their decision tree policy, though it is still not as expressive as synthesized loops.
\citet{Inala2020Synthesizing}
learns programmatic policies as finite state machines by imitating a teacher policy, although
finite state machine complexity can scale quadratically with the number of states, making them difficult to scale to more complex behaviors.

Another line of work instead synthesizes programs structured in Domain Specific Languages (DSLs), 
allowing humans to design tokens (\eg conditions and operations) and control flows (\eg while loops, if statements, reusable functions) 
to induce desired behaviors
and can produce human interpretable programs.
\citet{verma2018programmatically, verma2019imitation} distill neural network policies into programmatic policies.
Yet, the initial programs are constrained
to a set of predefined program templates. 
This significantly limits the scope of synthesizable programs 
and requires designing such templates for each task. 
In contrast, our method can synthesize diverse programs, without templates, 
which can flexibly represent the complex behaviors required to solve various tasks.

\begin{wrapfigure}[11]{R}{0.45\textwidth}
\centering
\vspace{-0.5cm}
\begin{mdframed}
\vspace{-0.2cm}
{%
 \dslfontsize
    \begin{align*}
    \text{Program}\ \rho &\coloneqq \text{DEF}\  \text{run}\ \text{m}(\ s\ \text{m})\\
    \text{Repetition} \ n &\coloneqq \text{Number of repetitions}\\
    \text{Perception} \ h & \coloneqq \text{Domain-dependent perceptions}\\
    \text{Condition} \ b &\coloneqq \text{perception h} \ | \ \text{not} \ \text{perception h} \\
    \text{Action} \ a &\coloneqq \text{Domain-dependent actions}\\
    \text{Statement}\ s &\coloneqq \text{while}\ \text{c}(\ b\ \text{c})\ \text{w}(\ s\ \text{w}) \ | \ s_1 ; s_2 \ | \ a \ | \\ 
    & \ \text{repeat}\ \text{R=}n\ \text{r}(\ s\ \text{r}) \ | \ \text{if}\ \text{c}(\ b\ \text{c})\ \text{i}(\ s\ \text{i}) \ | \\ 
    & \ \text{ifelse}\ \text{c}(\ b\ \text{c})\ \text{i}(\ s_1\ \text{i}) \  \text{else}\ \text{e}(\ s_2\ \text{e}) \\
    \end{align*}
}
\vspace{-0.93cm}
\end{mdframed}
    \vspace{-0.25cm}
    \caption[]{
        \small
        The domain-specific language (DSL) for constructing programs.  
        \label{fig:dsl}
    }
\end{wrapfigure}

\vspacesection{Problem Formulation}
\label{sec:problem}

We are interested in learning to synthesize a program structured
in a given DSL that can be executed to solve a 
given task described by an MDP, purely from reward. 
In this section, we formally define our definition of a program and DSL, tasks described by MDPs, 
and the problem formulation.

\noindent \textbf{Program and Domain Specific Language.} 
The programs, or programmatic policies, considered in this work are defined
based on a DSL as shown in \myfig{fig:dsl}. 
The DSL consists of control flows and an agent's perceptions and actions. 
A perception indicates circumstances in the environment (\eg
\texttt{frontIsClear()}) that can be perceived by an agent, while an
action defines a certain behavior that can be performed by an agent (\eg
\texttt{move()}, \texttt{turnLeft()}). 
Control flow includes
\texttt{if/else} statements, loops, 
and boolean/logical operators to compose more sophisticated conditions.
A policy considered in this work is described by a program $\rho$ 
which is executed to produce a sequence of actions given perceptions from the environment.

\noindent \textbf{MDP.} 
We consider finite-horizon discounted MDPs with initial
state distribution $\mu(s_o)$ and discount factor $\gamma$.
For a fixed sequence $\{(s_0 , a_0), ... , (s_t , a_t)\}$ of states and
actions obtained from a rollout of a given policy,
the performance of the policy is evaluated based on a discounted return
$\sum_{t=0}^{T} \gamma^t r_t$, where $T$ is the horizon of the episode and
$r_t = \mathcal{R}(s_t, a_t)$ the reward function.

\noindent \textbf{Objective.} 
Our objective is 
\begin{math}
    \max_{\rho} \mathbb{E}_{a \sim \text{EXEC}(\rho), s_0 \sim \mu}[\sum_{t=0}^T \gamma^t
    r_t]
\end{math},
where EXEC returns the actions induced by executing a program policy $\rho$ in the environment. 
Note that one can view this objective as a special case of the standard RL objective,
where the policy is represented as a program which follows the grammar of the DSL and
the policy rollout is obtained by executing the program.

\vspacesection{Approach}
\label{sec:appraoch}

Our goal is to develop a framework that can synthesize 
a program (\ie a programmatic policy)
structured in a given DSL that can be executed to solve a task of interest. 
This requires the ability to synthesize a program that is 
not only valid for execution (\eg grammatically correct)
but also describes desired behaviors for solving the task 
from only the reward.
Yet, learning to synthesize such a program from scratch for every new task can be difficult and inefficient.

\begin{figure}[h!]
\centering
    \includegraphics[width=.99\textwidth]{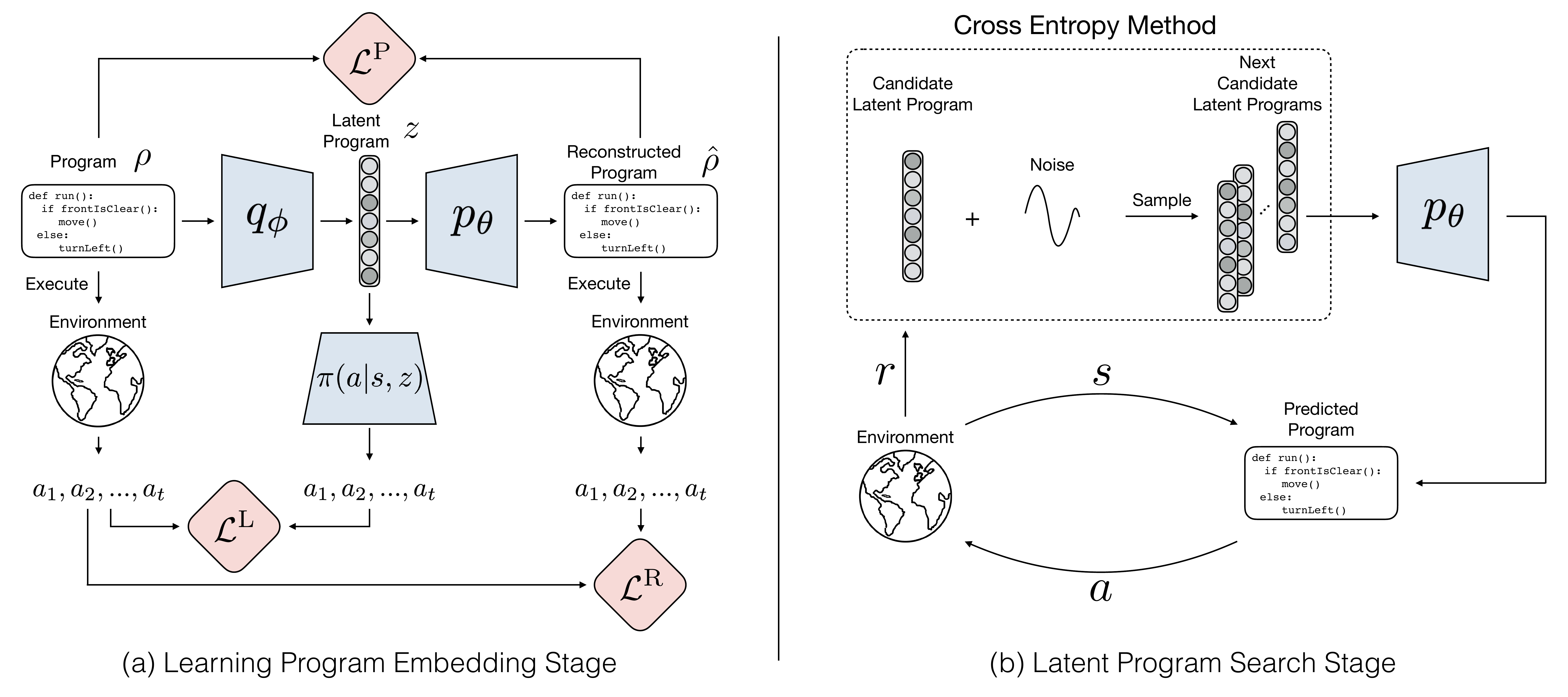}

    \caption[]{
        \small 
        (a) \textbf{Learning program embedding stage}:
        we propose to learn a program embedding space
        by training
        a program encoder $q_\phi$ that encodes a program as a latent program $z$, 
        a program decoder $p_\theta$ that decodes the latent program $z$ back to a reconstructed program $\hat{\rho}$, 
        and a policy $\pi$ that conditions on the latent program $z$ and acts as a neural program executor to produce the execution trace of the latent program $z$.
        The model optimizes a combination of 
        a program reconstruction loss $\mathcal{L}^{\text{P}}$, 
        a program behavior reconstruction loss $\mathcal{L}^{\text{R}}$,
        and a latent behavior reconstruction loss $\mathcal{L}^{\text{L}}$.
        $a_1, a_2, .., a_t$ denotes actions produced by either the policy $\pi$ or program execution.
        (b) \textbf{Latent program search stage}: 
        we use the Cross Entropy Method
        to iteratively search for the best candidate latent programs 
        that can be decoded and executed to maximize the reward to solve given tasks.
        \label{fig:model}
    }
\end{figure}

To this end, we propose our
\textbf{L}earning \textbf{E}mbeddings for l\textbf{A}tent \textbf{P}rogram \textbf{S}ynthesis framework, 
dubbed \method\\, as illustrated in \myfig{fig:model}.
\method\\ first learns a latent program embedding space 
that continuously parameterizes diverse behaviors 
and a program decoder that decodes a latent program to a program consisting of a sequence of program tokens.
Then, when a task is given, we iteratively search over this embedding space and decode each candidate latent program using the decoder to find a program that maximizes the reward.
This two-stage learning scheme 
not only enables learning to synthesize programs to acquire desired behaviors described by MDPs solely from reward,
but also allows reusing the learned embedding space to solve different tasks without retraining.

In the rest of this section, 
we describe how we construct the model and our learning objectives for the latent program embedding space
in~\mysecref{subsec:stage1}.
Then, 
we present how a program 
that describes desired behaviors for a given task
can be found through a search algorithm
in~\mysecref{subsec:stage2}.

\vspacesubsection{Learning a Program Embedding Space}
\label{subsec:stage1}
To learn a 
latent program embedding space, 
we propose to train a variational autoencoder (VAE)~\cite{vae} that consists of 
a program encoder $q_\phi$ which encodes a program $\rho$ to a latent program $z$
and a program decoder $p_\theta$ which reconstructs the program from the latent. 
Specifically, the VAE is trained through 
reconstruction of randomly generated programs 
and the behaviors they induce in the environment in an unsupervised manner. Architectural details are listed in \mysecref{sec:training}.

\sunnote{goal: smooth interpolation allows search}

Since we aim to iteratively search over the learned embedding space to achieve certain behaviors when a task is given,
we want this embedding space to allow for smooth behavior interpolation (\ie programs that exhibit similar behaviors are encoded closer in the embedding space).
To this end, we propose to train the model by optimizing the following three objectives.

\vspacesubsubsection{Program Reconstruction}
\label{subsec:ploss}

To learn a program embedding space,
we train a program encoder $q_\phi$ and a program decoder $p_\theta$ to 
reconstruct programs composed of sequences of program tokens.
Given an input program $\rho$ consisting of a sequence of program tokens,
the encoder processes the input program one token
at a time and produces a latent program embedding $z$.
Then, the decoder outputs program tokens one by one from the latent program embedding $z$ to synthesize a reconstructed program $\hat{\rho}$.
Both the encoder and the decoder are recurrent neural networks and are trained to 
optimize the $\beta$-VAE \citep{higgins2016beta} loss:

\begin{equation}
\label{eq:ploss} 
\mathcal{L}_{\theta,\phi}^{\text{P}}(\rho) = -
\mathbb{E}_{\mathbf{z} \sim q_\phi(\mathbf{z}\vert\mathbf{\rho})} [\log
p_\theta(\boldsymbol{\rho}\vert\mathbf{z})] + \beta
D_\text{KL}(q_\phi(\mathbf{z}\vert\boldsymbol{\rho})\|p_\theta(\mathbf{z}))
.
\end{equation}

\vspacesubsubsection{Program Behavior Reconstruction} 
\label{subsec:rloss}

While the loss in~\myeq{eq:ploss} enforces that the model 
encodes syntactically similar programs close to each other in the embedding space,
we also want to encourage programs with the same semantics to have similar program embeddings.
An example that demonstrates the importance of this is the \textit{program aliasing} issue, where
different programs have identical program semantics 
(\eg 
\texttt{repeat(2): move()} and \texttt{move() move()}).
Thus, we introduce an objective that compares 
the execution traces of the input program and the reconstructed program.
Since the program execution process is not differentiable, 
we optimize the model via 
REINFORCE~\citep{williams1992simple}:  

\begin{equation}
\label{eq:rloss}
\mathcal{L}_{\theta, \phi}^\text{R}(\rho) 
= - \mathbb{E}_{{z} \sim
q_\phi(z\vert\rho)}[R_\text{mat}(p_\theta(\rho|z),
\rho)],
\end{equation}
where $R_{\text{mat}}(\hat{\rho}, \rho)$, the reward for matching the input program's behavior, is defined as
\begin{equation}
\label{eq:RL Reward}
R_{\text{mat}}(\hat{\rho}, \rho) =
\mathbb{E}_{\mu}[
\frac{1}{N}\sum_{t=1}^{N}\underbrace{\mathbbm{1}\{\text{EXEC}_i(\mathbf{\hat{\rho}}) == \text{EXEC}_i(\mathbf{\rho}) \; \forall i =
1, 2, ... t\}}_{\text{stays 0 after the first } t \text{ where } \text{EXEC}_t(\hat{\rho})\; != \;\text{EXEC}_t(\rho)}],
\end{equation}
where $N$ is the maximum of the lengths of the execution traces of both programs,
and $\text{EXEC}_i(\rho)$ represents the action taken by program $\rho$ at time $i$.
Thus this objective encourages the model to embed behaviorally similar yet possibly syntactically 
different programs to similar latent programs.

\vspacesubsubsection{Latent Behavior Reconstruction} 
\label{subsec:lloss}

To further encourage learning a program embedding space 
that allows for smooth behavior interpolation, 
we devise another source of supervision by 
learning a program embedding-conditioned policy.
Denoted $\pi(a|z, s_t)$, 
this recurrent policy takes the program embedding $z$ produced by the program encoder
and learns to predict corresponding agent actions.
One can view this policy as a neural program executor 
that allows gradient propagation through the policy and the program encoder by optimizing 
the cross entropy between 
the actions obtained by executing the input program $\rho$
and the actions predicted by the policy:

\begin{equation}
    \label{eq:lloss}
    \mathcal{L}_\pi^{\text{L}}(\rho, \pi) = - \mathbb{E}_{\mu}[\sum_{t=1}^{M} \sum_{i=1}^{|\mathcal{A}|} \mathbbm{1}\{\text{EXEC}_i(\mathbf{\hat{\rho}}) == \text{EXEC}_i(\mathbf{\rho})\} \log \pi(a_i|z, s_t)],
\end{equation}
where $M$ denotes the length of the execution of $\rho$.
Optimizing this objective directly encourages the 
program embeddings, through supervised learning instead of RL as in $\mathcal{L}^{\text{R}}$, to be useful for action reconstruction, 
thus further ensuring that similar behaviors are encoded together 
and allowing for smooth interpolation.
Note that this policy is only used for improving learning the program embedding space not for solving the tasks of interest in the later stage.

In summary, we propose to optimize three sources of supervision to learn the program embedding space 
that allows for smooth interpolation and can be used to search for desired agent behaviors: 
(1) $\mathcal{L}^{\text{P}}$ (\myeq{eq:ploss}), the $\beta$-VAE 
objective for program reconstruction, (2) $\mathcal{L}^{\text{R}}$ (\myeq{eq:rloss}), an RL environment-state
matching loss for the reconstructed program, and (3) $\mathcal{L}^{\text{L}}$ (\myeq{eq:lloss}), a
supervised learning loss to encourage predicting the ground-truth agent
action sequences. 
Thus our combined objective is:
\begin{equation}
    \label{eq:fullloss}
    \min_{\theta, \phi, \pi} 
    \lambda_1 \mathcal{L}_{\theta, \phi}^{\text{P}}(\rho) + 
    \lambda_2 \mathcal{L}_{\theta, \phi}^{\text{R}}(\rho) + 
    \lambda_3 \mathcal{L}_\pi^{\text{L}}(\rho, \pi),
\end{equation}
where $\lambda_1$, $\lambda_2$, and $\lambda_3$
are hyperparameters
controlling the importance of each loss.
Optimizing the combination of these losses encourages 
the program embedding to be both semantically and syntactically informative.
More training details can be found in \mysecref{sec:training}.

\vspacesubsection{Latent Program Search: Synthesizing a Task-Solving Program} 
\label{subsec:stage2}

Once the program embedding space is learned,
our goal becomes searching for a latent program 
that maximizes the reward described by a given task MDP.
To this end, 
we adapt the Cross Entropy Method (CEM)
\citep{rubinstein1997optimization}, 
a gradient-free continuous search algorithm, 
to iteratively search over the program embedding space.
Specifically, 
we 
(1) sample a distribution of latent programs,
(2) decode the sampled latent programs into programs using the learned program decoder $p_\theta$,
(3) execute the programs in the task environment and obtain the corresponding rewards,
and (4) update the CEM sampling distribution based on the rewards.
This process is repeated until either convergence or the
maximum number of sampling steps has been reached.

\vspacesection{Experiments}

\label{sec:experiments}

We first introduce the environment and the tasks in~\mysecref{sec:karel} 
and describe the experimental setup in~\mysecref{sec:setup}. 
Then, we justify the design of \method\\ by
conducting extensive ablation studies in~\mysecref{sec:ablation}.
We describe the baselines used for comparison in~\mysecref{sec:baselines}, 
followed by the experimental results presented in~\mysecref{sec:result}.
In~\mysecref{sec:scalability}, 
we conduct experiments to evaluate the ability of our method
to generalize to a larger state space without further learning.
Finally, we investigate how \method\\' interpretability can be leveraged by conducting experiments that allow
humans to debug and improve the programs synthesized by \method\\
in~\mysecref{sec:interpretability}

\begin{figure*}
    \centering
    \begin{subfigure}[b]{0.1475\textwidth}
    \centering
    \includegraphics[trim=30 16 30 45, clip,height=\textwidth]{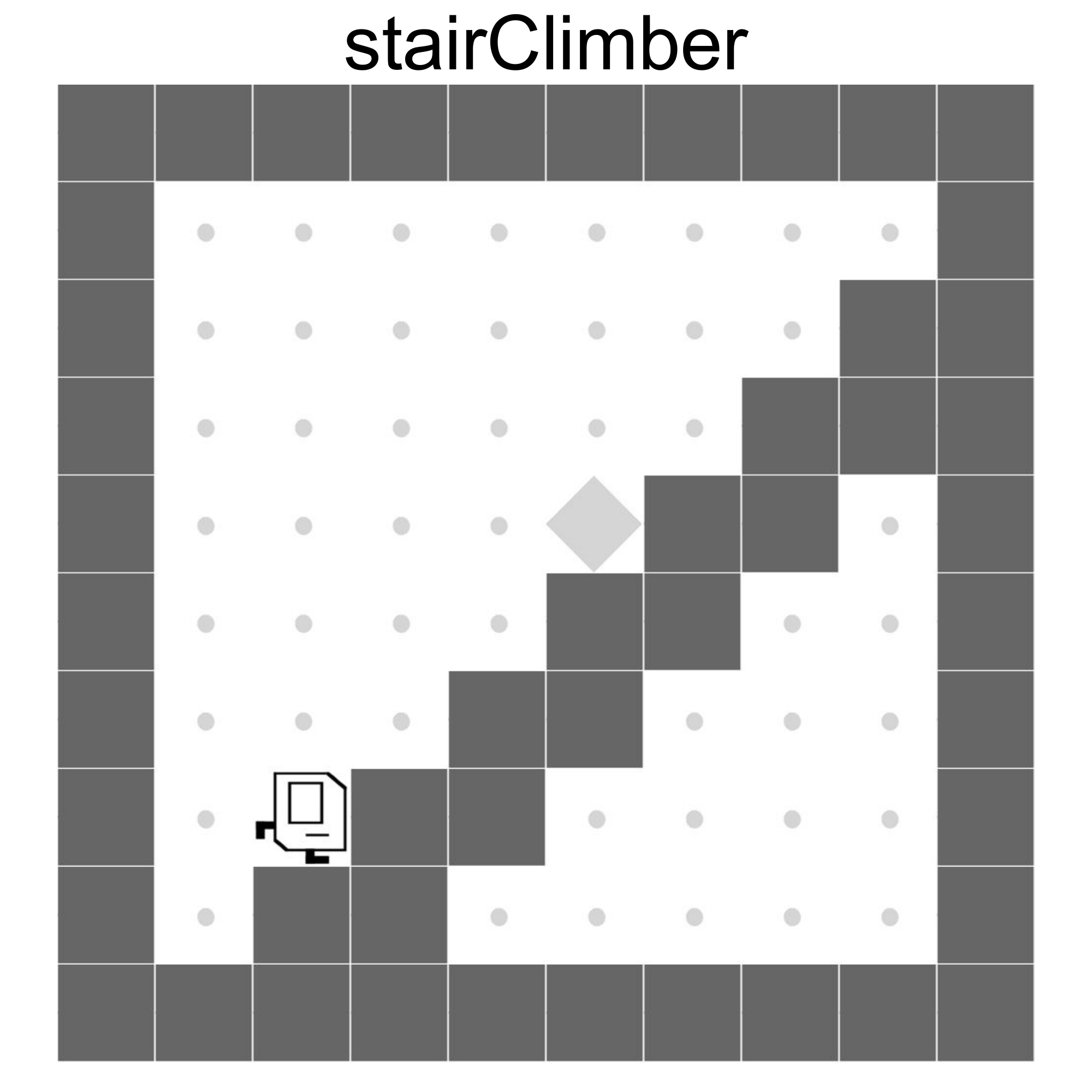}
    \SmallCaption{\textsc{StairClimber}}
    \end{subfigure}
    \begin{subfigure}[b]{0.1475\textwidth}
    \centering
    \includegraphics[trim=30 16 30 45, clip,height=\textwidth]{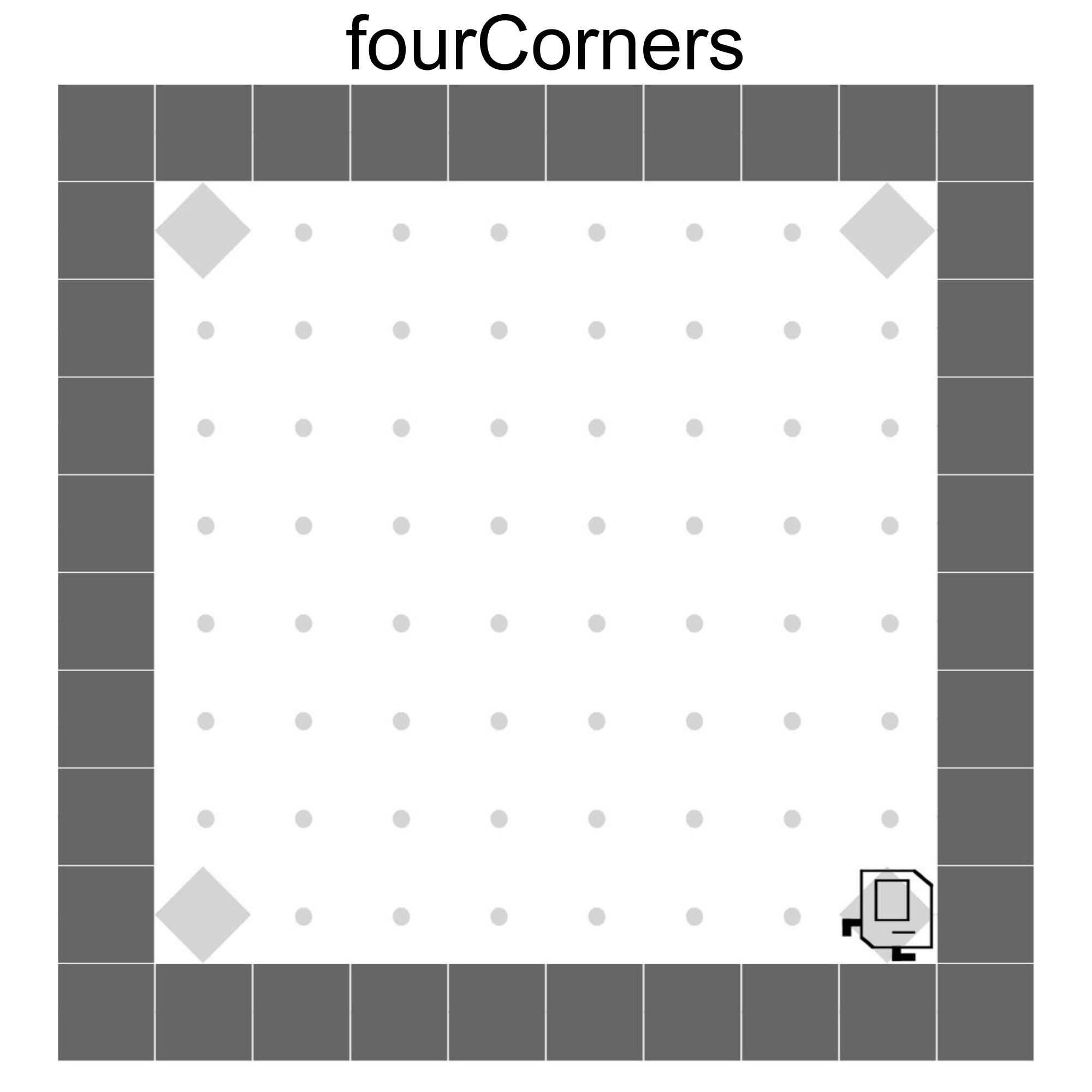}
    \SmallCaption{\textsc{FourCorner}}
    \end{subfigure}
    \begin{subfigure}[b]{0.1475\textwidth}
    \centering
    \includegraphics[trim=30 16 30 45, clip,height=\textwidth]{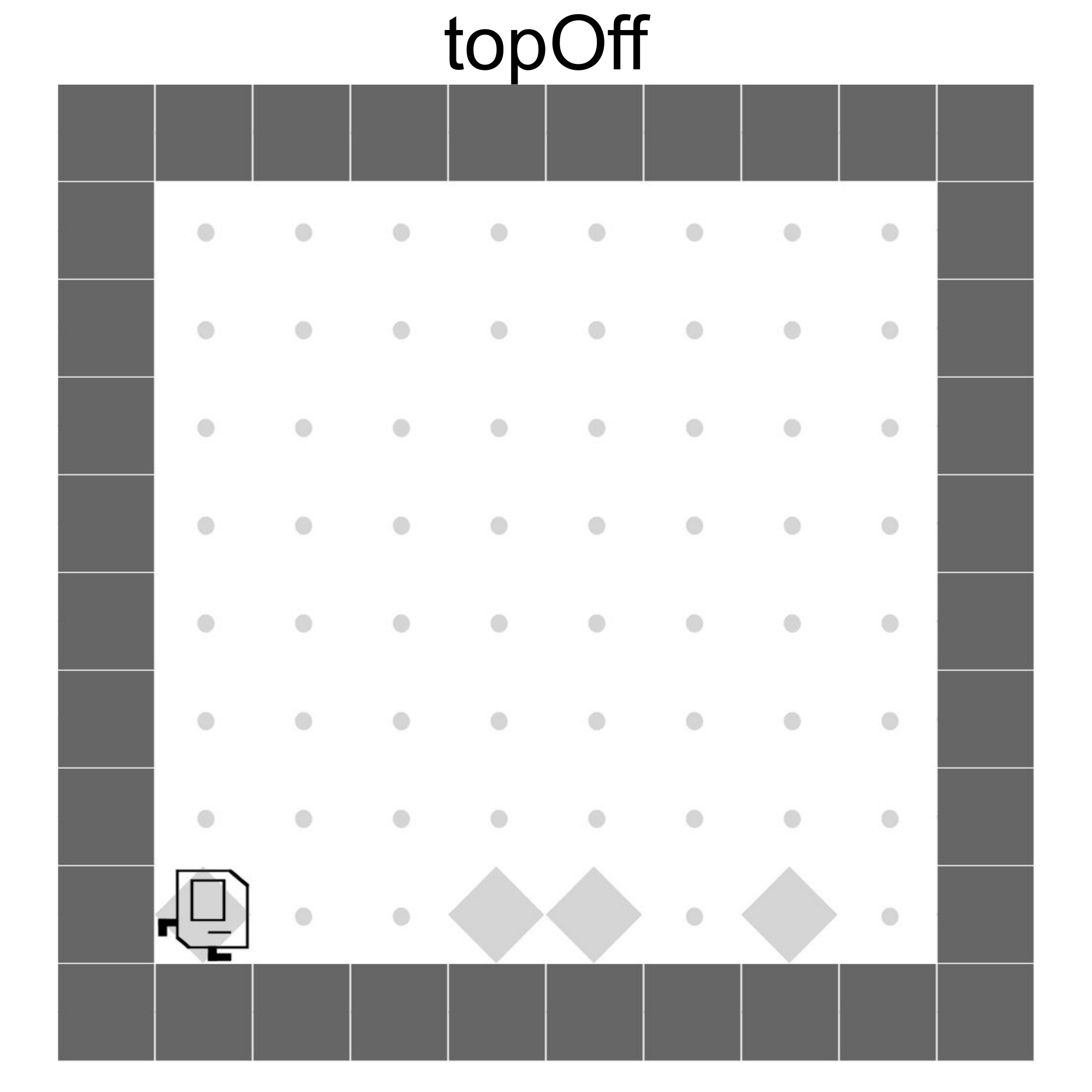}
    \SmallCaption{\textsc{TopOff}}
    \end{subfigure}
    \begin{subfigure}[b]{0.1475\textwidth}
    \centering
    \includegraphics[trim=30 16 30 45, clip,height=\textwidth]{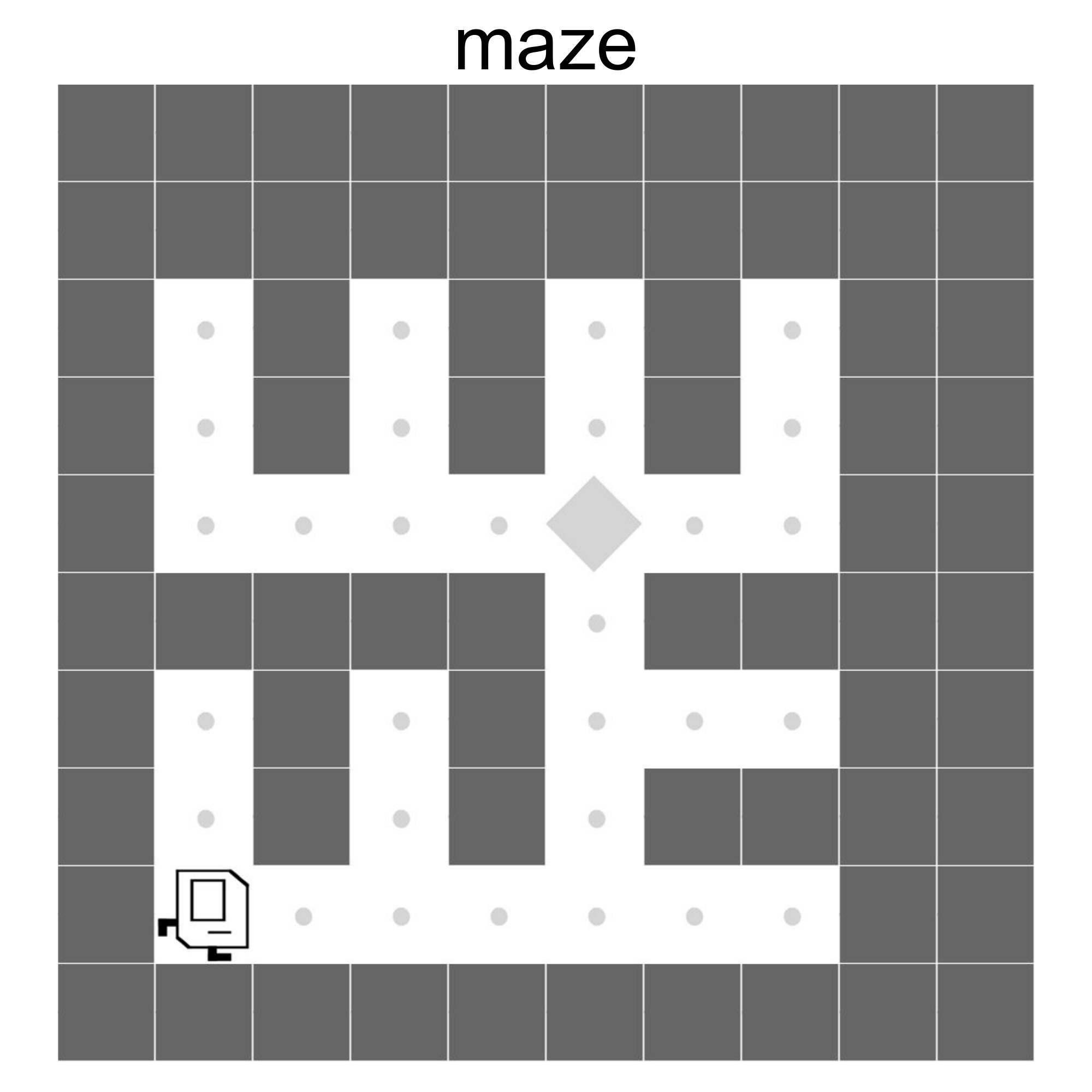}
    \SmallCaption{\textsc{Maze}}
    \end{subfigure}
    \begin{subfigure}[b]{0.23325\textwidth}
    \centering
    \includegraphics[trim=50 0 50 40, clip,width=\textwidth]{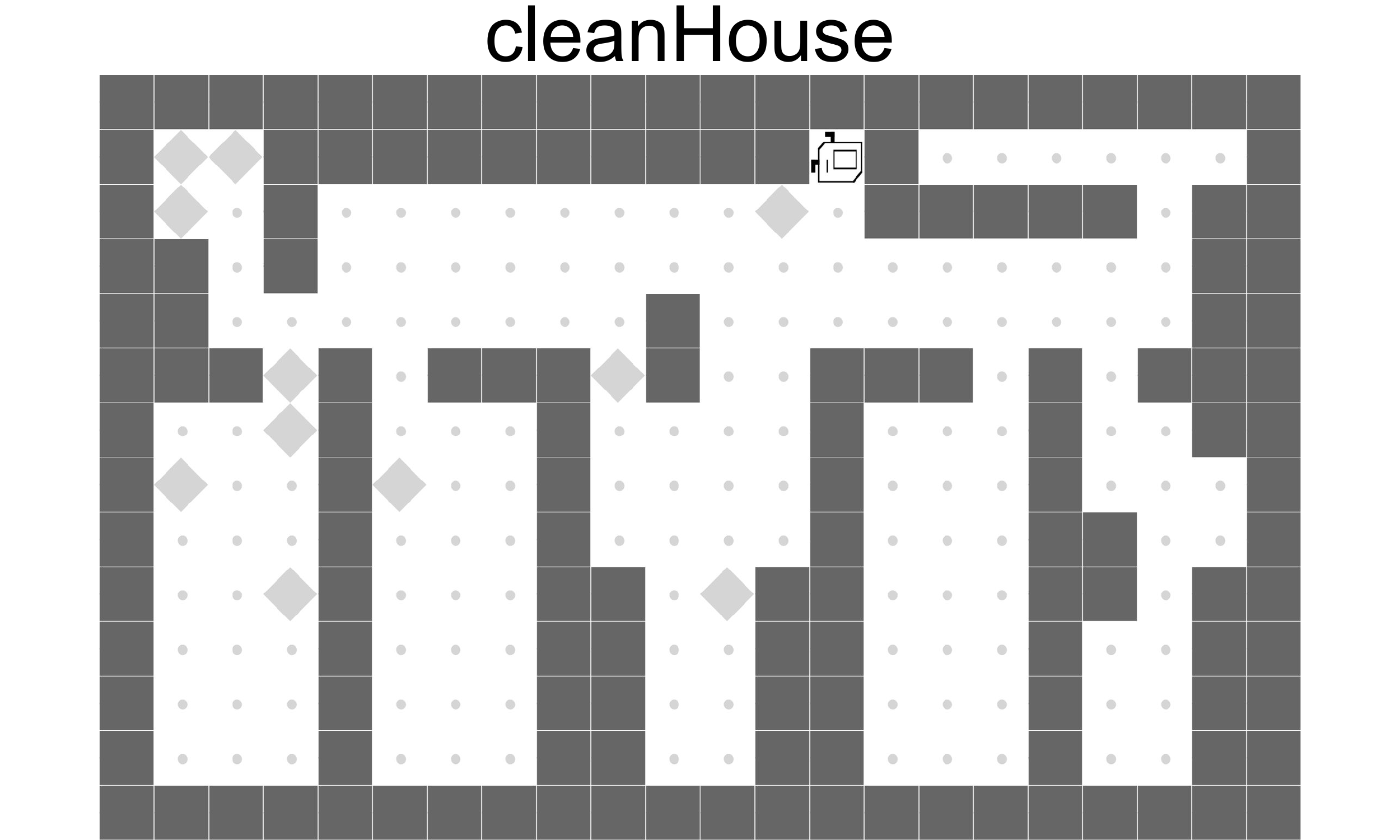}
    \SmallCaption{\textsc{CleanHouse}}
    \end{subfigure}
    \begin{subfigure}[b]{0.1475\textwidth}
    \centering
    \includegraphics[trim=30 16 30 45, clip,height=\textwidth]{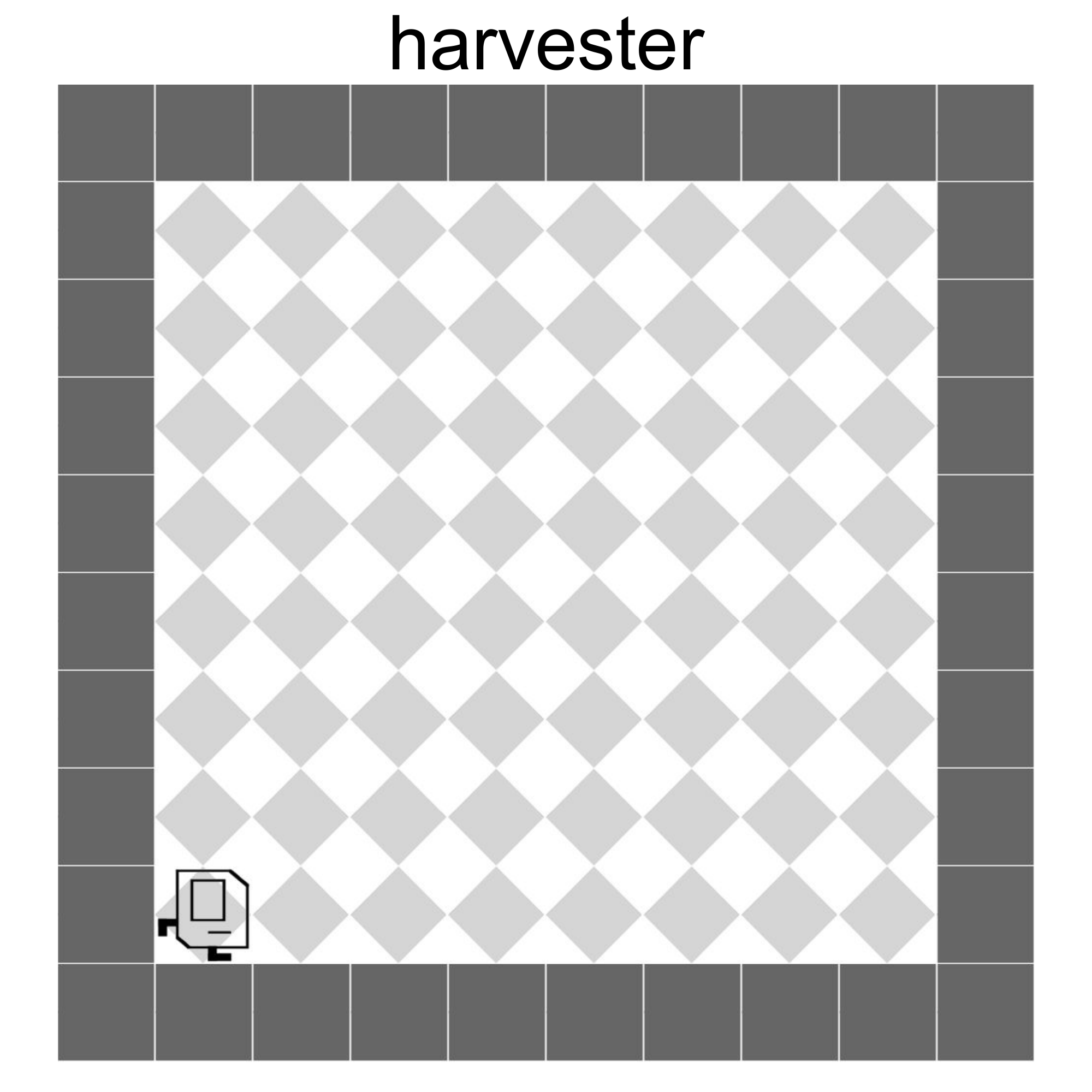}
    \SmallCaption{\textsc{Harvester}}
    \end{subfigure}
    \caption[]{
    \textbf{The Karel problem set}: 
    the domain features an agent navigating through a gridworld with walls and interacting with markers,
    allowing for designing tasks that demand certain behaviors.
    The tasks are further described in~\mysecref{sec:env details} 
    with visualizations 
    in~\myfig{fig:karel both envs}.}
    \label{fig:karel envs}
\end{figure*}

\vspacesubsection{Karel domain}
\label{sec:karel}

To evaluate the proposed framework, 
we consider the Karel domain~\citep{pattis1981karel}, 
as featured in~\cite{bunel2018leveraging, shin2018improving, sun2018neural},
which features an agent navigating through a gridworld with walls 
and interacting with markers. 
The agent has 5 actions for moving 
and interacting with marker 
and 5 perceptions for detecting obstacles and markers. 
The tasks of interest are shown in~\myfig{fig:karel envs}. Note that most tasks have randomly sampled agent, wall, marker, and/or goal configurations. When either training or evaluating, we randomly sample initial configurations upon every episode reset.
More details can be found in~\mysecref{sec:env details}.

\vspacesubsection{Programs}
\label{sec:setup}

To produce programs for learning the program embedding space,
we randomly generated a dataset of 50,000 unique programs. 
Note that the programs are generated independently of any Karel tasks; 
each program is created only by sampling tokens from the DSL, 
similar to the procedures used in~\cite{
devlin2017robustfill, bunel2018leveraging, chen2018executionguided, shin2018improving, sun2018neural, plans}. 
This dataset is split into 
a training set with 35,000 programs 
a validation set with 7,500 programs, 
and a testing set with 7,500 programs.
The validation set is used to select 
the learned program embedding space
to use for the program synthesis stage.

For each program, 
we sample random Karel states and execute the program
on them from different starting states 
to obtain $10$ environment rollouts
to compute the program behavior reconstruction loss 
$\mathcal{L}^{\text{R}}$
and the latent behavior reconstruction loss
$\mathcal{L}^{\text{L}}$
when learning the program embedding space.
We perform checks to ensure rollouts cover all execution branches
in the program so that they are representative of all aspects
of the program's behavior.
The maximum length of the programs is $44$ tokens and
the average length is $17.9$. 
We plot a histogram of their lengths in~\myfig{fig:data histogram} (in Appendix).
More dataset generation details can be found 
in~\mysecref{sec:dataset details}.

\vspacesubsection{Ablation Study}
\label{sec:ablation}
We first ablate various components of our proposed framework in order to (1) justify the necessity of the proposed two-stage learning scheme and (2) identify the effects of the proposed objectives.
We consider the following baselines and ablations of our method
(illustrated \mysecref{sec:illustration}).

\begin{itemize}
    \item Na\"{i}ve: a 
    program synthesis baseline that learns to directly synthesize a program from scratch by recurrently predicting a sequence of program tokens.
    This baseline investigates if an end-to-end learning method can solve the problem.
    More details can be found in \mysecref{sec:RL-rho appendix}.
    \item \method\\-P: 
    the simplest ablation of \method\\, in which
    the program embedding space is learned by only optimizing the program reconstruction loss $\mathcal{L}^{\text{P}}$ (\myeq{eq:ploss}). 
    \item \method\\-P+R: 
    an ablation of \method\\ which optimizes both 
    the program reconstruction loss $\mathcal{L}^{\text{P}}$ (\myeq{eq:ploss})
    and the program behavior reconstruction loss $\mathcal{L}^{\text{R}}$ (\myeq{eq:rloss}). 
    \item \method\\-P+L: 
    an ablation of \method\\ which optimizes both 
    the program reconstruction loss $\mathcal{L}^{\text{P}}$ (\myeq{eq:ploss}) 
    and the latent behavior reconstruction loss $\mathcal{L}^{\text{L}}$ (\myeq{eq:lloss}).
    \item \method\\ (\method\\-P+R+L): 
    \method\\ with all the losses, optimizing our full objective in \myeq{eq:fullloss}. 
    \item \method\\-rand-\{8/64\}: 
    similar to \method\\, this ablation also optimizes the full objective (\myeq{eq:fullloss}) for learning the program embedding space. 
    Yet, when searching latent programs, instead of CEM,
    it simply randomly samples 8/64 candidate latent programs and 
    chooses the best performing one. 
    These baselines justify the effectiveness of using CEM for searching latent programs.
\end{itemize}

\begin{table}[ht]
\centering
\caption[]{Program behavior reconstruction rewards (standard deviations) across all methods.}
\scalebox{0.8}{
\begin{tabular}{@{}cccccc@{}}\toprule
& \textsc{WHILE} & \textsc{IFELSE+WHILE} & \textsc{2IF+IFELSE} & \textsc{WHILE+2IF+IFELSE} & Avg Reward\\ 
\cmidrule{2-6}
Na\"{i}ve & 0.65 (0.33) & 0.83 (0.07) & 0.61 (0.33) & 0.16 (0.06) & 0.56 \\ 
\method\\-P & 0.95 (0.13) & 0.82 (0.08) & 0.58 (0.35) & 0.33 (0.17) & 0.67\\
\method\\-P+R & 0.98 (0.09) & 0.77 (0.05) & 0.63 (0.25) & 0.52 (0.27) & 0.72\\
\method\\-P+L & \textbf{1.06} (0.00) & 0.84 (0.10) & 0.77 (0.23) & 0.33 (0.13) & 0.75\\
\method\\-rand-8 & 0.62 (0.24) & 0.49 (0.09) & 0.36 (0.18) & 0.28 (0.14) & 0.44\\
\method\\-rand-64 & 0.78 (0.22) & 0.63 (0.09) & 0.55 (0.20) & 0.37 (0.09) & 0.58\\
\method\\ & \textbf{1.06} (0.08) & \textbf{0.87} (0.13) & \textbf{0.85} (0.30) & \textbf{0.57} (0.23) & \textbf{0.84} \\ 
\bottomrule
\end{tabular}}
\label{table:program convergence}
\end{table}

\textbf{Program Behavior Reconstruction.}
To determine the effectiveness of the proposed two-stage learning scheme and the learning objectives,
we measure how effective each ablation is at reconstructing
the behaviors of input programs.
We use programs from the test set (shown in \myfig{fig:reconstruction tasks} in Appendix), 
and utilize the environment state matching reward
$R_{\text{mat}}(\hat{\rho}, \rho)$ (\myeq{eq:RL Reward}),
with a $0.1$ bonus for synthesizing a syntactically correct program. 
Thus the return ranges between $[0, 1.1]$. 
We report the mean cumulative return, over 5 random seeds, 
of the final programs after convergence. 

The results are reported in~\mytable{table:program
convergence}. 
Each test is named after its control flows 
(\eg \textsc{IFELSE+WHILE} has an if-else statement and a while loop).
The na\"{i}ve program synthesis baseline fails on the complex \textsc{WHILE+2IF+IFELSE} program,
as it rarely synthesizes conditional
and loop statements, 
instead generating long sequences of 
action tokens that attempt to replicate the desired behavior
of those statements (see synthesized programs in~\myfig{fig:program reconstruction examples}).
We believe that this is because it is incentivized to initially predict action tokens to
gain more immediate reward, making it less likely to synthesize other tokens.
\method\\ and its variations perform better 
and synthesize more complex programs, 
demonstrating the importance of the proposed two-stage learning scheme in 
biasing program search.
We also note that \method\\-P 
achieves the worst performance out of the CEM search \method\\ ablations, 
indicating that optimizing the program reconstruction loss 
$\mathcal{L}^{\text{P}}$ (the VAE loss) alone does not yield
satisfactory results.
Jointly optimizing $\mathcal{L}^{\text{P}}$ with either 
the program behavior reconstruction loss $\mathcal{L}^{\text{R}}$ 
or the latent behavior reconstruction loss $\mathcal{L}^{\text{L}}$
improves the performance,
and optimizing our full objective with all three achieves the best performance across all tasks,
indicating the effectiveness of the proposed losses.
Finally, \method\\ outperforms \method\\-rand-8/64, 
suggesting the necessity of adopting better search algorithms such as CEM.

\begin{wraptable}[10]{r}{0.54\textwidth}
    \vspace{-0.4cm}
    \centering
    \caption[]{\small
    Program embedding space smoothness.
    For each program, 
    we execute the ten nearest programs 
    in the learned embedding space of each model 
    to calculate
    the mean state-matching reward $R_{\text{mat}}$ against the original program.
    We report $R_{\text{mat}}$ averaged over all programs
    in each dataset.
    \vspace{-0.1cm}
    }
    \label{table:latent smoothness}
    \scalebox{0.76}{\begin{tabular}{@{}cccccccccccc@{}}\toprule
& \method\\-P & \method\\-P+R & \method\\-P+L & \method\\ \\
\cmidrule{2-6}
\textsc{Training} 
& 0.22 
& 0.22 & \textbf{0.31}  & \textbf{0.31} \\ 
\textsc{Validation} 
& 0.22 
& 0.21 & \textbf{0.27}  & \textbf{0.27} \\
\textsc{Testing} 
& 0.22
& 0.22 & \textbf{0.28} & 0.27
\\
\bottomrule
\end{tabular}}
\end{wraptable}

\textbf{Program Embedding Space Smoothness.}
We investigate if 
the program and latent behavior reconstruction losses 
encourage learning a behaviorially smooth embedding space.
To quantify behavioral smoothness, 
we measure how much a change 
in the embedding space corresponds to a
change in behavior 
by comparing execution traces. 
For all programs we compute the pairwise Euclidean distance between
their embeddings in each model. 
We then calculate the environment state matching distance $R_{\text{mat}}$ between the
decoded programs by executing them from the same initial state.

The results are reported in~\mytable{table:latent smoothness}. 
\method\\ and \method\\-P+L perform the best, suggesting
that optimizing the latent behavior reconstruction objective
$\mathcal{L}^{\text{L}}$, in \myeq{eq:lloss}, is essential
for improving the smoothness of the latent space in terms of execution behavior. 
We further analyze and visualize the learned program embedding space in~\mysecref{sec:appendix latent space} and~\myfig{fig:latent vis} (in Appendix).

\vspacesubsection{Baselines}
\label{sec:baselines}

We evaluate \method\\ against the following baselines (illustrated in \myfig{fig:baselines} in Appendix \mysecref{sec:illustration}).
\begin{itemize}
    \item DRL: a neural network policy trained on each task 
    and taking raw states (Karel grids) as input.
    \item DRL-abs: a recurrent neural network policy directly trained on each Karel task but taking \textit{abstract} states as input
    (\ie it sees the same perceptions as \method\\,
    \eg \texttt{frontIsClear()==true}).
    \item DRL-abs-t: a DRL transfer learning baseline 
    in which for each task, 
    we train DRL-abs policies on all other tasks,
    then fine-tune them on the current
    task. Thus it acquires a prior by learning to first solve other Karel tasks.
    Rewards are reported for the policies
    from the task that transferred with highest return.
    We only transfer DRL-abs policies as some tasks have different state spaces. 
    \item HRL: a hierarchical RL baseline in which a VAE is first trained on 
    action sequences from program execution traces used by \method\\. 
    Once trained, the decoder is utilized as a low-level policy
    for learning a high-level policy to sample actions from.
    Similar to \method\\, this baseline utilizes the dataset to produce a prior of the domain.
    It takes raw states (Karel grids) as input.
    \item HRL-abs: the same method as HRL but taking 
    abstract states (\ie 
    local perceptions) as input. 
    \item VIPER~\citep{bastani2018verifiable}: A decision-tree programmatic policy which imitates the behavior of a deep RL teacher policy via a modified DAgger algorithm \citep{ross2011reduction}. This decision tree policy cannot synthesize loops, allowing us to highlight the performance advantages of more expressive program representation that \method\\ is able to take advantage of.
\end{itemize}

All the baselines are trained with PPO \citep{schulman2017proximal} 
or SAC \citep{haarnoja18b}, 
including the VIPER teacher policy.
More training details can be found in \mysecref{sec:hyperparameters}.

\vspacesubsection{Results}
\label{sec:result}

\begin{table}
\centering
\caption[]{Mean return (standard deviation) of all methods across Karel tasks, 
evaluated over 5 random seeds. DRL methods, program synthesis baselines, and \method\\ ablations are separately grouped.}
\scalebox{0.8}{\begin{tabular}{@{}ccccccc@{}}\toprule
& \textsc{StairClimber} & \textsc{FourCorner} & \textsc{TopOff} & \textsc{Maze} & \textsc{CleanHouse} & \textsc{Harvester}\\
\cmidrule{2-7}
DRL & \textbf{1.00} (0.00) & 0.29 (0.05) & 0.32 (0.07) & \textbf{1.00} (0.00) & 0.00 (0.00) & \textbf{0.90} (0.10)\\
DRL-abs & 0.13 (0.29) & 0.36 (0.44) & 0.63 (0.23) & \textbf{1.00} (0.00) & 0.01 (0.02) & 0.32 (0.18) \\
DRL-abs-t & 0.00 (0.00) & 0.05 (0.10) & 0.17 (0.11) & \textbf{1.00} (0.00) & 0.01 (0.02) & 0.16 (0.18) \\
HRL & -0.51 (0.17) & 0.01 (0.00) & 0.17 (0.11) & 0.62 (0.05) & 0.01 (0.00) & 0.00 (0.00) \\
HRL-abs & -0.05 (0.07) & 0.00 (0.00) & 0.19 (0.12) & 0.56 (0.03) & 0.00 (0.00) & -0.03 (0.02)\\
\midrule
Na\"{i}ve & 0.40 (0.49) & 0.13 (0.15) & 0.26 (0.27) & 0.76 (0.43) & 0.07 (0.09) & 0.21 (0.25) \\
VIPER & 0.02 (0.02) & 0.40 (0.42) & 0.30 (0.06) & 0.69 (0.05) & 0.00 (0.00) & 0.51 (0.07)\\
\midrule
\method\\-rand-8 & 0.10 (0.17) & 0.10 (0.14) & 0.28 (0.05) & 0.40 (0.50) & 0.00 (0.00) & 0.07 (0.06) \\
\method\\-rand-64 & 0.18 (0.40) & 0.20 (0.11) & 0.33 (0.07) & 0.58 (0.41) & 0.03 (0.06) & 0.12 (0.05) \\
\method\\ & \textbf{1.00} (0.00) & \textbf{0.45} (0.40) & \textbf{0.81} (0.07) & \textbf{1.00} (0.00) & \textbf{0.18} (0.14) & 0.45 (0.28) \\
\bottomrule
\end{tabular}}
\label{table:karel comparison}
\end{table}

We present the results of the baselines and our method
evaluated on the Karel task set based on the environment rewards 
in~\mytable{table:karel comparison}.
The reward functions are sparse for all tasks, 
and are normalized such that the 
final cumulative return is within $[-1, 1]$ 
for tasks with penalties and $[0, 1]$ for tasks without;
reward functions for each task are detailed
in~\mysecref{sec:env details}. 

\textbf{Overall Task Performance.} 
Across all but one task, \method\\ yields the best performance. 
The \method\\-rand baselines perform significantly worse than \method\\ on all Karel tasks, 
demonstrating the need for using a search algorithm 
like CEM during synthesis.
The performance of VIPER is bounded by its RL teacher policy, 
and therefore is outperformed by the DRL baselines
on most of the tasks.
Meanwhile, DRL-abs-t is generally
unable to improve upon DRL-abs across the board,
suggesting that transferring Karel behaviors with RL 
from one task to another is ineffective. 
Furthermore, both the HRL baselines achieve poor performance, 
likely because agent actions alone provide insufficient supervision for a VAE 
to encode useful action trajectories on unseen tasks---unlike programs.
Finally, the poor performance
of the na\"{i}ve program synthesis baseline
highlights the difficulty and inefficiency of
learning to synthesize programs from scratch using only 
rewards.
In the appendix, we present programs synthesized by 
\method\\ in~\myfig{fig:karel program examples}, 
example optimal programs for each task in 
\mysecref{sec:appendix programmatic policies}
(\myfig{fig:reconstruction tasks}), 
rollout visualizations
in \myfig{fig:example rollouts}, and
additional results analysis in~\mysecref{sec:additional_analysis}.

\textbf{Repetitive Behaviors.} 
Solving \textsc{StairClimber} and \textsc{FourCorner} requires
acquiring repetitive (or looping) behaviors.
\textsc{StairClimber},
which can be solved by repeating a short, 4-step stair-climbing 
behavior until the goal marker is reached, 
is not solved by DRL-abs. 
\method\\ fully solves the task given the same perceptions,
as this behavior can be simply represented 
with a while loop that repeats the stair-climbing skill. 
However VIPER performs poorly as its decision tree 
cannot represent such loops.
Similarly, the baselines are unable
to perform as well on \textsc{FourCorner}, 
a task in which the agent must pickup a marker 
located in each corner of the grid. 
This behavior takes at least 14 timesteps to complete, 
but can be represented by two nested loops.
Similar to \textsc{StairClimber}, the bias introduced
by the DSL and our generated dataset (which includes nested
loops), 
results in \method\\ being able to perform much better.

\textbf{Exploration.} 
\textsc{TopOff}
rewards the agent for adding markers to locations with
existing markers. 
However, there are no restrictions for the agent to wander elsewhere around the environment, 
thus making exploration a problem for the RL baselines, 
and thereby also constraining VIPER. 
\method\\ performs best on this task, 
as the ground-truth program can be represented by a simple loop 
that just moves forward and places markers when a marker is detected.
\textsc{Maze} also involves exploration, however its small size ($8 \times 8$) results in many methods, 
including \method\\, solving the task.

\textbf{Complexity.} 
Solving \textsc{Harvester} and \textsc{CleanHouse}
requires acquiring complex behaviors,
resulting in poor performance from all methods.
\textsc{CleanHouse} requires an agent to navigate
through a house and pick up all markers along
the walls on the way. 
This requires repeated execution of a skill, of varied
length,
which navigates around the house, 
turns into rooms, and picks up markers. 
As such, all baselines perform very poorly. 
However, \method\\ is able to perform 
substantially better because these behaviors 
can be represented by a program of medium complexity 
with a while loop and some nested conditional statements.
On the other hand, 
\textsc{Harvester} involves simply navigating to and picking up
a marker on every spot on the grid. 
However, this is a difficult program to synthesize 
given our random dataset generation process; 
the program we manually 
derive to solve \textsc{Harvester} is long
and more syntactically complex than most training programs. 
As a result, DRL and VIPER outperform \method\\ on this task.

\textbf{Learned Program Embedding Space.} More analysis on our learned program embedding space can be found in the appendix.
We present CEM search trajectory visualizations in ~\mysecref{sec:cem_vis},
demonstrating how the search population's rewards change over time.
To qualitatively investigate the smoothness of the learned program embedding space, 
we linearly interpolate between pairs of latent programs
and display their corresponding decoded programs in
\mysecref{sec:interpolation}.
In~\mysecref{sec:evolution}, 
we illustrate how predicted programs evolve over the course of CEM search.

\vspacesubsection{Generalization}
\label{sec:scalability}

\begin{wraptable}[6]{r}{0.35\textwidth}
    \vspace{-0.4cm}
    \centering
    \caption[]{\small
    Rewards on $100\times100$ grids.
    \vspace{-0.55cm}
    }
    \label{table:transfer learning comparison}
    \scalebox{0.76}{\begin{tabular}{@{}ccc@{}}\toprule
    & \textsc{StairClimber} & \textsc{Maze} \\
    \centering
    \small
    DRL & 0.00 (0.00) & 0.00 (0.00) \\
    DRL-abs & 0.00 (0.00) & 0.04 (0.05)\\
    VIPER & 0.00 (0.00) & 0.10 (0.12)\\
    \method\\ & \textbf{1.00} (0.00) & \textbf{1.00} (0.00)\\
    \bottomrule
    \end{tabular}}
\end{wraptable}

We are also interested in learning whether 
the baselines and the programs synthesized by \method\\ 
can generalize to novel scenarios without further learning.
Specifically, we investigate how well they can generalize to larger state spaces.
We expand both \textsc{StairClimber}
and \textsc{Maze} to $100 \times 100$ grid sizes 
(from $12 \times 12$ and $8 \times 8$, respectively).
We directly evaluate the policies or programs
obtained from the original tasks
with smaller state spaces for all methods except DRL 
(its observation space changes), 
which we retrain from scratch.
The results are shown in~\mytable{table:transfer learning comparison}.
All baselines perform significantly worse than before on both tasks.
On the contrary, the programs synthesized by \method\\ for the smaller task instances  
achieve zero-shot generalization to larger task instances without losing any performance. 
Larger grid size experiments for the other Karel tasks and additional unseen configuration experiments can be found in
\mysecref{sec:appendix_generalization}.

\vspacesubsection{Interpretability}
\label{sec:interpretability}

Interpretability in machine learning~\cite{lipton2018mythos, shen2020} is particularly crucial
when it comes to learning a policy that interacts with the environment~\citep{carl-zhang20esafe, Hewing_2019safe, FisacAZKGT17safe, 
hakobyan2015safe, Sadigh-RSS-16safe, berkenkamp2017safe, 
saved-thananjeyansafe, aswani2013provablysafe}.
The proposed framework produces programmatic policies 
that are interpretable from the following aspects as outlined in~\citep{shen2020}.
\begin{itemize}
    \item Trust: 
    interpretable machine learning methods and models may more easily be trusted since humans tend to be reluctant to trust systems that they do not understand.
    Programs synthesized by \method\\ can naturally be better trusted since one can simply read and interpret them.
    \item Contestability: 
    the program execution traces produce a chain of reasoning for each action, providing insights on the induced behaviors and thus allowing for contesting improper decisions.
    \item Safety: 
    synthesizing readable programs
    allows for diagnosing issues earlier (\ie before execution)
    and provides opportunities to intervene,
    which is especially critical for safety-critical tasks.
\end{itemize}

In the rest of this section, we investigate how 
the proposed framework enjoys interpretability from the three aforementioned aspects.
\Skip{
In the rest of this section, we investigate how 
the proposed framework can satisfy the safety property of interpretability, \ie how the policy produced by our framework provides the opportunities to diagnose issues earlier and allows for interventions.
}
Specifically, synthesized programs are not only readable to human users but also interactive, allowing non-expert users with a basic understanding of programming to diagnose and make edits to improve their performance. 
To demonstrate this, we asked non-expert humans to read, interpret, and edit suboptimal \method\\ policies to improve their performance. 
Participants edited \method\\ programs on 3 Karel tasks with suboptimal reward: \textsc{TopOff}, \textsc{FourCorner}, and \textsc{Harvester}.
With just 3 edits, participants obtained a mean reward improvement of 
97.1\%, and with 5 edits, participants improved it by 125\%. 
This justifies how our synthesized policies can be manually diagnosed and improved, 
a property which DRL methods lack. 
More details and discussion can be found in~\mysecref{sec:debug}.

\vspacesection{Discussion}
\label{sec:conclusion}
We propose a framework for solving tasks described by MDPs by producing programmatic policies that are more interpretable and generalizable than neural network policies learned by deep reinforcement learning methods. Our proposed framework adopts a flexible program representation and requires only minimal supervision compared to prior programmatic reinforcement learning and program synthesis works. Our proposed two-stage learning scheme not only alleviates the difficulty of learning to synthesize programs from scratch but also enables reusing its learned program embedding space for various tasks. The experiments demonstrate that our proposed framework outperforms DRL and programmatic baselines on a set of Karel tasks by producing expressive and generalizable programs that can consistently solve the tasks. Ablation studies justify the necessity of the proposed two-stage learning scheme as well as the effectiveness of the proposed learning objectives.

While the proposed framework achieves promising results,
we would like to acknowledge two assumptions that are implicitly made in this work.
First, we assume the existence of a program executor that can produce execution traces of programs.
This program executor needs to be able to return perceptions from the environment state as well as
apply actions to the environment.
While this assumption is widely made in program synthesis works, 
a program executor can still be difficult to obtain
when it comes to real-world robotic tasks. 
Fortunately, in research fields such as computer vision or robotics,
a great amount of effort has been put into
satisfying this assumption
such as designing modules that can return high-level abstraction of raw sensory input (\eg with object detection networks, proximity/tactile sensors, etc.).

Secondly, we assume that it is possible to generate a distribution of programs whose behaviors are 
at least remotely related to the desired behaviors 
for solving the tasks of interest.
It can be difficult to synthesize programs 
which represent behaviors that are more complex
than ones in the training program distribution, although
one possible solution is to 
employ
a better program generation process 
to generate programs that induce more complex behaviors.
Also, the choice of DSL plays an important role in how complex the programs can be.
Ideally, employing a more complex DSL would allow our proposed framework to synthesize more advanced agent behaviors.

In the future, we hope to extend the proposed framework to 
more challenging domains such real-world robotics.
We believe this framework would allow for 
deploying robust, interpretable policies 
for safety-critical tasks 
such as robotic surgeries. 
One way to make \method\\ applicable to robotics domains would be
to simultaneously learn perception modules and action controllers. 
Other possible solutions include incorporating program execution methods~\citep{andreas2016modular, oh2017zero, sun2020program, yang2021program, lee2019composing, zhao2021proto}
that are designed to allow program execution or designing DSLs 
that allow pre-training of perception modules and action controllers. 
Also, the proposed framework shares some characteristics
with works in multi-task RL~\cite{oh2017zero, andreas2016modular, teh2017distral, shu2018hierarchical, sohn2018hierarchical, pmlr-v119-jain20b, pertsch2020spirl} 
and meta-learning~\cite{snell2017prototypical, vuorio2019multimodal, vinyals2016matching, finn_model-agnostic_2017, vuorio2018toward, chen2018a, lee_gradient-based_2018, nichol2018reptile, NEURIPS2020_e6385d39, chen2020learning}.
Specifically, it learns a program embedding space
from a distribution of tasks/programs. 
Once the program embedding space is learned,
it can be be reused to solve different tasks without retraining.

Yet, extending \method\\ to such domains can potentially lead to some negative
societal impacts. 
For example, our framework can still capture unintended bias during learning or 
suffer from adversarial attacks. 
Furthermore, policies deployed in the real world
can create great economic impact by causing job losses in some sectors.
Therefore, we would encourage further work to 
investigate the biases, safety issues, and potential economic impacts to ensure that the deployment in the field does not cause far-reaching, negative societal impacts.

\section*{Acknowledgments}
\label{sec:ack}

The authors appreciate the fruitful discussions with 
Karl Pertsch, Youngwoon Lee, Ayush Jain, and Grace Zhang.
The authors would like to thank Ting-Yu S. Su for contributing to the learned program embedding spaces visualizations.
The authors are grateful to Sidhant Kaushik, Laura Smith, and Siddharth Verma 
for offering their time to participate in the human debugging interpretability experiment.

\section*{Funding Transparency Statement}
    This project was partially supported by USC startup funding and by
NAVER AI Lab.
\label{sec:funding}

\renewcommand{\bibname}{References}
\bibliographystyle{unsrtnat}
\bibliography{ref}

\clearpage
\appendix

\appendix

\section*{Appendix}

\vspace{-1.8cm}

\part{} %
\parttoc %
\listoffigures
\listoftables

\clearpage

\section{Program Embedding Space Visualizations}
\label{sec:appendix latent space}

\begin{figure}[t]
    \centering
    \hfill
    \begin{subfigure}[t]{0.49\textwidth}
    	\centering
    	\includegraphics[width=\textwidth]{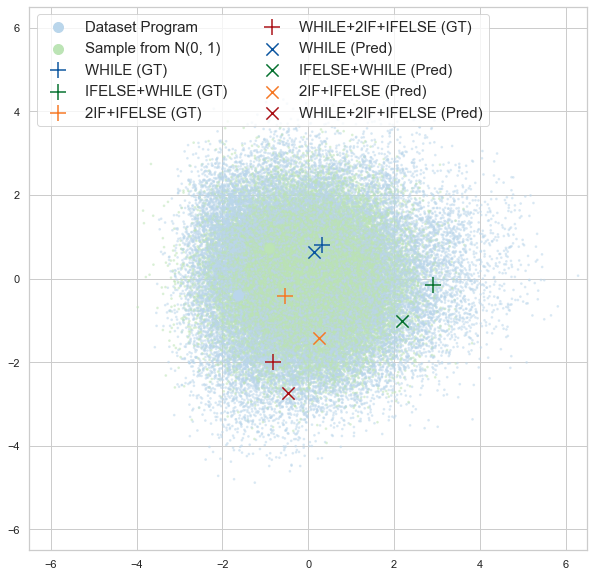}
        \caption{\textsc{\method\\-P}}
        \label{fig:vis_latent_leapsp}
    \end{subfigure}
    \hfill
    \begin{subfigure}[t]{0.49\textwidth}
    	\centering
    	\includegraphics[width=\textwidth]{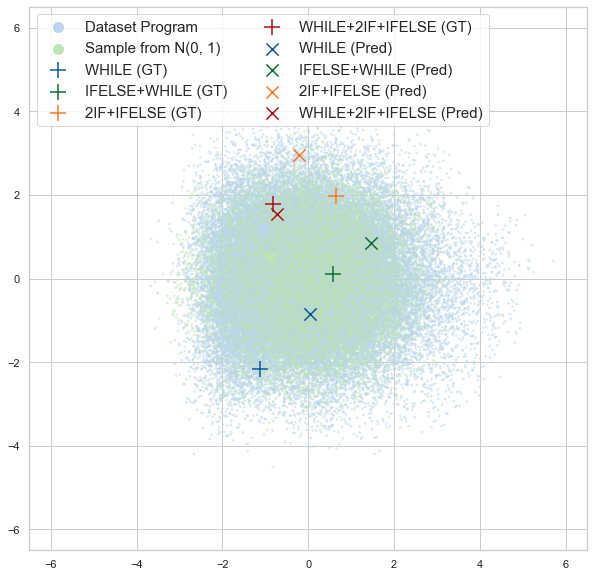}
        \caption{\textsc{\method\\-P+R}}
        \label{fig:vis_latent_leapspr}
    \end{subfigure}
    \hfill
    \\
    \hfill
    \begin{subfigure}[t]{0.49\textwidth}
    	\centering
    	\includegraphics[width=\textwidth]{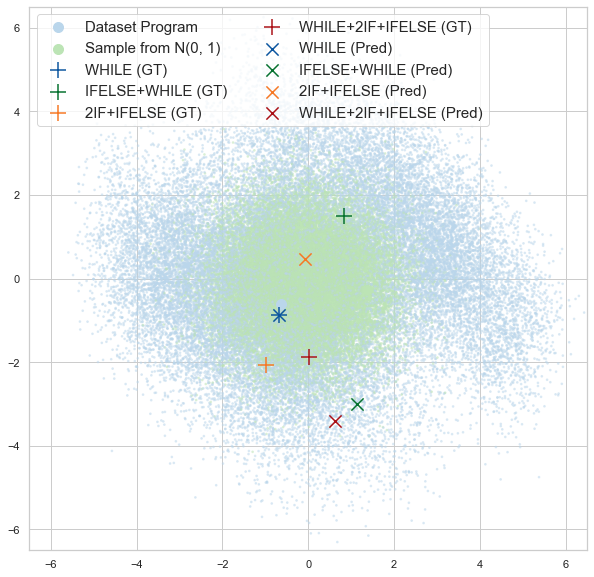}
        \caption{\textsc{\method\\-P+L}}
        \label{fig:vis_latent_leapspl}
    \end{subfigure}
    \hfill
    \begin{subfigure}[t]{0.49\textwidth}
    	\centering
    	\includegraphics[width=\textwidth]{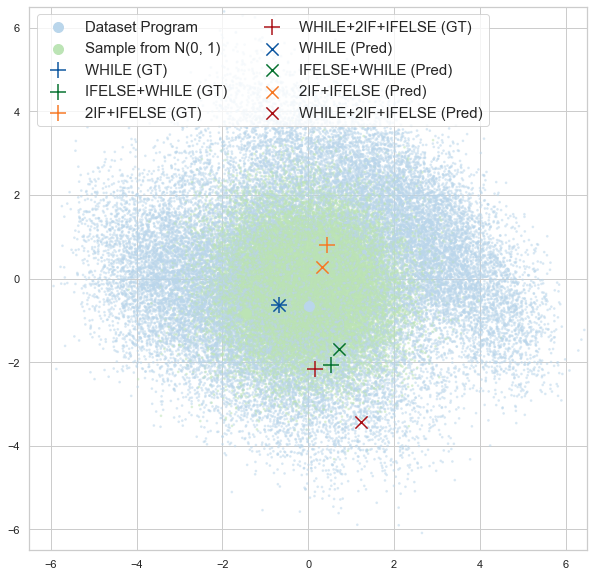}
        \caption{\textsc{\method\\}}
        \label{fig:vis_latent_leaps}
    \end{subfigure}
    \caption[Visualizations of Learned Program Embedding Space]{\textbf{Visualizations of learned program embedding space.}
    We perform dimensionality reduction with PCA
    to embed 
    encoded programs from the training dataset, 
    samples drawn from a normal distribution,
    programs from the testing dataset, 
    and programs reconstructed by models
    to a 2D space.
    The shape of the latent training programs in 
    the program embedding spaces learned by \method\\-P and \method\\-P+R are similar to a normal distribution,
    while in the program embedding spaces learned by \method\\ and \method\\-P+L, the shape is more twisted, 
    suggesting the effectiveness of the proposed latent behavior reconstruction objective.
    Moreover, the distances between pairs of ground-truth programs and their reconstructions are smaller
    in the program embedding space learned by \method\\,
    highlighting the advantage of employing both of the two proposed behavior reconstruction objectives.
    \Skip{
    2-Dimensional plots of the latent space shape---obtained with a PCA projection---for all \method\\ ablations
    (best seen in color and digitally).
    Green points are all $35,000$ training programs encoded into the latent
    program embedding space, and blue points are samples from a
    $\mathcal{N}(0_d, I_d)$ prior distribution projected down to two dimensions. The conditional policy
    ablations have a ``twisted'' latent space due to how it explicitly encourages
    the embedding to store behavioral information. We show both ground-truth programs
    and a synthesized program in the latent space for each method.}
    }
    \label{fig:latent vis}
\end{figure}

In this section, 
we present and analyze visualizations providing insights on the program embedding spaces learned by 
\method\\ and its variations.
To investigate the learned program embedding space,
we perform dimensionality reduction with PCA~\cite{pca}
to embed the following data to a 2D space for visualizations shown in~\myfig{fig:latent vis}:

\begin{itemize}
    \item Latent programs from the training dataset encoded by a learned encoder $q_\phi$, 
    visualized as blue scatters. There are 35k training programs.
    \item Samples drawn from a normal distribution $\mathcal{N}(0, 1)$, 
    visualized as green scatters.
    This is to show how a distribution would look like if
    the embedding space is learned by using a highly weighted KL-divergence penalty 
    (\ie a large $\beta$ value the VAE loss).
    We compared this against the latent program distribution
    learned by our method to justify
    the effectiveness of the proposed objectives: 
    the program behavior reconstruction loss ($\mathcal{L}^{\text{R}}$) and
    the latent behavior reconstruction loss ($\mathcal{L}^{\text{L}}$).
    \item Ground-truth (GT) test programs from the testing dataset, 
    encoded by a learned decoder $q_\phi$, 
    visualized as plus signs ($+$) with different colors.
    We selected 4 test programs.
    \item Reconstructed programs which are predicted (Pred)
    by each method given 
    visualized as crosses ($\times$) with different colors.
    Since there are 4 test programs selected,
    4 reconstructed programs are visualized.
    Each pair of test program and predicted program 
    is visualized with the same color.
    These predicted (\ie synthesized) programs are also 
    shown in~\myfig{fig:program reconstruction examples}.
\end{itemize}

\noindent \textbf{Embedding Space Coverage.}
Even though the testing programs are not in the training program dataset, 
and therefore are unseen to models,
their embedding vectors still lie in 
the distribution learned by all the models.
This indicates that the learned embedding spaces cover 
a wide distribution of programs.

\noindent \textbf{Latent Program Distribution vs.\ Normal Distribution.}
We now compare two distributions: 
the latent program distribution formed by
encoding all the training programs to the program embedding space and
a normal distribution $\mathcal{N}(0, 1)$.
One can view the normal distribution as the distribution obtained by
heavily enforcing the weight of the KL-divergence term 
when training a VAE model.
We discuss the shape of the latent program distribution 
in the learned program embedding space as follows:

\begin{itemize}
    \item \method\\-P: since \method\\+P simply
    optimizes the $\beta$-VAE loss (the program reconstruction loss $\mathcal{L}^{\text{P}}$), 
    which puts a lot of emphasis on the KL-divergence term,
    the shape of the latent program distribution 
    is very similar to a normal distribution 
    as shown in~\myfig{fig:latent vis} (a).
    \item \method\\-P+R: while \method\\+P+R additionally optimizes the program behavior reconstruction loss $\mathcal{L}^{\text{R}}$, 
    the shape of the latent program distribution 
    is still similar to a normal distribution,
    as shown in~\myfig{fig:latent vis} (b).
    We hypothesize that it is because the program behavior reconstruction loss 
    alone might not be strong or explicit enough to introduce a change.
    \item \method\\-P+L: 
    the shape of the latent program distribution 
    in the program embedding space learned by \method\\+P+L
    is significantly different from a normal distribution, 
    as shown in~\myfig{fig:latent vis} (c).
    This suggest that employing 
    the latent behavior reconstruction loss $\mathcal{L}^{\text{L}}$
    dramatically contributes to the learning.
    We believe it is because the latent behavior reconstruction loss
    is optimized with direct gradients 
    and therefore provides a stronger learning signal
    especially compared to 
    the program behavior reconstruction loss $\mathcal{L}^{\text{R}}$, 
    which is optimized using REINFORCE~\cite{williams1992simple}.
    \item \method\\ (\method\\-P+R+L): 
    \method\\ optimizes the full objective that includes
    all three proposed objectives 
    and form a similar distribution shape as the one
    learned by \method\\+P+L. 
    However, the distance between each pair of the
    ground-truth testing program 
    and the predicted program is much closer 
    in the program embedding space learned by \method\\
    compared to the space learned by \method\\+P+L.
    This justifies the effectiveness of the proposed 
    program behavior reconstruction loss $\mathcal{L}^{\text{R}}$,
    which can bring the programs with similar behaviors closer in the embedding space.
\end{itemize}

\noindent \textbf{Summary.}
The visualizations of the program embedding spaces 
learned by \method\\ and its ablations 
qualitatively justify the effectiveness of the proposed learning objectives,
as complementary to the quantitative results presented in the main paper.

\section{Cross Entropy Method Trajectory Visualization}
\label{sec:cem_vis}

\begin{figure}[t]
    \centering
    \hfill
    \begin{subfigure}[t]{0.32\textwidth}
    	\centering
    	\includegraphics[width=\textwidth]{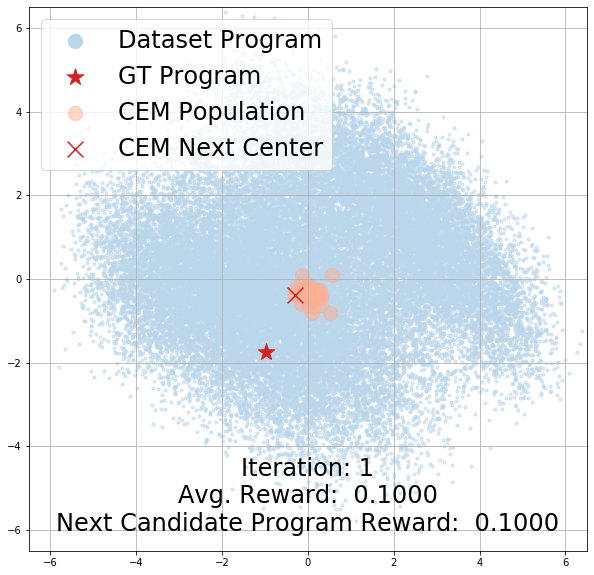}
    	\caption{Iteration 1}
    \end{subfigure}
    \hfill
    \begin{subfigure}[t]{0.32\textwidth}
    	\centering
    	\includegraphics[width=\textwidth]{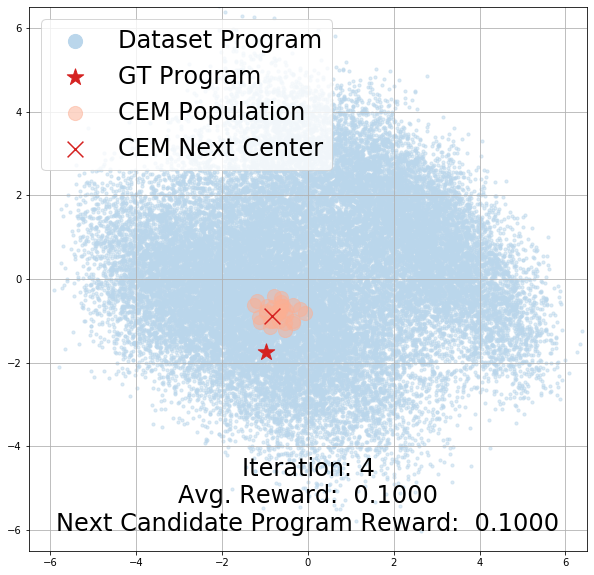}
    	\caption{Iteration 4}
    \end{subfigure}
    \hfill
    \begin{subfigure}[t]{0.32\textwidth}
    	\centering
    	\includegraphics[width=\textwidth]{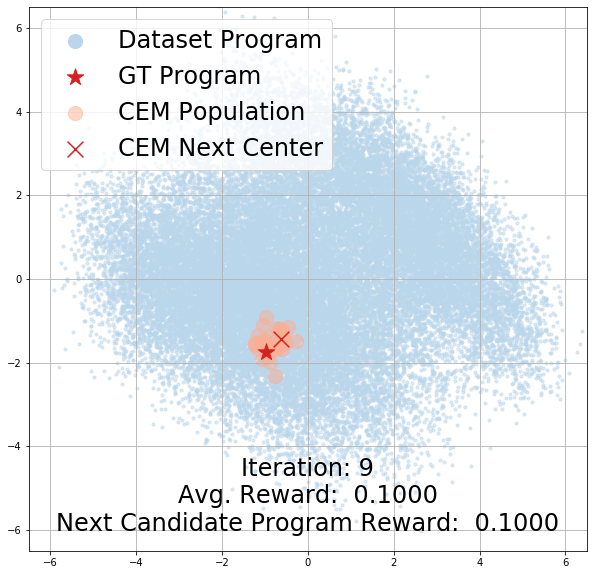}
    	\caption{Iteration 9}
    \end{subfigure}
    \hfill
    \\
    \hfill
    \begin{subfigure}[t]{0.32\textwidth}
    	\centering
    	\includegraphics[width=\textwidth]{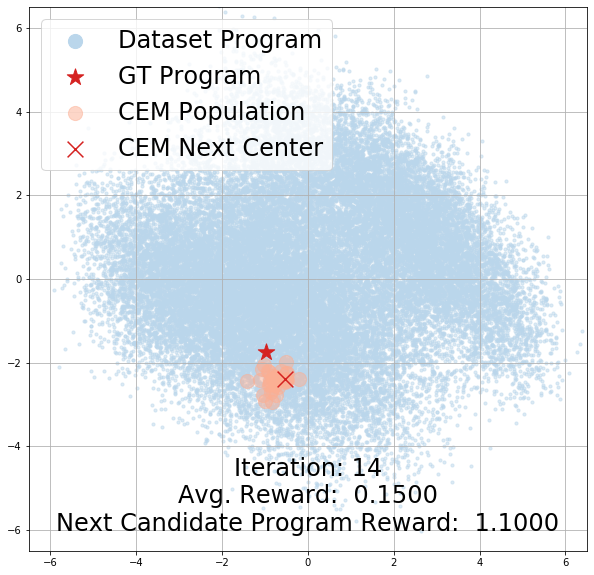}
    	\caption{Iteration 14}
    \end{subfigure}
    \hfill
    \begin{subfigure}[t]{0.32\textwidth}
    	\centering
    	\includegraphics[width=\textwidth]{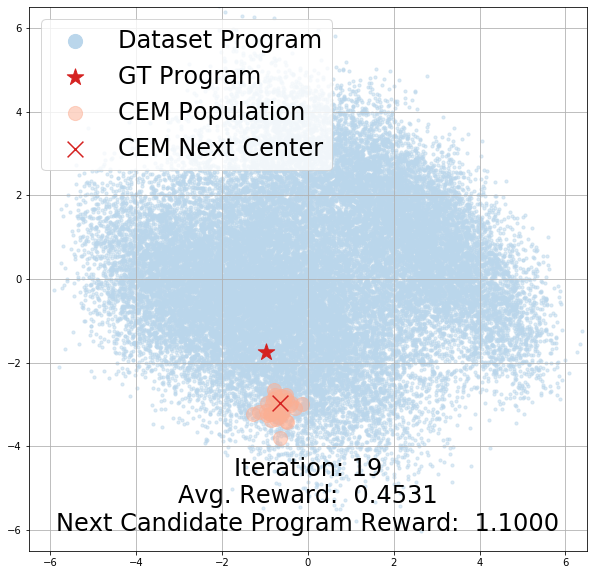}
    	\caption{Iteration 19}
    \end{subfigure}
    \hfill
    \begin{subfigure}[t]{0.32\textwidth}
    	\centering
    	\includegraphics[width=\textwidth]{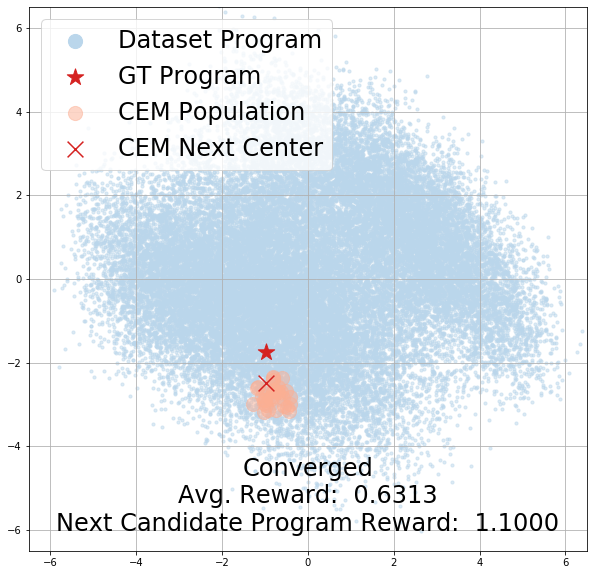}
    	\caption{Iteration 23}
    \end{subfigure}

    \caption[\textsc{StairClimber} CEM Trajectory Visualization]{
    \textbf{\textsc{StairClimber} CEM Trajectory Visualization.}
    Latent training programs from the training dataset, 
    a ground-truth program for \textsc{StairClimber} task,
    CEM populations, and CEM next candidate programs
    are embedded to a 2D space using PCA.
    Both the average reward of the entire population 
    and the reward of the next candidate program 
    (CEM Next Center) consistently increase 
    as the number of iterations increase.
    Also, the CEM population gradually 
    moves toward where the ground-truth program is located.
    }
    \label{fig:cem vis stairclimber}
\end{figure}

\begin{figure}[t]
    \centering
    \hfill
    \begin{subfigure}[t]{0.32\textwidth}
    	\centering
    	\includegraphics[width=\textwidth]{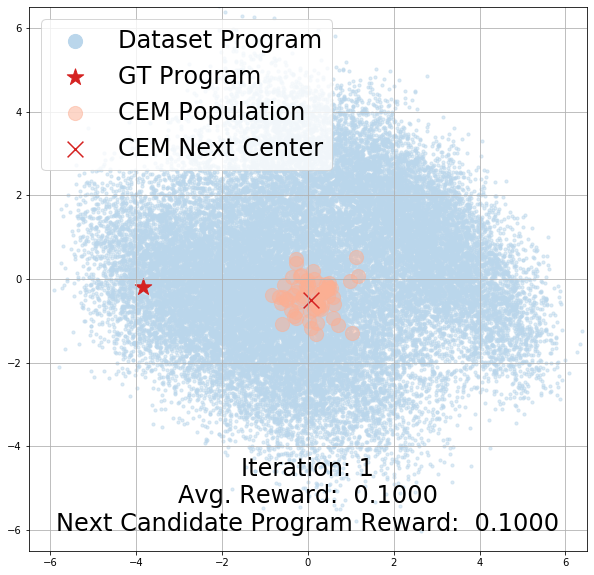}
    	\caption{Iteration 1}
    \end{subfigure}
    \hfill
    \begin{subfigure}[t]{0.32\textwidth}
    	\centering
    	\includegraphics[width=\textwidth]{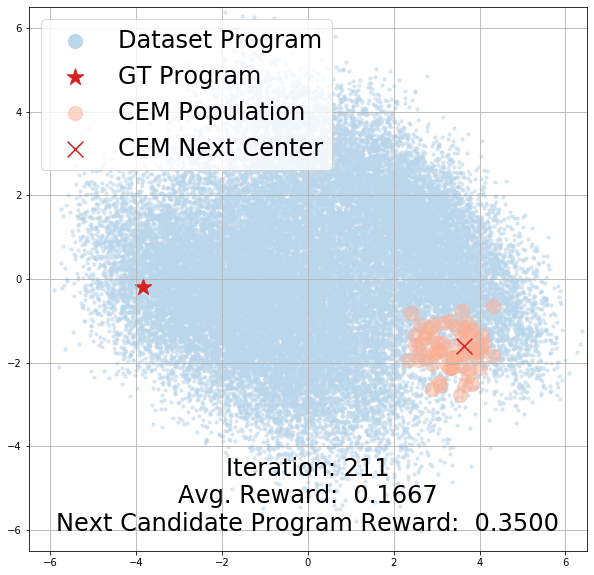}
    	\caption{Iteration 211}
    \end{subfigure}
    \hfill
    \begin{subfigure}[t]{0.32\textwidth}
    	\centering
    	\includegraphics[width=\textwidth]{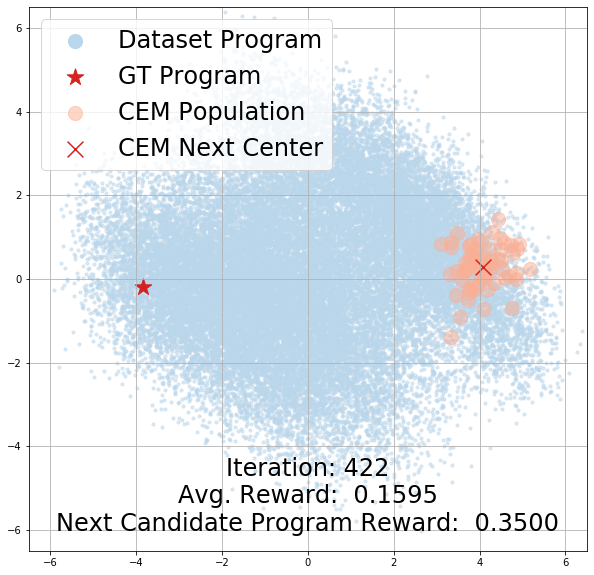}
    	\caption{Iteration 422}
    \end{subfigure}
    \hfill
    \\
    \hfill
    \begin{subfigure}[t]{0.32\textwidth}
    	\centering
    	\includegraphics[width=\textwidth]{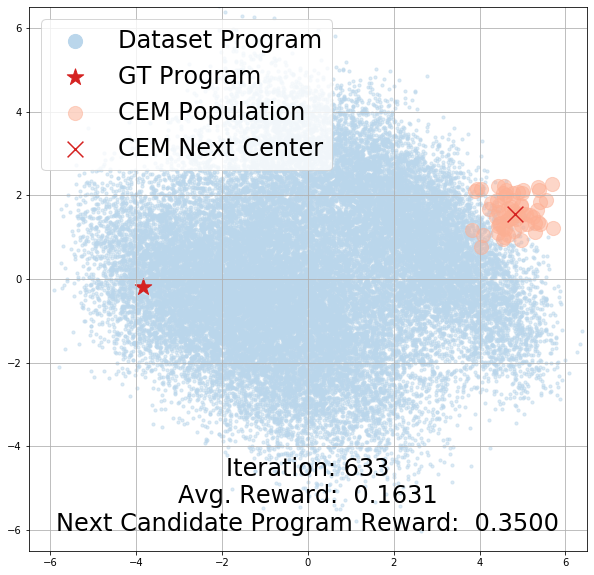}
    	\caption{Iteration 633}
    \end{subfigure}
    \hfill
    \begin{subfigure}[t]{0.32\textwidth}
    	\centering
    	\includegraphics[width=\textwidth]{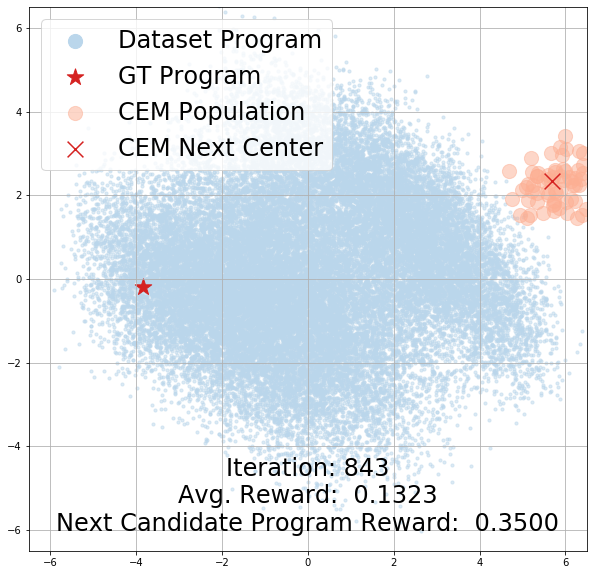}
    	\caption{Iteration 843}
    \end{subfigure}
    \hfill
    \begin{subfigure}[t]{0.32\textwidth}
    	\centering
    	\includegraphics[width=\textwidth]{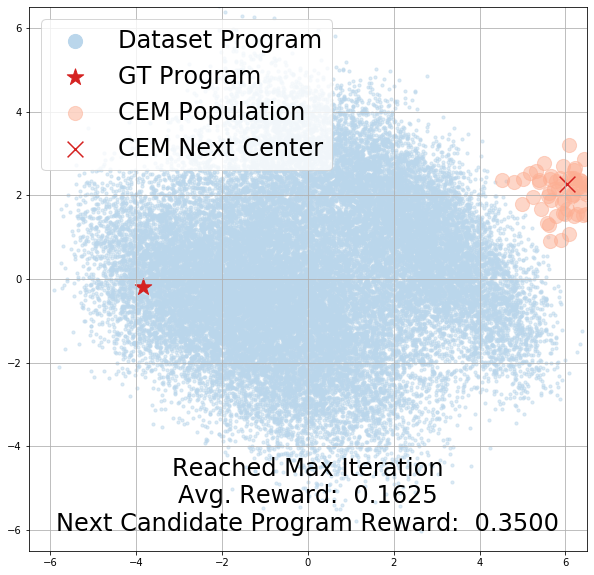}
    	\caption{Iteration 1000}
    \end{subfigure}

    \caption[\textsc{FourCorner} CEM Trajectory Visualization]{
    \textbf{\textsc{FourCorner} CEM Trajectory Visualization.}
    Latent training programs from the training dataset, 
    a ground-truth program for the \textsc{FourCorner} task,
    CEM populations, and CEM next candidate programs
    are embedded to a 2D space using PCA.
    The CEM trajectory does not converge.
    The ground-truth program lies far away from the initial sampled distribution, 
    which might contribute to the difficulty of converging.
    }
    \label{fig:cem vis fourcorner}
\end{figure}

As described in the main paper,
once the program embedding space is learned by \method\\,
our goal becomes searching for a latent program 
that maximizes the reward described by a given task MDP.
To this end, 
we adapt the Cross Entropy Method (CEM)
\citep{rubinstein1997optimization}, 
a gradient-free continuous search algorithm, 
to iteratively search over the program embedding space.
Specifically, 
we iteratively perform the following steps:

\begin{enumerate}
    \item Sample a distribution of candidate latent programs.
    \item Decode the sampled latent programs into programs using the learned program decoder $p_\theta$.
    \item Execute the programs in the task environment and obtain the corresponding rewards.
    \item Update the CEM sampling distribution based on the rewards.
\end{enumerate}

This process is repeated until either convergence or 
the maximum number of sampling steps has been reached.

We perform dimensionality reduction with PCA~\cite{pca}
to embed the following data to a 2D space;
the visualizations of CEM trajectories are shown 
in~\myfig{fig:cem vis stairclimber} and~\myfig{fig:cem vis fourcorner}:

\begin{itemize}
    \item Latent programs from the training dataset encoded by a learned encoder $q_\phi$, 
    visualized as blue scatters. There are 35k training programs.
    This is to visualize the shape of the program distribution in the learned program embedding space.
    This is also visualized 
    in~\myfig{fig:latent vis}.
    \item Ground-truth (GT) programs 
    that exhibit optimal behaviors for solving the Karel tasks, 
    visualized as red stars ($\star	$).
    Ideally, the CEM population should 
    iteratively move toward where
    the GT programs are located.
    \item CEM population is a batch of sampled candidate latent programs at each iteration,
    visualized as red scatters.
    Each candidate latent program can be decoded as a program
    that can be executed in the task environment to obtain a reward.
    By averaging the reward obtained by every candidate latent program,
    we can calculate the average reward of this population and show it in the figures as Avg. Reward.
    \item CEM Next Center,
    visualized as cross signs ($\times$),
    indicates the center vector around which
    the next batch of candidate latent programs will be sampled.
    This vector is calculated based on 
    a set of candidate latent programs 
    that achieve best reward (\ie elite samples)
    at each iteration.
    In this case, it is a weighted average based on the reward each candidate gets from its execution.
\end{itemize}

From~\myfig{fig:cem vis stairclimber}, 
we observe that both the average reward of 
the entire population 
and the reward of the next candidate program (CEM Next Center) consistently increase as the number of iterations increases,
justifying the effectiveness of CEM.
Moreover, we observe that the CEM population gradually 
moves toward where the ground-truth program is located,
which aligns well with the fact that
our proposed framework can reliably synthesize task-solving programs.

Yet, the populations might not always exactly converge to where the ground-truth latent program is. We hypothesize this could be attributed to the following reasons:

\begin{enumerate}
    \item 
    CEM convergence: while the CEM search converges, it can still be suboptimal. Since the search terminates when the next candidate latent program obtains the maximum reward (1.1 as shown in the figure) for 10 iterations, it might not exactly converge to where a ground-truth program is.
    \item
    Dimensionality reduction: we visualized the trajectories and programs by performing dimensionality reduction from 256 to 2 dimensions with PCA, which could cause visual distortions.
    \item
    Suboptimal learned program embedding space: while we aim to learn a program embedding space where all the programs inducing the same behaviors are mapped to the same spot in the embedding space, it is still possible that programs that induce the desired behavior can distribute to more than one location in a learned program embedding space. Therefore, CEM search can converge to somewhere that is different from the ground-truth latent program.
\end{enumerate}

On the other hand, the CEM trajectory shown in~\myfig{fig:cem vis fourcorner} does not converge and terminates when reaching the maximum number of iterations.
The ground-truth program lies far away from the initial sampled distribution, 
which might contribute to the difficulty of converging.
This aligns with the relatively unsatisfactory performance achieved by \method\\.
Employing a more sophisticated searching algorithm 
or conducting a more thorough hyperparameter search could potentially improve the performance
but it is not the main focus of this work.
\section{Program Embedding Space Interpolations}
\label{sec:interpolation}
\begin{table}[]
    \centering
    \small
    \caption[\method\\ Close Latent Program Interpolation]{Decoded linear interpolations of programs close to each other in the latent space.} 
    \begin{tabular}{cp{0.7\textwidth}}\toprule
        Latent Program & Decoded Program\\
        \cmidrule{1-2}
         \texttt{START} & {\begin{lstlisting}[frame=none, belowskip=-6.5pt, aboveskip=-6.5pt]
DEF run m( turnRight move WHILE c( frontIsClear c) w( move w) WHILE c( not c( frontIsClear c) c) w( move w) IF c( frontIsClear c) i( move i) m)
         \end{lstlisting}}\\
         \texttt{1} & {\begin{lstlisting}[frame=none, belowskip=-6.5pt, aboveskip=-6.5pt]
DEF run m( turnRight move WHILE c( frontIsClear c) w( move w) WHILE c( not c( frontIsClear c) c) w( move w) IF c( frontIsClear c) i( move i) m)
         \end{lstlisting}}\\
         \texttt{2} & {\begin{lstlisting}[frame=none, belowskip=-6.5pt, aboveskip=-6.5pt]
 DEF run m( turnRight move WHILE c( frontIsClear c) w( move w) IF c( not c( frontIsClear c) c) i( move i) m)
         \end{lstlisting}}\\
         \texttt{3} & {\begin{lstlisting}[frame=none, belowskip=-6.5pt, aboveskip=-6.5pt]
DEF run m( turnRight move WHILE c( frontIsClear c) w( move w) IF c( not c( frontIsClear c) c) i( move i) m)
         \end{lstlisting}}\\
         \texttt{4} & {\begin{lstlisting}[frame=none, belowskip=-6.5pt, aboveskip=-6.5pt]
DEF run m( turnRight move WHILE c( frontIsClear c) w( move w) IF c( not c( frontIsClear c) c) i( move i) m)
         \end{lstlisting}}\\
         \texttt{5} & {\begin{lstlisting}[frame=none, belowskip=-6.5pt, aboveskip=-6.5pt]
DEF run m( turnRight move WHILE c( frontIsClear c) w( move w) IF c( not c( frontIsClear c) c) i( move i) m)
         \end{lstlisting}}\\
         \texttt{6} & {\begin{lstlisting}[frame=none, belowskip=-6.5pt, aboveskip=-6.5pt]
DEF run m( turnRight move WHILE c( frontIsClear c) w( move w) IF c( not c( frontIsClear c) c) i( move i) m)
         \end{lstlisting}}\\
         \texttt{7} & {\begin{lstlisting}[frame=none, belowskip=-6.5pt, aboveskip=-6.5pt]
DEF run m( turnRight move turnLeft WHILE c( frontIsClear c) w( move w) IF c( not c( frontIsClear c) c) i( putMarker i) m)
         \end{lstlisting}}\\
         \texttt{8} & {\begin{lstlisting}[frame=none, belowskip=-6.5pt, aboveskip=-6.5pt]
DEF run m( turnRight move turnLeft WHILE c( frontIsClear c) w( move w) IF c( not c( frontIsClear c) c) i( putMarker i) m)
         \end{lstlisting}}\\
         \texttt{END} & {\begin{lstlisting}[frame=none, belowskip=-6.5pt, aboveskip=-6.5pt]
DEF run m( turnRight move turnLeft WHILE c( frontIsClear c) w( move w) IF c( not c( frontIsClear c) c) i( putMarker i) m)
         \end{lstlisting}}\\
    \bottomrule
    \end{tabular}
    \label{tab:program interpolation close}
\end{table}

\begin{table}[]
    \centering
    \small
    \caption[\method\\ Far Latent Program Interpolation]{Decoded linear interpolations of programs far from each other in the latent space.} 
    \begin{tabular}{cp{0.7\textwidth}}\toprule
        Latent Program & Decoded Program\\
        \cmidrule{1-2}
         \texttt{START} & {\begin{lstlisting}[frame=none, belowskip=-6.5pt, aboveskip=-6.5pt]
DEF run m( turnRight turnLeft turnLeft move turnRight putMarker move m)
         \end{lstlisting}}\\
         \texttt{1} & {\begin{lstlisting}[frame=none, belowskip=-6.5pt, aboveskip=-6.5pt]
DEF run m( turnRight turnLeft turnLeft move turnRight putMarker move m)
         \end{lstlisting}}\\
         \texttt{2} & {\begin{lstlisting}[frame=none, belowskip=-6.5pt, aboveskip=-6.5pt]
DEF run m( turnRight turnLeft turnLeft move WHILE c( frontIsClear c) w( putMarker w) turnRight move m)
         \end{lstlisting}}\\
         \texttt{3} & {\begin{lstlisting}[frame=none, belowskip=-6.5pt, aboveskip=-6.5pt]
DEF run m( turnRight turnLeft move turnLeft WHILE c( frontIsClear c) w( putMarker w) move m)
         \end{lstlisting}}\\
         \texttt{4} & {\begin{lstlisting}[frame=none, belowskip=-6.5pt, aboveskip=-6.5pt]
DEF run m( turnRight turnLeft move WHILE c( frontIsClear c) w( turnLeft w) IF c( not c( frontIsClear c) c) i( move i) m)
         \end{lstlisting}}\\
         \texttt{5} & {\begin{lstlisting}[frame=none, belowskip=-6.5pt, aboveskip=-6.5pt]
DEF run m( turnRight move turnLeft WHILE c( frontIsClear c) w( move w) IF c( not c( frontIsClear c) c) i( putMarker i) m)
         \end{lstlisting}}\\
         \texttt{6} & {\begin{lstlisting}[frame=none, belowskip=-6.5pt, aboveskip=-6.5pt]
DEF run m( move turnRight turnLeft move WHILE c( frontIsClear c) w( IF c( not c( rightIsClear c) c) i( putMarker i) w) m)
         \end{lstlisting}}\\
         \texttt{7} & {\begin{lstlisting}[frame=none, belowskip=-6.5pt, aboveskip=-6.5pt]
DEF run m( move turnRight turnLeft move WHILE c( frontIsClear c) w( IF c( not c( rightIsClear c) c) i( turnLeft i) w) m)
         \end{lstlisting}}\\
         \texttt{8} & {\begin{lstlisting}[frame=none, belowskip=-6.5pt, aboveskip=-6.5pt]
DEF run m( move turnRight move WHILE c( frontIsClear c) w( IF c( not c( rightIsClear c) c) i( turnLeft i) w) m)
         \end{lstlisting}}\\
         \texttt{END} & {\begin{lstlisting}[frame=none, belowskip=-6.5pt, aboveskip=-6.5pt]
DEF run m( move turnRight move WHILE c( frontIsClear c) w( IF c( not c( rightIsClear c) c) i( turnLeft i) w) m)
         \end{lstlisting}}\\
    \bottomrule
    \end{tabular}
    \label{tab:program interpolation far}
\end{table}
To learn a program embedding space that allows for smooth interpolation, 
we propose three sources of supervision.
We aim to verify the effectiveness of it by investigating interpolations
in the learned program embedding space.
To this end, we follow the procedure described below to produce results shown in \mytable{tab:program interpolation close} and \mytable{tab:program interpolation far}.

\begin{enumerate}
    \item Sampling a pair of programs from the dataset (\texttt{START} program and \texttt{END} program).
    \item Encoding the two programs into the learned program embedding space.
    \item Linearly interpolating between the two latent programs to obtain a number of (eight) interpolated latent programs.
    \item Decoding the latent programs to obtain interpolated programs (program \texttt{1} to program \texttt{8}).
\end{enumerate}

We show two pairs of programs and their interpolations in between below as examples. 
Specifically, the first pair of programs, shown in \mytable{tab:program interpolation close}, are closer to each other in the latent space and the second pair of programs, shown in \mytable{tab:program interpolation far}, are further from each other. 
We observe that the interpolations between the closer program pair exhibit smoother transitions 
and the interpolations between the further program pair display more dramatic change.
\section{Program Evolution}
\label{sec:evolution}
\begin{table}[]
    \centering
    \small
    \caption[Program Evolution Over CEM Search]{How predicted programs evolve throughout the course of CEM search for \textsc{StairClimber}. See \myfig{fig:cem vis stairclimber} for the corresponding visualization of this CEM search.} 
    \begin{tabular}{cp{0.75\textwidth}}\toprule
        Search Iteration & Best Predicted Program\\
        \cmidrule{1-2}
         \texttt{Iteration: 1} & {\begin{lstlisting}[frame=none, belowskip=-6.5pt, aboveskip=-6.5pt]
DEF run m( IF c( frontIsClear c) i( pickMarker i) WHILE c( leftIsClear c) w( move w) IFELSE c( frontIsClear c) i( turnRight move i) ELSE e( move e) m)
         \end{lstlisting}}\\
         \texttt{Iteration: 2} & {\begin{lstlisting}[frame=none, belowskip=-6.5pt, aboveskip=-6.5pt]
DEF run m( WHILE c( markersPresent c) w( move w) IFELSE c( frontIsClear c) i( turnLeft i) ELSE e( move e) WHILE c( leftIsClear c) w( move w) m)
         \end{lstlisting}}\\
         \texttt{Iteration: 3} & {\begin{lstlisting}[frame=none, belowskip=-6.5pt, aboveskip=-6.5pt]
DEF run m( WHILE c( not c( frontIsClear c) c) w( move turnRight w) WHILE c( leftIsClear c) w( turnLeft move w) m)
         \end{lstlisting}}\\
         \texttt{Iteration: 4} & {\begin{lstlisting}[frame=none, belowskip=-6.5pt, aboveskip=-6.5pt]
DEF run m( WHILE c( not c( frontIsClear c) c) w( pickMarker move w) WHILE c( leftIsClear c) w( turnLeft move w) m)
         \end{lstlisting}}\\
         \texttt{Iteration: 5} & {\begin{lstlisting}[frame=none, belowskip=-6.5pt, aboveskip=-6.5pt]
DEF run m( WHILE c( not c( frontIsClear c) c) w( pickMarker turnRight w) WHILE c( leftIsClear c) w( move turnLeft w) m)
         \end{lstlisting}}\\
         \texttt{Iteration: 6} & {\begin{lstlisting}[frame=none, belowskip=-6.5pt, aboveskip=-6.5pt]
DEF run m( WHILE c( not c( frontIsClear c) c) w( pickMarker turnRight w) WHILE c( leftIsClear c) w( move turnLeft w) m)
         \end{lstlisting}}\\
         \texttt{Iteration: 7} & {\begin{lstlisting}[frame=none, belowskip=-6.5pt, aboveskip=-6.5pt]
DEF run m( WHILE c( not c( leftIsClear c) c) w( turnRight w) IFELSE c( frontIsClear c) i( move i) ELSE e( turnLeft e) WHILE c( rightIsClear c) w( move w) m)
         \end{lstlisting}}\\
         \texttt{Iteration: 8} & {\begin{lstlisting}[frame=none, belowskip=-6.5pt, aboveskip=-6.5pt]
DEF run m( WHILE c( not c( leftIsClear c) c) w( turnRight move w) WHILE c( markersPresent c) w( turnLeft move w) m)
         \end{lstlisting}}\\
         \texttt{Iteration: 9} & {\begin{lstlisting}[frame=none, belowskip=-6.5pt, aboveskip=-6.5pt]
DEF run m( WHILE c( not c( noMarkersPresent c) c) w( turnRight move w) WHILE c( not c( frontIsClear c) c) w( turnLeft move w) m)
         \end{lstlisting}}\\
         \texttt{Iteration: 10} & {\begin{lstlisting}[frame=none, belowskip=-6.5pt, aboveskip=-6.5pt]
DEF run m( WHILE c( not c( noMarkersPresent c) c) w( turnRight move w) WHILE c( leftIsClear c) w( turnLeft move w) m)
         \end{lstlisting}}\\
         \texttt{Iteration: 11} & {\begin{lstlisting}[frame=none, belowskip=-6.5pt, aboveskip=-6.5pt]
DEF run m( WHILE c( not c( leftIsClear c) c) w( turnRight move w) WHILE c( noMarkersPresent c) w( turnLeft move w) m)
         \end{lstlisting}}\\   
         \texttt{Iteration: 12} & {\begin{lstlisting}[frame=none, belowskip=-6.5pt, aboveskip=-6.5pt]
DEF run m( WHILE c( not c( leftIsClear c) c) w( turnRight move w) WHILE c( noMarkersPresent c) w( turnLeft move w) m)
         \end{lstlisting}}\\   
         \texttt{Iteration: 13} & {\begin{lstlisting}[frame=none, belowskip=-6.5pt, aboveskip=-6.5pt]
DEF run m( WHILE c( not c( leftIsClear c) c) w( turnRight move w) WHILE c( noMarkersPresent c) w( turnLeft move w) m)
         \end{lstlisting}}\\   
         \texttt{Iteration: 14} & {\begin{lstlisting}[frame=none, belowskip=-6.5pt, aboveskip=-6.5pt]
DEF run m( WHILE c( not c( markersPresent c) c) w( turnRight move w) WHILE c( rightIsClear c) w( move turnLeft w) m)
         \end{lstlisting}}\\            
         \texttt{Iteration: 15} & {\begin{lstlisting}[frame=none, belowskip=-6.5pt, aboveskip=-6.5pt]
DEF run m( WHILE c( not c( markersPresent c) c) w( turnRight move w) WHILE c( rightIsClear c) w( move turnLeft w) m)
         \end{lstlisting}}\\   
         \texttt{Iteration: 16} & {\begin{lstlisting}[frame=none, belowskip=-6.5pt, aboveskip=-6.5pt]
DEF run m( WHILE c( not c( markersPresent c) c) w( turnRight move w) WHILE c( rightIsClear c) w( move turnLeft w) m)
         \end{lstlisting}}\\   
         \texttt{Iteration: 17} & {\begin{lstlisting}[frame=none, belowskip=-6.5pt, aboveskip=-6.5pt]
DEF run m( WHILE c( not c( markersPresent c) c) w( turnRight move w) WHILE c( rightIsClear c) w( move turnLeft w) m)
         \end{lstlisting}}\\   
         \texttt{Iteration: 18} & {\begin{lstlisting}[frame=none, belowskip=-6.5pt, aboveskip=-6.5pt]
DEF run m( WHILE c( not c( markersPresent c) c) w( turnRight move w) WHILE c( rightIsClear c) w( move turnLeft w) m)
         \end{lstlisting}}\\   
         \texttt{Iteration: 19} & {\begin{lstlisting}[frame=none, belowskip=-6.5pt, aboveskip=-6.5pt]
DEF run m( WHILE c( not c( markersPresent c) c) w( turnRight move w) WHILE c( rightIsClear c) w( move turnLeft w) m)
         \end{lstlisting}}\\   
         \texttt{Iteration: 20} & {\begin{lstlisting}[frame=none, belowskip=-6.5pt, aboveskip=-6.5pt]
DEF run m( WHILE c( not c( markersPresent c) c) w( turnRight move w) WHILE c( rightIsClear c) w( move turnLeft w) m)
         \end{lstlisting}}\\   
         \texttt{Iteration: 21} & {\begin{lstlisting}[frame=none, belowskip=-6.5pt, aboveskip=-6.5pt]
DEF run m( WHILE c( not c( markersPresent c) c) w( turnRight move w) WHILE c( rightIsClear c) w( move turnLeft w) m)
         \end{lstlisting}}\\   
         \texttt{Iteration: 22} & {\begin{lstlisting}[frame=none, belowskip=-6.5pt, aboveskip=-6.5pt]
DEF run m( WHILE c( not c( markersPresent c) c) w( turnRight move w) WHILE c( rightIsClear c) w( move turnLeft w) m)
         \end{lstlisting}}\\  
         \texttt{Converged} & {\begin{lstlisting}[frame=none, belowskip=-6.5pt, aboveskip=-6.5pt]
DEF run m( WHILE c( not c( markersPresent c) c) w( turnRight move w) WHILE c( rightIsClear c) w( turnLeft move w) m)
         \end{lstlisting}}\\  
    \bottomrule
    \end{tabular}
    \label{tab:program evolution}
\end{table}

In this section, we aim to investigate how predicted programs evolve over the course of searching. 
We visualize converged CEM search trajectories and the reward each program gets on the StairClimber task in Appendix \myfig{fig:cem vis stairclimber}. 
In~\mytable{tab:program evolution}, we present the predicted programs corresponding to 
the CEM search trajectory on the \textsc{StairClimber} task in~\myfig{fig:cem vis stairclimber}.
We observe that the sampled programs consistently improve as the number of iterations increases, 
justifying the effectiveness of the learned program embedding and the CEM search.

\section{Interpretability: Human Debugging of \method\\ Programs}
\label{sec:debug}
\begin{table}

    \caption[Human Debugging Experiment Results]{Mean return (standard deviation) [\% increase in performance]
    after debugging by non-expert humans of \method\\ synthesized programs
    for 3 statement edits and 5 statement edits. Chosen \method\\ programs
    are median-reward programs out of 5 \method\\ seeds for each task.
    }
    
    \label{tab:interpretability results}
    \centering
    
\scalebox{0.9}{
    \begin{tabular}{cccc}
    \toprule
    Karel Task & Original Program & 3 Edits & 5 Edits \\
    \midrule
    \textsc{TopOff} & 0.86 & 0.95 (0.07) [10.5\%] & 1.0 (0.00) [16.3\%] \\
    \textsc{FourCorner} & 0.25 & 0.75 (0.35) [200\%] & 0.92 (0.12) (268\%) \\
    \textsc{Harvester} & 0.47 & 0.85 (0.05) [80.9\%] & 0.89 (0.00) [89.4\%]\\
    \midrule 
    Average \% Increase & - & 97.1\% & 125\% \\
    \bottomrule
    \end{tabular}}
\end{table}

\begin{figure}
    \begin{mdframed}[]
    \centering
    \includegraphics[width=0.6\textwidth]{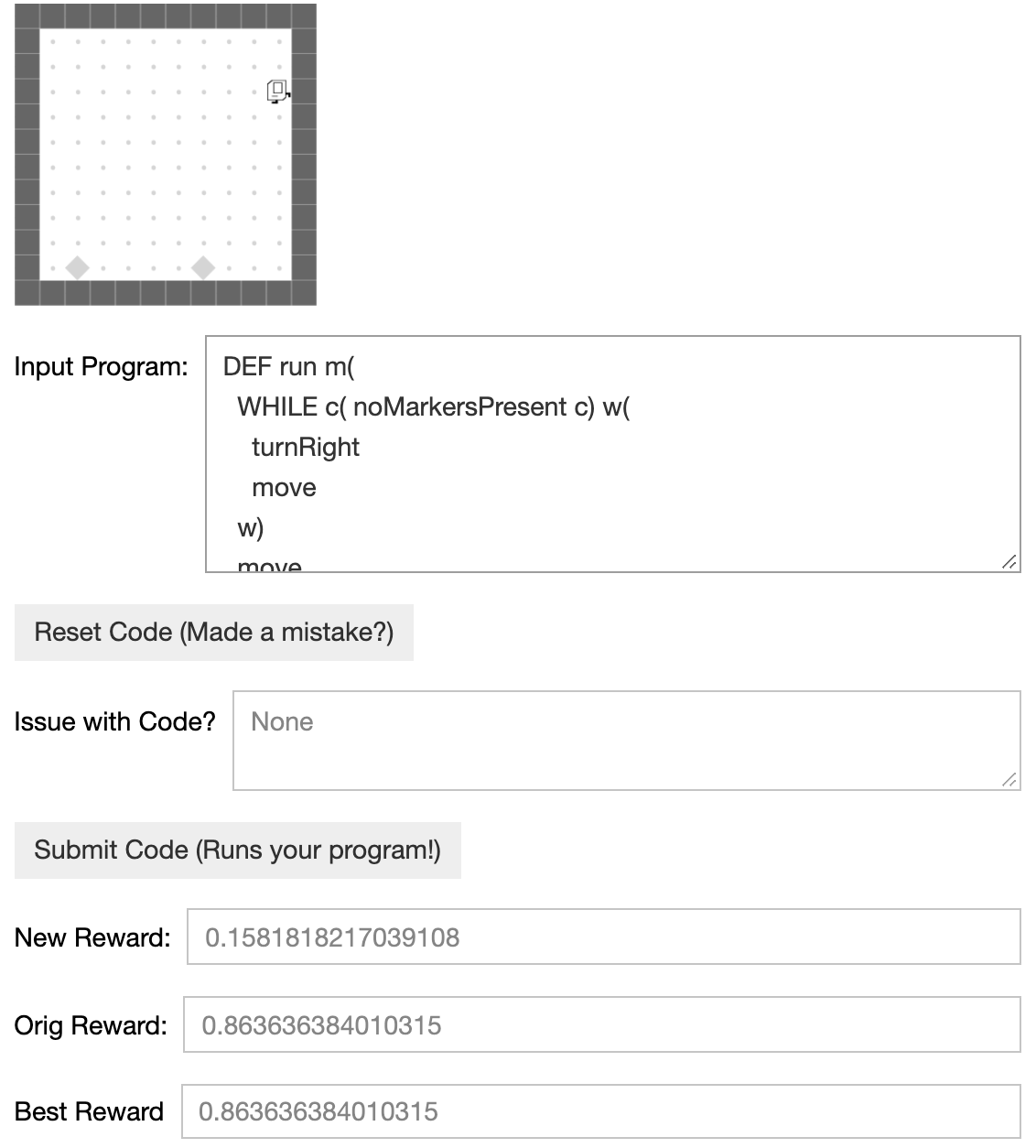}
    \end{mdframed}
    \caption[Human Debugging Experiment User Interface]{\textbf{User Interface for the Human Debugging Interpretability Experiments.}
            The top contains moving rollout visualizations of the current program in the ``Input Program'' box, which 
            users are allowed to edit. ``Input Program'' will first contain the program synthesized by \method\\.
            Syntax errors or other issues with code (such as the edit distance being too high) are displayed in the ``Issue with Code?'' box, the reward of the 
            current inputted program is in the ``New Reward'' box, and the reward of the original program synthesized
            by \method\\ is in the ``Orig Reward'' box. The user's best reward across all inputted programs is kept track of in the ``Best Reward'' box. }
    \label{fig:interpretability UI}
\end{figure}
\begin{figure}
    \ContinuedFloat
    \centering
    \begin{mdframed}[frametitle=\textsc{TopOff}]
    \begin{subfigure}[t]{0.32\textwidth}
    \textsc{\method\\ (Reward=0.86)}
    {\begin{lstlisting}
DEF run m( 
  WHILE c( noMarkersPresent c) w( 
    turnRight 
    move 
  w) 
  putMarker 
  move 
  WHILE c( noMarkersPresent c) w( 
    turnRight 
    move 
  w) 
  putMarker 
  move 
  WHILE c( noMarkersPresent c) w( 
    turnRight 
    move 
  w) 
  putMarker 
  move 
  WHILE c( noMarkersPresent c) w( 
    turnRight 
    move 
  w) 
  putMarker 
  move 
m) 
    \end{lstlisting}}
    \end{subfigure}
    \begin{subfigure}[t]{0.32\textwidth}
    \textsc{3 Edits (Reward=1.0)}
    {\begin{lstlisting}
DEF run m(
  REPEAT R=9 r( 
  WHILE c( noMarkersPresent c) w( 
    IF c( frontIsClear c) i( move i)
  w) 
  putMarker 
  move
  r) 
  WHILE c( noMarkersPresent c) w( 
    turnRight 
    move 
  w) 
  putMarker 
  move 
  WHILE c( noMarkersPresent c) w( 
    turnRight 
    move 
  w) 
  putMarker 
  move 
  WHILE c( noMarkersPresent c) w( 
    turnRight 
    move 
  w) 
  putMarker 
  move 
m) 
    \end{lstlisting}}
    
    \end{subfigure}
    \begin{subfigure}[t]{0.32\textwidth}
    \textsc{5 Edits (Reward=1.0)}
    {\begin{lstlisting}
DEF run m(
  WHILE c( frontIsClear c) w( 
    IF c( markersPresent c) i(
      putMarker
    i)
    move 
  w)
  WHILE c( frontIsClear c) w(
  WHILE c( noMarkersPresent c) w( 
    turnRight 
    move 
  w) 
  putMarker 
  move 
  WHILE c( noMarkersPresent c) w( 
    turnRight 
    move 
  w) 
  putMarker 
  move 
  WHILE c( noMarkersPresent c) w( 
    turnRight 
    move 
  w) 
  putMarker 
  move 
  WHILE c( noMarkersPresent c) w( 
    turnRight 
    move 
  w) 
  putMarker 
  move
  w)
m) 
    \end{lstlisting}}
    \end{subfigure}
    \end{mdframed}
    \begin{mdframed}[frametitle=\textsc{FourCorner}]
    \begin{subfigure}[t]{0.32\textwidth}
    \textsc{\method\\ (Reward=0.25)}
    {\begin{lstlisting}
DEF run m( 
  turnRight 
  turnRight 
  move 
  turnRight 
  turnRight 
  turnRight 
  turnRight 
  turnRight 
  turnRight 
  turnRight 
  turnRight 
  turnRight 
  turnRight 
  turnRight 
  turnRight 
  turnRight 
  turnRight 
  turnRight 
  turnRight 
  turnRight 
  turnRight 
  turnRight 
  turnRight 
  turnRight 
  turnRight 
  turnRight 
  turnRight 
  turnRight 
  turnRight 
  turnRight 
  turnRight 
  turnRight 
  turnRight 
  putMarker 
  turnRight 
  turnRight 
m) 
    \end{lstlisting}}
    \end{subfigure}
    \begin{subfigure}[t]{0.32\textwidth}
    \textsc{3 Edits (Reward=1.0)}
    {\begin{lstlisting}
DEF run m( 
  turnRight 
  turnRight 
  move 
  turnRight 
  turnRight 
  turnRight 
  turnRight 
  turnRight 
  turnRight 
  turnRight 
  turnRight 
  turnRight 
  turnRight 
  turnRight 
  turnRight 
  turnRight 
  turnRight 
  turnRight 
  turnRight 
  turnRight 
  turnRight 
  turnRight 
  turnRight 
  turnRight 
  turnRight 
  turnRight 
  turnRight 
  turnRight 
  turnRight 
  turnRight 
  turnRight 
  turnRight
  REPEAT R=4 r(
  REPEAT R=9 r(
  move
  r)
  putMarker
  turnRight
  r)
  turnRight
m)
    \end{lstlisting}}
    
    \end{subfigure}
    \begin{subfigure}[t]{0.32\textwidth}
    \textsc{5 Edits (Reward=1.0)}
    {\begin{lstlisting}
DEF run m( 
  turnRight 
  turnRight 
  move 
  turnRight 
  turnRight 
  turnRight 
  turnRight 
  turnRight 
  turnRight 
  turnRight 
  turnRight 
  turnRight 
  turnRight 
  turnRight 
  turnRight 
  turnRight 
  turnRight 
  turnRight 
  turnRight 
  turnRight 
  turnRight 
  turnRight 
  turnRight 
  turnRight 
  turnRight 
  turnRight 
  turnRight 
  turnRight 
  turnRight 
  turnRight 
  turnRight 
  turnRight
  putMarker
  REPEAT R=3 r(
  REPEAT R=9 r(
  move
  r)
  putMarker
  turnRight
  r)
m) 
    \end{lstlisting}}
    \end{subfigure}
    \end{mdframed} 
    \end{figure}
    \begin{figure}
    \begin{mdframed}[frametitle=\textsc{Harvester}]
    \begin{subfigure}[t]{0.32\textwidth}
    \textsc{\method\\ (Reward=0.47)}
    {\begin{lstlisting}
DEF run m( 
  turnLeft 
  turnLeft 
  pickMarker 
  move 
  pickMarker 
  move 
  pickMarker 
  move 
  pickMarker 
  move 
  pickMarker 
  move 
  turnLeft 
  pickMarker 
  move 
  pickMarker 
  move 
  pickMarker 
  move 
  pickMarker 
  move 
  turnLeft 
  pickMarker 
  move 
  pickMarker 
  move 
  pickMarker 
  move 
  turnLeft 
  pickMarker 
  move 
  pickMarker 
  move 
  pickMarker 
  move 
  turnLeft 
  pickMarker 
  move 
  pickMarker 
  move 
  pickMarker 
  move 
m) 

    \end{lstlisting}}
    \end{subfigure}
    \begin{subfigure}[t]{0.32\textwidth}
    \textsc{3 Edits (Reward=0.77)}
    {\begin{lstlisting}
DEF run m( 
  turnLeft 
  turnLeft 
  REPEAT R=4 r(
    pickMarker 
    move 
    pickMarker 
    move 
    pickMarker 
    move 
    pickMarker 
    move 
    pickMarker 
    move 
    pickMarker
    move
    turnLeft 
  r)
  pickMarker 
  move 
  pickMarker 
  move 
  pickMarker 
  move 
  pickMarker 
  move 
  pickMarker 
  move 
  turnLeft 
  pickMarker 
  move 
  pickMarker 
  move 
  pickMarker 
  move 
  turnLeft 
  pickMarker 
  move 
  pickMarker 
  move 
  pickMarker 
  move 
  turnLeft 
  pickMarker 
  move 
  pickMarker 
  move 
  pickMarker 
  move 
m) 
    \end{lstlisting}}
    
    \end{subfigure}
    \begin{subfigure}[t]{0.32\textwidth}
    \textsc{5 Edits (Reward=0.89)}
    {\begin{lstlisting}
DEF run m( 
  REPEAT R=3 r(
  pickMarker 
  move 
  pickMarker 
  move 
  pickMarker 
  move 
  pickMarker 
  move 
  pickMarker 
  move 
  turnLeft
  r)
  REPEAT R=3 r(
  pickMarker 
  move 
  pickMarker 
  move 
  pickMarker 
  move 
  pickMarker 
  move 
  turnLeft
  r)
  pickMarker 
  move 
  pickMarker 
  move 
  pickMarker 
  move 
  turnLeft 
  pickMarker 
  move 
  pickMarker 
  move 
  pickMarker 
  move 
  turnLeft 
  pickMarker 
  move 
  pickMarker 
  move 
  pickMarker 
  move 
m) 
    \end{lstlisting}}
    \end{subfigure}
    \end{mdframed}
    \caption[Human Debugging Experiment Example Programs]{\textbf{Human Debugging Experiment Example Programs.} Example original and human-edited programs for each Karel
    task for edit distances 3 and 5.}
    \label{fig:interpretability example programs}
\end{figure}
Interpretability in Machine Learning
is crucial for several reasons~\cite{lipton2018mythos, shen2020}.
First, \textit{trust} -- interpretable machine learning methods and models may more easily be trusted
since humans tend to be reluctant to trust systems that they do not understand.
Second, interpretability can improve the \textit{safety} of 
machine learning systems.
A machine learning system that is interpretable allows for
diagnosing issues (\eg the distribution shift from training data to testing data)
earlier and provides more opportunities to intervene.
This is especially important for safety-critical tasks such as medical diagnosis~\cite{bakator2018deep,shen2017deep,ghoshal2020estimating,chang2021behavioral,singh2020explainable} and real-world robotics~\cite{campeau2019kinova, gu2017deep, openai2018learning, hafner2019dream, yamada2020motion, zhang2021policy, lee2019follow} tasks.
Finally, interpretability can lead to \textit{contestability}, 
by producing a chain of reasoning, providing insights on how a decision is made and therefore allowing humans to contest unfair or improper decisions.

We believe interpretability is especially crucial when it comes to learning a policy that interacts with the environment.
In this work, we propose a framework that offers an effective way to acquire an interpretable programmatic policy structured in a program.
In the following, we discuss how the proposed framework enjoys interpretability from the three aforementioned aspects.
Programs synthesized by the proposed framework can naturally be better \textit{trusted} since one can simply read and understand them.
Also, through the program execution trace produced by executing a program, 
each decision made by the policy (\ie the program)
is traceable and therefore satisfies the \textit{contestability} property.
Finally, the programs produced by our framework satisfy the \textit{safety} property of interpretability as humans can diagnose and correct for issues by reading and editing the programs.

Our synthesized programs are not only readable to human users
but also interactable, 
allowing \textit{non-expert} users with a basic understanding of programming to diagnose and make edits to improve their performance.
To test this hypothesis, we asked people with programming experience who are 
unfamiliar with our DSL or Karel tasks to edit suboptimal \method\\ programs to improve
performance as much as possible on 3 Karel tasks: \textsc{TopOff}, \textsc{FourCorner}, and
\textsc{Harvester} through a user interface displayed in \myfig{fig:interpretability UI}. Each person was given 1.5 hours (30 minutes per program), including time required
to understand what the \method\\ programs were doing, understand the DSL tokens, and fully debug/test
their edited programs. For each program, participants were required to modify up to 5 statements, then
attempt the task again with up to only 3 modifications as calculated by the Levenshtein distance metric~\citep{Levenshtein_SPD66}.
A single statement modification is defined as any modification/removal/addition of a IF, WHILE, IFELSE, REPEAT, or ELSE statement, or a removal/addition/change of an action statement (\eg move, turnLeft, etc.).
Participants were allowed to ask clarification questions, but we would not answer questions regarding
how to specifically improve the performance of their program.

We display example edited programs in \myfig{fig:interpretability example programs}, and the aggregated results of editing in Table~\ref{tab:interpretability results}. We see a significant increase in performance 
in all three tasks, with an average 97.1\% increase in performance with 3 edits and an
average 125\% increase in performance with 5. These numbers are averaged over 3 people, with standard deviations reported in the table. Thus we see that even slight modifications to suboptimal \method\\ programs
can enable much better Karel task performance when edited by non-expert humans.

Our experiments in this section make an interesting connection to works in
program/code repair (\ie automatic bug fixing)~\cite{yasunaga20a, nguyen2013semfix, jiang2018shaping, xin2017leveraging,goues2019automated,schulte2010automated,koyuncu2020fixminer,durieux2016dynamoth,li2020dlfix,chen2017contract,white2019sorting,gupta2017deepfix,wang2017dynamic,mesbah2019deepdelta},
where the aim is to develop algorithms and models 
that can find bugs or even repair programs without the intervention of a human programmer.
While the goal of these works is to fix programs produced by humans,
our goal in this section is to allow humans to improve programs 
synthesized by the proposed framework.

Another important benefit of programmatic policies is \textit{verifiability} - the ability to verify different properties of policies such as 
correctness, stability, smoothness, robustness, safety, etc.
Since programmatic policies are highly structured, 
they are more amenable to formal verification methods developed for traditional software systems as compared to neural policies. 
Recent works \cite{bastani2018verifiable, verma2018programmatically, verma2019imitation, zhu2019inductive} show that various properties of programmatic policies (programs written using DSLs, decision trees) 
can be verified using existing verification algorithms,
which can also be applied to programs synthesized by 
the proposed framework.

\begin{figure}[H]
    \begin{subfigure}[t]{0.32\textwidth}
    \textsc{WHILE:  }
    {\begin{lstlisting} 
DEF run m( 
    WHILE c( frontIsClear c) w( 
        turnRight
        move 
        pickMarker
        turnRight
        w)
    m)
    \end{lstlisting}} 
    \end{subfigure}
    \begin{subfigure}[t]{0.32\textwidth}
    \textsc{IFELSE+WHILE: }
    {\begin{lstlisting}
DEF run m( 
    IFELSE c( markersPresent c) i( 
        move turnRight 
        i) ELSE e( 
            move 
        e) 
    move 
    move 
    WHILE c( leftIsClear c) w( 
        turnLeft 
        w) 
    m)
    \end{lstlisting}}
    \end{subfigure}
    \begin{subfigure}[t]{0.32\textwidth}
    \textsc{2IF+IFELSE: }
    {\begin{lstlisting}
DEF run m( 
    IF c( frontIsClear c) i( 
        putMarker 
    i) 
    move 
    IF c( rightIsClear c) i( 
        move 
    i) 
    IFELSE c( frontIsClear c) i( 
        move 
        i) ELSE e( 
            move
        e)
    m)
    \end{lstlisting}}
    \end{subfigure}
    \begin{subfigure}[t]{0.32\textwidth}
    \textsc{WHILE+2IF+IFELSE: }
    {\begin{lstlisting}
DEF run m( 
    WHILE c( leftIsClear c) w( 
        turnLeft 
    w) 
    IF c( frontIsClear c) i( 
        putMarker move 
    i) 
    move 
    IF c( rightIsClear c) i( 
        turnRight 
        move 
    i) 
    IFELSE c( frontIsClear c) i( 
        move 
        i) ELSE e( 
            turnLeft move 
        e) 
    m)
    \end{lstlisting}}
\end{subfigure}
    \begin{subfigure}[t]{0.32\textwidth}
     \textsc{StairClimber:} 
     {\begin{lstlisting} 
DEF run m( 
    WHILE c( noMarkersPresent c) w( 
        turnLeft 
        move 
        turnRight 
        move 
        w) 
    m)
     \end{lstlisting}}
\end{subfigure}
    \begin{subfigure}[t]{0.32\textwidth}
     \textsc{TopOff:} 
     {\begin{lstlisting} 
DEF run m( 
    WHILE c( frontIsClear c) w( 
        IF c( markersPresent c) i( 
            putMarker 
            i) 
        move 
        w) 
    m)
     \end{lstlisting}}
\end{subfigure}
    \begin{subfigure}[t]{0.32\textwidth}
     \textsc{CleanHouse:}
     {\begin{lstlisting}
DEF run m( 
    WHILE c( noMarkersPresent c) w( 
        IF c( leftIsClear c) i( 
            turnLeft 
            i) 
        move 
        IF c( markersPresent c) i( 
            pickMarker 
            i) 
        w) 
    m)
     \end{lstlisting}}
\end{subfigure}
    \begin{subfigure}[t]{0.32\textwidth}
     \textsc{FourCorner:} 
     {\begin{lstlisting}
DEF run m( 
    WHILE c( noMarkersPresent c) w( 
        WHILE c( frontIsClear c) w( 
            move 
            w) 
        IF c( noMarkersPresent c) i( 
            putMarker 
            turnLeft 
            move 
            i) 
        w) 
    m)
     \end{lstlisting}}
\end{subfigure}
    \begin{subfigure}[t]{0.32\textwidth}
     \textsc{Maze:} 
     {\begin{lstlisting}
DEF run m( 
    WHILE c( noMarkersPresent c) w( 
        IFELSE c( rightIsClear c) i( 
            turnRight 
            i) ELSE e( 
                WHILE c( not c( frontIsClear c) c) w( 
                    turnLeft 
                    w) 
            e) 
            move 
        w) 
    m)
     \end{lstlisting}}
\end{subfigure}
    \begin{subfigure}[t]{\textwidth}
     \textsc{Harvester:} 
     {\begin{lstlisting}
DEF run m( 
    WHILE c( markersPresent c) w( 
        WHILE c( markersPresent c) w( 
            pickMarker 
            move 
            w) 
        turnRight 
        move 
        turnLeft
        WHILE c( markersPresent c) w( 
            pickMarker 
            move 
            w) 
        turnLeft 
        move 
        turnRight
        w) 
    m)
     \end{lstlisting}}
\end{subfigure}
    \caption[Ground-Truth Test Programs and Karel Programs]{\textbf{Ground-Truth Test and Karel Programs.}
        Here we display ground-truth test set programs used for reconstruction experiments and 
        example ground-truth programs that we write which can solve the Karel tasks (there are an infinite number of programs that can solve each task). Conditionals are enclosed in \texttt{c( c)}, while loops are enclosed in \texttt{w( w)}, if statements are enclosed in \texttt{i( i)}, and the main program is enclosed in \texttt{DEF run m( m)}.
        \label{fig:reconstruction tasks}
    }
\end{figure}

\begin{figure*}[!h]
\centering

\begin{mdframed}[frametitle={Na\"{i}ve}]
 \dslfontsize
 \begin{subfigure}[t]{0.49\textwidth}
 \textsc{WHILE}
 {\begin{lstlisting}
DEF run m(
    WHILE c(frontIsClear c) w(
        turnRight
        move
        pickMarker
        turnRight
        w)
    m)
 \end{lstlisting}}
 \end{subfigure}
 \begin{subfigure}[t]{0.49\textwidth}
 \textsc{IFELSE+WHILE}
 {\begin{lstlisting}
DEF run m(
    move
    move
    move
    turnLeft
    turnLeft
    m)
 \end{lstlisting}}
 \end{subfigure}
 \begin{subfigure}[t]{0.49\textwidth}
 \textsc{2IF+IFELSE}
 {\begin{lstlisting}
DEF run m(
    putMarker
    move
    move
    move
    m)
 \end{lstlisting}}
 \end{subfigure}
 \begin{subfigure}[t]{0.49\textwidth}
 \textsc{WHILE+2IF+IFELSE}
 {\begin{lstlisting}
DEF run m(
    turnLeft
    putMarker
    move
    move
    WHILE c( markersPresent c) w(
        pickMarker
        pickMarker
        pickMarker
        w)
    m)
 \end{lstlisting}}
 \end{subfigure}
\end{mdframed}

\begin{mdframed}[frametitle={\method\\-P}]
\dslfontsize
\begin{subfigure}[t]{0.49\textwidth}
\textsc{WHILE} {\begin{lstlisting}
DEF run m( 
    IF c( frontIsClear c) i( 
        turnRight 
        move 
        pickMarker 
        turnRight 
        i) 
    m)
\end{lstlisting}}
\end{subfigure}
\begin{subfigure}[t]{0.49\textwidth}
\textsc{IFELSE+WHILE} 
{\begin{lstlisting}
DEF run m( 
    IFELSE c( rightIsClear c) i( 
        move 
        i) ELSE e( 
        move 
        e) 
    move 
    move 
    IF c( leftIsClear c) i( 
        turnLeft 
        i) 
    m)
\end{lstlisting}}
\end{subfigure}
\begin{subfigure}[t]{0.49\textwidth}
\textsc{2IF+IFELSE} 
{\begin{lstlisting}
DEF run m( 
    IFELSE c( not c( frontIsClear c) c) i( 
        move 
        i) ELSE e( 
        putMarker 
        move 
        e) 
    move 
    move 
    m)
\end{lstlisting}}
\end{subfigure}
\begin{subfigure}[t]{0.49\textwidth}
\textsc{WHILE+2IF+IFELSE}
{\begin{lstlisting}
DEF run m( 
    WHILE c( leftIsClear c) w( 
        turnLeft 
        w) 
    putMarker 
    move 
    move 
    turnRight 
    move 
    move 
    m)
\end{lstlisting}}
\end{subfigure}
\end{mdframed}
\begin{mdframed}[frametitle={\method\\-P+R}]
\dslfontsize
\begin{subfigure}[t]{0.49\textwidth}
\textsc{WHILE}
{\begin{lstlisting}
DEF run m( 
    WHILE c( rightIsClear c) w( 
        WHILE c( frontIsClear c) w( 
            turnRight 
            move 
            pickMarker 
            turnRight 
            w) 
        w) 
    m)
\end{lstlisting}}
\end{subfigure}
\begin{subfigure}[t]{0.49\textwidth}
\textsc{IFELSE+WHILE}
{\begin{lstlisting}
DEF run m( 
    REPEAT R=1 r( 
        move 
        r) 
    REPEAT R=2 r( 
        move 
        r) 
    m)
\end{lstlisting}}
\end{subfigure}
\begin{subfigure}[t]{0.49\textwidth}
\textsc{2IF+IFELSE}
{\begin{lstlisting}
DEF run m( 
    IFELSE c( not c( frontIsClear c) c) i( 
        move 
        i) ELSE e( 
        putMarker 
        e) 
    IFELSE c( rightIsClear c) i( 
        move 
        i) ELSE e( 
        move 
        e)
    IF c( rightIsClear c) i( 
        move 
        i) 
    move 
    m)
\end{lstlisting}}
\end{subfigure}
\begin{subfigure}[t]{0.49\textwidth}
\textsc{WHILE+2IF+IFELSE}
{\begin{lstlisting}
DEF run m( 
    WHILE c( leftIsClear c) w( 
        turnLeft 
        w) 
    putMarker 
    move 
    move 
    turnRight 
    move 
    move 
    m)
\end{lstlisting}}
\end{subfigure}
\end{mdframed}
\end{figure*}
\begin{figure*}[!h]
\ContinuedFloat
\centering
\begin{mdframed}[frametitle={\method\\-P+L}]
\dslfontsize
\begin{subfigure}[t]{0.49\textwidth}
\textsc{WHILE}
{\begin{lstlisting}
DEF run m( 
    WHILE c( frontIsClear c) w( 
        turnRight 
        move 
        pickMarker 
        turnRight 
        w) 
    m)
\end{lstlisting}}
\end{subfigure}
\begin{subfigure}[t]{0.49\textwidth}
\textsc{IFELSE+WHILE}
{\begin{lstlisting}
DEF run m( 
    move 
    move 
    move 
    WHILE c( leftIsClear c) w( 
        turnLeft 
        w) 
    m)
\end{lstlisting}}
\end{subfigure}
\begin{subfigure}[t]{0.49\textwidth}
\textsc{2IF+IFELSE}
{\begin{lstlisting}
DEF run m( 
    IFELSE c( frontIsClear c) i( 
        REPEAT R=0 r( 
            turnRight 
            r) 
        putMarker 
        move 
        i) ELSE e( 
        move 
        e) 
    move 
    move 
    m)
\end{lstlisting}}
\end{subfigure}
\begin{subfigure}[t]{0.49\textwidth}
\textsc{WHILE+2IF+IFELSE}
{\begin{lstlisting}
DEF run m( 
    WHILE c( leftIsClear c) w( 
        turnLeft 
        w) 
    WHILE c( leftIsClear c) w( 
        turnLeft 
        w) 
    WHILE c( leftIsClear c) w( 
        turnLeft 
        w)
    WHILE c( leftIsClear c) w( 
        turnLeft 
        w) 
    IF c( frontIsClear c) i( 
        putMarker 
        move 
        i) 
    move 
    move 
    m)
\end{lstlisting}}
\end{subfigure}
\end{mdframed}
\begin{mdframed}[frametitle={\method\\}]
\dslfontsize
\begin{subfigure}[t]{0.49\textwidth}
\textsc{WHILE}
{\begin{lstlisting}
DEF run m( 
    WHILE c( frontIsClear c) w( 
        turnRight 
        move 
        pickMarker 
        turnRight 
        w) 
    m)
\end{lstlisting}}
\end{subfigure}
\begin{subfigure}[t]{0.49\textwidth}
\textsc{IFELSE+WHILE}
{\begin{lstlisting}
DEF run m( 
    IFELSE c( not c( noMarkersPresent c) c) i( 
        move 
        turnRight 
        i) ELSE e( 
        move 
        e) 
    REPEAT R=2 r( 
        move 
        r)
    WHILE c( leftIsClear c) w( 
        turnLeft 
        w) 
    m)   
\end{lstlisting}}
\end{subfigure}
\begin{subfigure}[t]{0.49\textwidth}
\textsc{2IF+IFELSE}
{\begin{lstlisting}
DEF run m( 
    IFELSE c( frontIsClear c) i( 
        putMarker 
        move 
        i) ELSE e( 
        move 
        e) 
    IF c( rightIsClear c) i( 
        move 
        i) 
    move 
    m)    
\end{lstlisting}}
\end{subfigure}
\begin{subfigure}[t]{0.49\textwidth}
\textsc{WHILE+2IF+IFELSE}
{\begin{lstlisting}
DEF run m( 
    WHILE c( leftIsClear c) w( 
        turnLeft 
        w) 
    IF c( frontIsClear c) i( 
        putMarker 
        move 
        i) 
    move 
    move 
    m)    
\end{lstlisting}}
\end{subfigure}
\end{mdframed}

\caption[Program Reconstruction Task Synthesized Programs]{\textbf{Example program reconstruction task programs generated by all methods.} The programs
that achieve the highest reward while being representative of programs generated by most seeds are shown. 
The na\"{i}ve program synthesis baseline
usually generates the simplest programs, with fewer conditional statements and loops than the \method\\ ablations. Notably, it fails to generate 
\text{IFELSE} statements on these examples, while \method\\ has no problem doing so.}
\label{fig:program reconstruction examples}
\end{figure*}

\section{Optimal and Synthesized Programs}
\label{sec:appendix programmatic policies}
In this section, 
we present the programs from the testing set 
which are selected for conducting ablation studies in the main paper
in~\myfig{fig:reconstruction tasks}.
Also, we manually write programs that
induce optimal behaviors to solve the Karel tasks
and present them in~\myfig{fig:reconstruction tasks}.
Note that while we only show one optimal program for each task,
there exist multiple programs that exhibit the desired behaviors for each task.
Then, we analyze the program reconstructed by \method\\, its ablations, and the na\"{i}ve program synthesis baseline
in~\mysecref{sec:program behavior reconstruction ex}, 
and discuss the programs synthesized by \method\\ for Karel tasks in~\mysecref{sec:karel program ex}.

\begin{figure*}[h]
\centering
\begin{mdframed}[frametitle={\method\\} Karel Programs]
\begin{subfigure}[t]{0.32\textwidth}
\textsc{StairClimber}
{\begin{lstlisting}
DEF run m( 
    WHILE c( noMarkersPresent c) w( 
        turnRight 
        move 
        w) 
    WHILE c( rightIsClear c) w( 
        turnLeft 
        w) 
    m)
\end{lstlisting}}
\end{subfigure}
\begin{subfigure}[t]{0.32\textwidth}
\textsc{TopOff}
{\begin{lstlisting}
DEF run m( 
    WHILE c( noMarkersPresent c) w( 
        move 
        w) 
    putMarker 
    move 
    WHILE c( not c( markersPresent c) c) w( 
        move w)
    putMarker 
    move 
    WHILE c( not c( markersPresent c) c) w( 
        move 
        w) 
    putMarker 
    move 
    turnRight 
    turnRight 
    turnRight 
    turnRight 
    turnRight 
    turnRight
    turnRight 
    turnRight 
    m)
\end{lstlisting}}
\end{subfigure}
\begin{subfigure}[t]{0.32\textwidth}
\textsc{CleanHouse}
{\begin{lstlisting}
DEF run m( 
    WHILE c( noMarkersPresent c) w( 
        turnRight 
        move 
        move 
        turnLeft 
        turnRight 
        pickMarker 
        w) 
    turnLeft 
    turnRight 
    m)
\end{lstlisting}}
\end{subfigure}
\begin{subfigure}[t]{0.32\textwidth}
\textsc{FourCorner}
{\begin{lstlisting}
DEF run m( 
    turnRight 
    move 
    turnRight 
    turnRight 
    turnRight 
    WHILE c( frontIsClear c) w( 
        move 
        w) 
    turnRight 
    putMarker
    WHILE c( frontIsClear c) w( 
        move 
        w) 
    turnRight 
    putMarker 
    WHILE c( frontIsClear c) w( 
        move 
        w) 
    turnRight 
    putMarker
    WHILE c( frontIsClear c) w( 
        move 
        w) 
    turnRight 
    putMarker 
    m)
\end{lstlisting}}
\end{subfigure}
\begin{subfigure}[t]{0.32\textwidth}
\textsc{Maze}
{\begin{lstlisting}
DEF run m( 
    IF c( frontIsClear c) i( 
        turnLeft 
        i) 
    WHILE c( noMarkersPresent c) w( 
        turnRight 
        move 
        w) 
    m)
\end{lstlisting}}
\end{subfigure}
\begin{subfigure}[t]{0.32\textwidth}
\textsc{Harvester}
{\begin{lstlisting}
DEF run m( 
    turnLeft 
    turnLeft 
    pickMarker 
    move 
    pickMarker 
    pickMarker 
    move 
    pickMarker 
    move 
    pickMarker 
    move 
    pickMarker 
    move 
    turnLeft
    pickMarker 
    move 
    pickMarker 
    move 
    pickMarker 
    move 
    pickMarker 
    move 
    turnLeft 
    pickMarker 
    move 
    pickMarker 
    move 
    pickMarker 
    move
    pickMarker 
    move 
    turnLeft 
    pickMarker 
    move 
    pickMarker 
    move 
    pickMarker 
    move 
    m)
\end{lstlisting}}
\end{subfigure}
\end{mdframed}
\caption[\method\\ Karel Tasks Synthesized Programs]{\textbf{Example Karel programs generated by \method\\.} The programs that 
achieved the best reward out of all seeds are shown.}
\label{fig:karel program examples}
\end{figure*}
\subsection{Program Behavior Reconstruction}
\label{sec:program behavior reconstruction ex}

This section serves as a complement to the ablation studies in the main paper, 
where we aim to justify the effectiveness of the proposed framework and the learning objectives.
To this end, we select programs that are unseen to \method\\ and its ablations during the learning program embedding space from the testing set 
and reconstruct those programs using \method\\, its ablations and the na\"{i}ve program synthesis baseline.
Those selected programs are shown in~\myfig{fig:reconstruction tasks}
and the reconstructed programs are shown
in~\myfig{fig:program reconstruction examples}.

The na\"{i}ve program synthesis baseline fails on the complex \textsc{WHILE+2IF+IFELSE} program,
as it rarely synthesizes conditional
and loop statements, 
instead generating long sequences of 
action tokens that attempt to replicate the desired behavior
of those statements.
We believe that this is because it is incentivized to initially predict action tokens to
gain more immediate reward, making it less likely to synthesize other tokens.
\method\\ and its variations perform better 
and synthesize more complex programs, 
demonstrating the importance of the proposed two-stage learning scheme in biasing program search.
Also, \method\\ synthesizes programs that are more concise
and induce behaviors which are more similar to
given testing programs,
justifying the effectiveness of the proposed learning objectives.

\subsection{Karel Environment Tasks}
\label{sec:karel program ex}

This section is complementary to the main experiments in the main paper, 
where we compare \method\\ against the baselines 
on a set of Karel tasks, 
which is described in detail in~\mysecref{sec:env details}.
The programs synthesized by \method\\ are presented 
in~\myfig{fig:karel program examples}.

The synthesized programs solve both \textsc{StairClimber} and \textsc{Maze}.
For \textsc{TopOff}, since the average expected number 
of markers presented in the last row is $3$,
\method\\ synthesizes a sub-optimal program that conducts 
the topoff behavior three times. 
For \textsc{CleanHouse},
while all the baselines fail on this task,
the synthesized program achieves some performance by
simply moving around and try to pick up markers.
For \textsc{Harvester},
\method\\ fails to 
acquire the desired behavior that required nested loops 
but produces a sub-optimal program that contains only action tokens.

\section{Additional Generalization Experiments}
\label{sec:appendix_generalization}
Here, we present additional generalization experiments to complement those presented in \mysecref{sec:scalability}. In \mysecref{sec:appendix_100x100_extended}, we extend the 100x100 state size zero-shot generalization experiments to 3 additional tasks. In \mysecref{sec:appendix_unseen_config}, we analyze how well baseline methods and \method\\ can generalize to unseen configurations of a given task.
\subsection{Generalization on \textsc{FourCorner}, \textsc{TopOff}, and \textsc{Harvester}}
\label{sec:appendix_100x100_extended}
\begin{table} 
    \vspace{-0.4cm}
    \centering
    \caption[]{\small
    Extended reward comparison on original tasks with $8\times8$ or $12\times12$ grids and zero-shot generalization to $100\times100$ grids. 
    \method\\ achieves the best generalization performance on all the tasks except for \textsc{Harvester}.
    }
    \label{table:transfer learning comparison extended}
    \scalebox{0.8}{\begin{tabular}{@{}ccccccc@{}}\toprule
    &   & \textsc{StairClimber} & \textsc{Maze} & \textsc{FourCorner} & \textsc{TopOff} & \textsc{Harvester} \\
    \hline
    \centering
    \small
    \multirow{2}{*}{DRL} 
    & Original & \textbf{1.00} (0.00) & \textbf{1.00} (0.00) & 0.29 (0.05) & 0.32 (0.07) &  \textbf{0.90} (0.10) \\ 
    & 100x100  & 0.00 (0.00) & 0.00 (0.00) & 0.00 (0.00) & 0.01 (0.01) & 0.00 (0.00) \\
    \midrule
    \multirow{2}{*}{DRL-abs} 
    & Original & 0.13 (0.29) & \textbf{1.00} (0.00) & 0.36 (0.44) & 0.63 (0.23) &  0.32 (0.18) \\ 
    & 100x100  & 0.00 (0.00) & 0.04 (0.05) & 0.37 (0.44) & 0.15 (0.12) & 0.02 (0.01) \\
    \midrule
    \multirow{2}{*}{DRL-FCN} 
    & Original &  \textbf{1.00} (0.00) & 0.97 (0.03) & 0.20 (0.34) & 0.28 (0.12) & 0.46 (0.16) \\ 
    & 100x100  & -0.20 (0.10) & 0.01 (0.01) & 0.00 (0.00) & 0.01 (0.01) & 0.02 (0.00) \\
    \midrule
    \multirow{2}{*}{VIPER} 
    & Original & 0.02 (0.02) & 0.69 (0.05) & 0.40 (0.42) & 0.30 (0.06) & 0.51 (0.07) \\ 
    & 100x100  & 0.00 (0.00) & 0.10 (0.12) & 0.40 (0.42) & 0.03 (0.00) & \textbf{0.04} (0.00) \\
    \midrule
    \multirow{2}{*}{\method\\} 
    & Original & \textbf{1.00} (0.00) & \textbf{1.00} (0.00) & \textbf{0.45} (0.40) & \textbf{0.81} (0.07)  & 0.45 (0.28) \\ 
    & 100x100  & \textbf{1.00} (0.00) & \textbf{1.00} (0.00) & \textbf{0.45} (0.37) & \textbf{0.21} (0.03) & 0.00 (0.00) \\
    \bottomrule
    \end{tabular}}
\end{table}
Evaluating zero-shot generalization performance assumes methods to work reasonably well on the original tasks. For this reason (and due to space limitations) we present only \textsc{StairClimber} and \textsc{Maze} for generalization experiments in the main text in \mysecref{sec:scalability} because most methods achieve reasonable performance on these two tasks, with DRL and LEAPS both solving these tasks fully and DRL-abs solving Maze fully.

However, here we also present full results for all 
tasks except \textsc{CleanHouse} (as no method except \method\\ has a reasonable level of performance on it).
The results are summarized in~\mytable{table:transfer learning comparison extended}. We see that \method\\ generalizes well on \textsc{FourCorner} and maintains the best performance on \textsc{TopOff}. It is outperformed on \textsc{Harvester}, although none of the methods do well on \textsc{Harvester} as the highest obtained reward by any method is 0.04 (by VIPER). In summary, \method\\ performs the best on 4 out of these 5 tasks, further demonstrating its superior zero-shot generalization performance.

Furthermore, we note that it is possible that a DRL policy employing a fully convolutional network (FCN) as proposed in \citet{fcn} can handle varying observation sizes. FCNs were also demonstrated in \citet{silver2020few} to demonstrate better generalization performance than traditional convolutional neural network policies. However, we hypothesize that the generalization performance here will still be poor as there is a large increase in the number of features that the FCN architecture needs to aggregate when transferring from 8x8/12x12 state inputs to 100x100 inputs---a ~10x input size increase that FCN is not specifically designed to deal with.
We have included both FCN's zero-shot generalization results and its results on the original grid sizes in ~\mytable{table:transfer learning comparison extended}. DRL-FCN, where we have replaced the policy and value function networks of PPO with an FCN, does manage to perform zero-shot transfer marginally better than DRL performs when training from scratch (as it DRL's architecture cannot handle varied input sizes) on \textsc{Maze} and \textsc{Harvester}. However, it obtains a negative reward on \textsc{StairClimber} as it attempts to navigate away from the stairs when transferring to the $100 \times 100$ grid size. Its performance is still far worse than LEAPS and VIPER on most tasks, demonstrating that the programmatic structure of the policy is important for these tasks.

\subsection{Generalization to Unseen Configurations}
\label{sec:appendix_unseen_config}

\begin{table}
    \centering
    \caption[Unseen Configurations Performance]{Mean return (standard deviation) [\% change in performance]
    on generalizing to unseen configurations
    on \textsc{TopOff} and \textsc{Harvester} task.
    }
    \label{table:diff config}
    \scalebox{0.73}{
    \begin{tabular}{@{}cccccc@{}}\toprule
    \textsc{TopOff} & \multicolumn{5}{c}{Training configuration \%}
    \\ 
    & 75\% & 50\% & 25\% & 10\% & 5\% \\
    \midrule
    \centering
    \small
    DRL & 0.17 (0.05) [-46.8\%]
    & 0.12 (0.09) [-62.5\%] & 0.12 (0.06) [-62.5\%] & 0.17 (0.13) [-46.8\%] & 0.13 (0.04) [-59.4\%]
    \\
    DRL-abs & 0.23 (0.29) [-63.5\%] & 0.29 (0.36) [-54.0\%] & 0.45 (0.45) [-28.6\%] & 0.24 (0.38) [-61.9\%] & 0.26 (0.37) [-18.8\%] \\
    VIPER & 0.27 (0.03) [-10.0\%] & 0.28 (0.04) [-6.67\%] & 0.27 (0.06) [-10.0\%] & 0.27 (0.02) [-10.0\%] & 0.28 (0.03) [-6.67\%] \\
    \midrule
    \method\\ & 
    \textbf{0.68} (0.18) [-15.0\%] & \textbf{0.65} (0.13) [-18.8\%] &
    \textbf{0.61} (0.24) [-23.8\%] & \textbf{0.68} (0.21) [-15.0\%] & \textbf{0.67} (0.18) [-16.3\%]
    \\
    \bottomrule
    \vspace{0.1cm}
    \end{tabular}}
    
    \scalebox{0.73}{
    \begin{tabular}{@{}cccccc@{}}\toprule
    \textsc{Harvester} & \multicolumn{5}{c}{Training configuration \%}
    \\ 
    & 75\% & 50\% & 25\% & 10\% & 5\% \\
    \midrule
    \centering
    \small
    DRL & \textbf{0.64} (0.24) [-28.9\%] & 0.71 (0.29) [-21.1\%] & 0.21 (0.06) [-76.7\%] & 0.14 (0.09) [-84.4\%] & 0.04 (0.01) [-95.6\%]
    \\
    DRL-abs & 0.14 (0.21) [-56.3\%] & 0.24 (0.25) [-25.0\%] & 0.05 (0.06) [-84.4\%] & 0.13 (0.21) [-59.4\%] & 0.31 (0.31) [-3.13\%]  \\
    VIPER & 0.54 (0.01) [+5.88\%] & \textbf{0.54} (0.02) [+5.88\%] & \textbf{0.55} (0.01) [+7.84\%] & \textbf{0.54} (0.01) [+5.88\%] & \textbf{0.44} (0.22) [-13.7\%] \\
    \midrule
    \method\\ & 0.40 (0.30) [-13.0\%] & 0.42 (0.27) [-8.69\%] & 0.50 (0.35) [+08.69\%] & 0.12 (0.19) [-73.9\%] & 0.01 (0.03) [-97.6\%]
    \\
    \bottomrule
    \vspace{0.1cm}
    \end{tabular}}

\end{table}

\Skip{
\begin{table}
    \centering
    \caption{Mean return (standard deviation) 
    on generalizing to unseen configurations
    on \textsc{TopOff} task.
    }
    \label{table:diff config}
    \scalebox{0.8}{\begin{tabular}{@{}cccc@{}}\toprule
    & Training (25\% configurations) $\uparrow$
    & Testing (75\% configurations) $\uparrow$
    & Performance Drop (\%) $\downarrow$\\
    \midrule
    \centering
    \small
    DRL & 0.17 (0.09) & 0.14 (0.07) & 21.4\% \\
    DRL-abs & 0.33 (0.34) & 0.30 (0.35) & 10.0\% \\
    VIPER & 0.29 (0.04) & 0.26 (0.02) & 11.5\% \\
    \midrule
    \method\\ & 
    \textbf{0.82} (0.14) & \textbf{0.79} (0.15) & \textbf{3.65}\% \\
    \bottomrule
    \vspace{0.1cm}
    \end{tabular}}
    
    \Skip{
    \scalebox{0.8}{\begin{tabular}{@{}cccc@{}}\toprule
    & Training (25\% configurations) $\uparrow$
    & Testing (75\% configurations) $\uparrow$
    & Performance Drop (\%) $\downarrow$\\
    \midrule
    \centering
    \small
    DRL & 0.17 (0.09) & 0.14 (0.07) & 21.4\% \\
    DRL-abs & 0.33 (0.34) & 0.30 (0.35) & 10.0\% \\
    VIPER & 0.29 (0.04) & 0.26 (0.02) & 11.5\% \\
    \midrule
    \method\\ & 
    \textbf{0.82} (0.14) & \textbf{0.79} (0.15) & \textbf{3.65}\% \\
    \bottomrule
    \end{tabular}}
    }
    
\end{table}
}

We present a generalization experiment in the main paper to study how well 
the baselines and the programs synthesized by the proposed framework 
can generalize to 
larger state spaces that are unseen during training without further learning 
on the \textsc{StairClimber} and \textsc{Maze} tasks.
In this section, we investigate the ability of generalizing to different configurations,
which are defined based on the marker placement related to solve a task,
on both the \textsc{TopOff} task and \textsc{Harvester} task.

Since solving \textsc{TopOff} requires an agent to put markers on top of all markers on the last row,
the initial configurations are determined by the marker presence on the last row.
The grid has a size of $10 \times 10$ inside the surrounding wall.
We do not spawn a marker at the bottom right corner in the last row, 
leaving $9$ possible locations with marker,
allowing $2^{9}$ possible initial configurations.
On the other hand, \textsc{Harvester} requires an agent to pick up
all the markers placed in the grid.
The grid has a size of $6 \times 6$ inside the surrounding wall,
leaving $36$ possible locations in grid with a marker,
resulting in $2^{36}$ possible initial configurations.

We aim to test if methods can learn from only a small portion of configurations during training 
and still generalize to all the possible configurations without further learning.
To this end,
we experiment using $75\%$, $50\%$, $25\%$, $10\%$, $5\%$ of the configurations 
for training DRL, DRL-abs, and VIPER and for the program search stage of \method\\.
Then, we test zero-shot generalization of the learned models and programs 
on all the possible configurations.
We report the performance in~\mytable{table:diff config}.
We compare the performance each method achieves 
to its own performance learning from all the configurations (reported in the main paper)
to investigate how limiting training configurations affects the performance.
Note that the results of training and testing on $100\%$ configurations
are reported in the main paper, 
where no generalization is required.

\textbf{\textsc{TopOff.}}
\method\\ outperforms all the baselines on the mean return 
on all the experiments.
VIPER and \method\\ enjoy the lowest and the second lowest performance decrease 
when learning from only a portion of configurations,
which demonstrates the strength of programmatic policies.
DRL-abs slightly outperforms DRL, with better absolute performance and lower performance decrease.
We believe that this is because DRL takes entire Karel grids as input,
and therefore held out configurations are completely unseen to it.
In contrast, DRL-abs takes abstract states (\ie local perceptions) as input,
which can alleviate this issue.

\textbf{\textsc{Harvester.}}
VIPER outperforms almost all other methods on absolute performance and performance decrease, while
\method\\ achieves second best results,
which again justifies the generalization of programmatic policies.
Both DRL and DRL-abs are unable to generalize well when
learning from a limited set of configurations,
except in the case of DRL-abs learning from $5\%$ of configurations,
which can be attributed to the high-variance of DRL-abs results.%

\section{Additional Analysis on Experimental Results}
\label{sec:additional_analysis}

Due to the limited space in the main paper, we include additional analysis of the experimental results in this section.

\subsection{DRL vs. DRL-abs}

We hypothesize that DRL-abs does not always outperform DRL due to imperfect perception 
(i.e. state abstraction) design. 
DRL-abs takes abstract states as input (\ie \texttt{frontIsClear()}, \texttt{leftIsClear()}, \texttt{rightIsClear()}, \texttt{markerPresent()} in our design), 
which only describe local perception while omitting the information of the entire map.
Therefore, for tasks such as \textsc{StairClimber}, \textsc{Harvester}, and \textsc{CleanHouse}, 
which would be easier to solve with access to the entire Karel grid, 
DRL might outperform DRL-abs. 
In this work, DRL-abs’ abstract states are the perceptions from 
the DSL we synthesize programs with 
to make the comparisons fair against our method as well as 
analyzing the effects of abstract states in the DRL domain. 
However, a more sophisticated design for perception/state abstraction could potentially improve the performance of DRL-abs.

\subsection{VIPER generalization}

VIPER operates on the abstract state space which is invariant to grid size. 
However, for the reasons below, 
it is still unable to transfer the behavior to the larger grid 
despite its abstract state representation. 
We hypothesize that VIPER’s performance suffers on zero-shot generalization for two main reasons.
\begin{enumerate}
    \item It is constrained to imitate the DRL teacher policy during training, which is trained on the smaller grid sizes. Thus its learned policy also experiences difficulty in zero-shot generalization to larger grid sizes.
    \item Its decision tree policies cannot represent certain looping behaviors as they simply perform a one-to-one mapping from abstract state to action, thus making it difficult to learn optimal behaviors that require a one-to-many mapping between an abstract state and a set of desired actions. Empirically, we observed that training losses for VIPER decision trees were much higher for tasks such as \textsc{StairClimber} which require such behaviors.
\end{enumerate}
\section{Detailed Descriptions and Illustrations of Ablations and Baselines}
\label{sec:illustration}

This section provides details on 
the variations of \method\\ used for ablations studies
and the baselines which we compare against.
The descriptions of the ablations of \method\\ are presented in~\mysecref{sec:appendix_ablation} and
the illustrations are shown in \myfig{fig:appendix_ablation}.
The na\"{i}ve program synthesis baseline is illustrated in \myfig{fig:baselines} (c)
for better visualization.
Then, the descriptions of the baselines are presented in~\mysecref{sec:appendix_baseline} and
the illustrations are shown in \myfig{fig:baselines}.

\subsection{Ablations}
\label{sec:appendix_ablation}
We first ablate various components of our proposed framework in order to (1) justify the necessity of the proposed two-stage learning scheme and (2) identify the effects of the proposed objectives.
We consider the following baselines and ablations of our method.

\begin{figure*}[h]
    \centering
    \includegraphics[width=\textwidth]{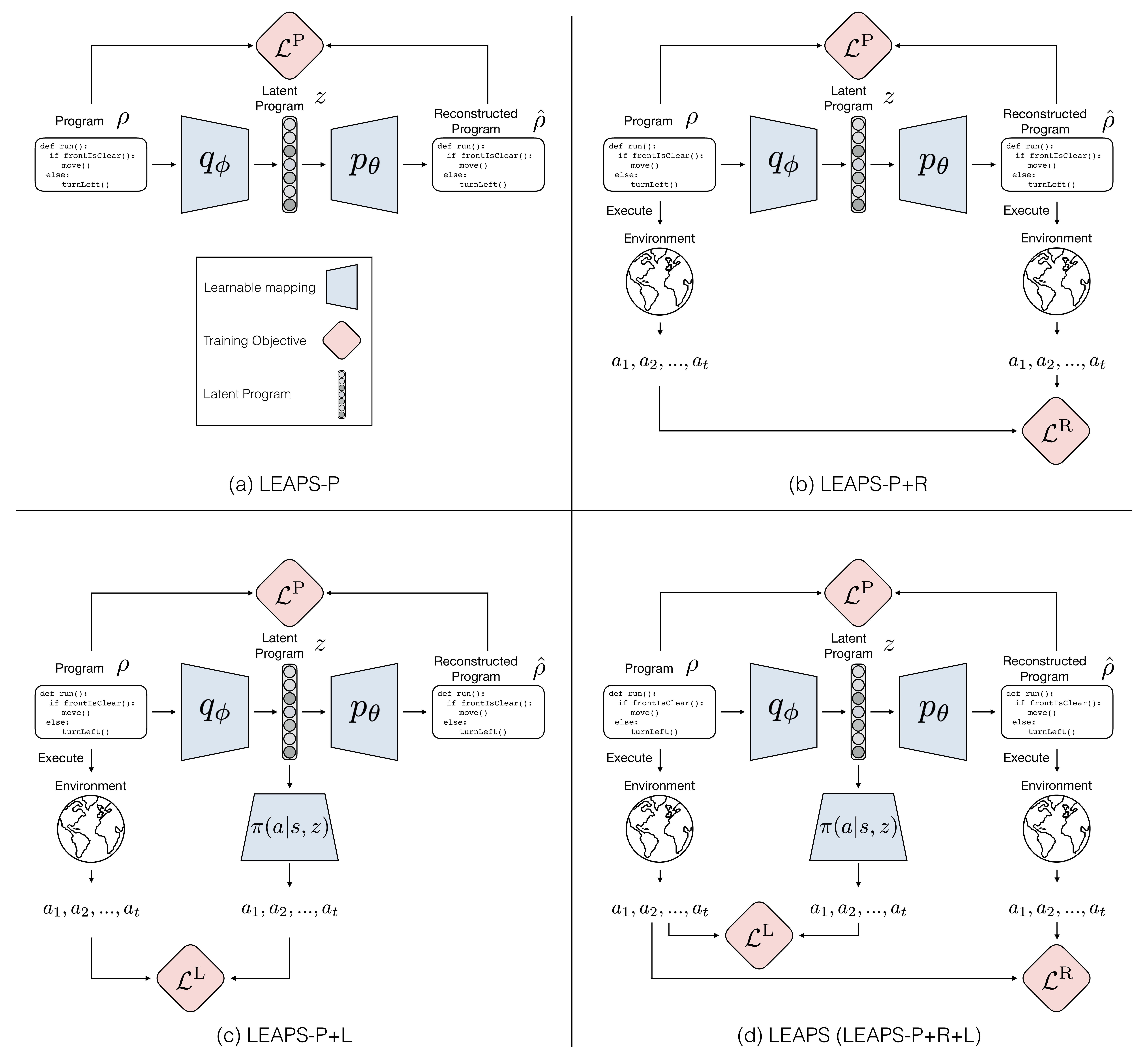}
    \caption[\method\\ Ablations Illustrations]{
    \textbf{\method\\ Variations Illustrations.}
    Blue trapezoids represent the modules whose parameters are being learned in the learning program embedding stage.
    Red diamonds represent the learning objectives.
    Gray rounded rectangle represent latent programs (\ie program embeddings), which are vectors.
    (a) \method\\-P: 
    the simplest ablation of \method\\, in which
    the program embedding space is learned by only optimizing the program reconstruction loss $\mathcal{L}^{\text{P}}$.
    (b) \method\\-P+R: 
    an ablation of \method\\ which optimizes both 
    the program reconstruction loss $\mathcal{L}^{\text{P}}$
    and the program behavior reconstruction loss $\mathcal{L}^{\text{R}}$. 
    (c) \method\\-P+L: 
    an ablation of \method\\ which optimizes both 
    the program reconstruction loss $\mathcal{L}^{\text{P}}$
    and the latent behavior reconstruction loss $\mathcal{L}^{\text{L}}$.
    (d) \method\\ (\method\\-P+R+L): 
    our proposed framework that optimizes all
    the proposed objectives.
    }
    \label{fig:appendix_ablation}
\end{figure*}

\begin{itemize}
    \item Na\"{i}ve: the na\"{i}ve program synthesis baseline is a policy that learns to directly synthesize a program from scratch by recurrently predicting a sequence of program tokens.
    The architecture of this baseline is a recurrent neural network
    which takes an initial starting token as the input at the first time step,
    and then sequentially outputs a program token at each time step to compose a program 
    until an end token is produced. 
    Note that the observation of this baseline is its own previously outputted program token 
    instead of the state of the task environment (\eg Karel grids). 
    Also, at each time step,
    this baseline produces a distribution over all the possible program tokens in the given DSL 
    instead of a distribution over agent's action in the task environment (\eg \texttt{move()}).
    This baseline investigates if an end-to-end learning method can solve the problem.
    This baseline is illustrated in \myfig{fig:baselines} (c).
    
    \item \method\\-P: 
    the simplest ablation of \method\\, in which
    the program embedding space is learned by only optimizing the program reconstruction loss $\mathcal{L}^{\text{P}}$.
    This baseline is illustrated in \myfig{fig:appendix_ablation} (a).
    
    \item \method\\-P+R: 
    an ablation of \method\\ which optimizes both 
    the program reconstruction loss $\mathcal{L}^{\text{P}}$
    and the program behavior reconstruction loss $\mathcal{L}^{\text{R}}$. 
    This baseline is illustrated in \myfig{fig:appendix_ablation} (b).
    
    \item \method\\-P+L: 
    an ablation of \method\\ which optimizes both 
    the program reconstruction loss $\mathcal{L}^{\text{P}}$
    and the latent behavior reconstruction loss $\mathcal{L}^{\text{L}}$.
    This baseline is illustrated in \myfig{fig:appendix_ablation} (c).
    
    \item \method\\ (\method\\-P+R+L): 
    \method\\ with all the losses, optimizing our full objective. 
    \item \method\\-rand-\{8/64\}: 
    like \method\\, this ablation also optimizes the full objective for learning the program embedding space. 
    But when searching latent programs, instead of CEM,
    it simply randomly samples 8/64 candidate latent programs and 
    chooses the best performing one. 
    These baselines justify the effectiveness of using CEM for searching latent programs.
\end{itemize}

\subsection{Baselines}
\label{sec:appendix_baseline}

\begin{figure*}[h]
    \centering
    \includegraphics[width=\textwidth]{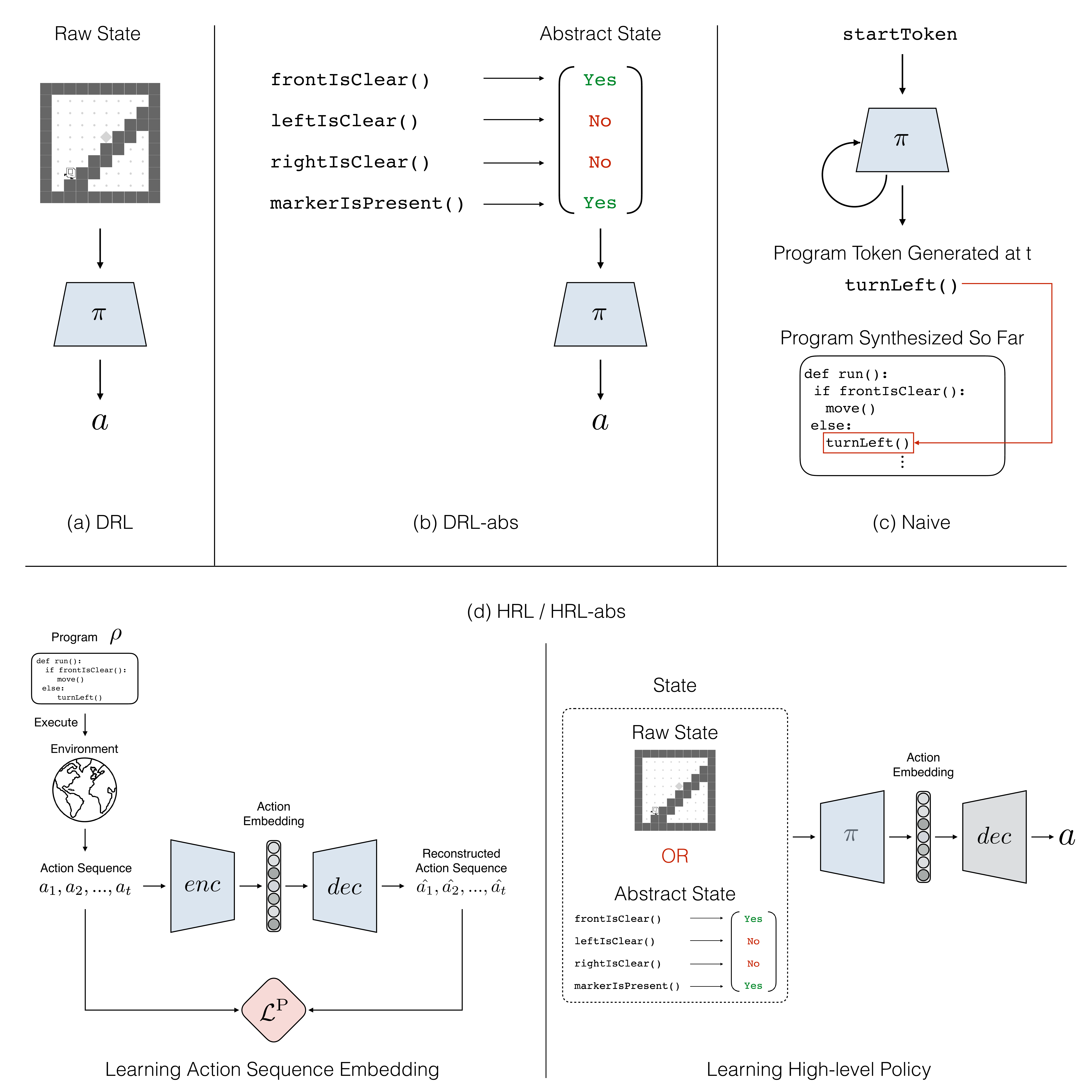}
    \caption[Baseline Methods Illustrations]{
    \textbf{Baseline Methods Illustrations.}
    (a) DRL: a DRL policy that takes raw state input (\ie a Karel grid represented as a $W \times H \times 12$ binary tensor as there are 12 possible states for each grid square).
    (b) DRL-abs: a DRL policy that takes abstract state input, containing a vector of returned values of perceptions, \eg \texttt{frontIsClear()==true} and \texttt{markersPresent()==false}.
    (c) Naive: a na\"{i}ve program synthesis baseline that learns to directly synthesize a program from scratch by recurrently predicting a sequence of program tokens.
    (d) HRL/HRL-abs: 
    a hierarchical RL baseline in which a VAE, 
    consisting of a encoder $enc$ and a decoder $dec$, 
    is first trained to reconstruct action sequences 
    from program execution traces used by \method\\. 
    Once the action embedding space is learned, 
    it employs a high-level policy $\pi$ that learns from scratch
    to solve task by predicting a distribution in the learned 
    action embedding space.
    Note that the parameters of the decoder $dec$ are frozen
    (represented in gray) when the high-level policy is learning.
    The HRL policy takes raw state input (same as the DRL baseline) 
    and the HRL-abs policy takes abstract state input (same as the DRL-abs baseline).
    }
    \label{fig:baselines}
\end{figure*}

We evaluate \method\\ against the following baselines (illustrated in \myfig{fig:baselines}).

\begin{itemize}
    \item DRL: a neural network policy trained on each task 
    and taking raw states (Karel grids) as input.
    A Karel grid is represented as a binary tensor with dimension $W \times H \times 16$ (there are 16 possible states for each grid square)
    instead of an image.
    This baseline is illustrated in \myfig{fig:baselines} (a).
    
    \item DRL-abs: a recurrent neural network policy directly trained on each Karel task 
    but instead of taking raw states (Karel grids) as input
    it takes
    \textit{abstract} states as input
    (\ie it sees the same perceptions as \method\\).
    Specifically, all returned values
    of perceptions including \texttt{frontIsClear()==true},
    \texttt{leftIsClear()==false},
    \texttt{rightIsClear()==true},
    \texttt{markersPresent()==false},
    and \texttt{noMarkersPresent()==true}
    are concatenated as a binary vector,
    which is then fed to the DRL-abs policy as its input.
    This baseline allows for a fair comparison to \method\\ 
    since the program execution process also utilizes abstract state information.
    This baseline is illustrated in \myfig{fig:baselines} (b).
    
    \item DRL-abs-t: a DRL \textit{transfer} learning baseline 
    in which for each task, 
    we train DRL-abs policies on all other tasks,
    then fine-tune them on the current
    task. 
    Thus it acquires a prior by learning to first solve other Karel tasks.
    Rewards are reported for the policies
    from the task that transferred with highest return.
    We only transfer DRL-abs policies as some tasks have different state spaces 
    so that transferring a DRL policy trained on a task to another task with a different state space is not possible. 
    
    This baseline is designed to investigate if acquiring task related priors 
    allows DRL policies to perform better on our Karel tasks.
    Unlike \method\\, which acquires priors from a dataset consisting of randomly generated programs and the behaviors those program induce in the environment,
    DRL-abs-t allows for acquiring priors from goal-oriented behaviors (\ie other Karel tasks).
    
    \item HRL: a hierarchical RL baseline in which a VAE is first trained on 
    action sequences from program execution traces used by \method\\. 
    Once trained, the decoder is utilized as a low-level policy
    for learning a high-level policy to sample actions from.
    Similar to \method\\, this baseline utilizes the dataset to produce a prior of the domain.
    It takes raw states (Karel grids) as input.
    
    This baseline is also designed to investigate if acquiring priors 
    allow DRL policies to perform better.
    Similar to \method\\, which acquires priors from a dataset consisting of randomly generated programs and the behaviors those program induce in the environment,
    HRL is trained to acquire priors by learning to reconstruct the behaviors induced by the programs.
    One can also view this baseline as a version of the framework proposed in~\cite{hausman2018learning} with some simplifications, 
    which also learns an embedding space using a VAE 
    and then trains a high-level policy to utilize this embedding space together with the low-level policy whose parameters are frozen.
    This baseline is illustrated in \myfig{fig:baselines} (d).
    
    \item HRL-abs: the same method as HRL but taking 
    abstract states (\ie 
    local perceptions) as input. 
    This baseline is illustrated in \myfig{fig:baselines} (d).
    
    \item VIPER~\citep{bastani2018verifiable}: A decision-tree programmatic policy which imitates the behavior of a deep RL teacher policy via a modified DAgger algorithm \citep{ross2011reduction}. This decision tree policy cannot synthesize loops, allowing us to highlight the performance advantages of more expressive program representation that \method\\ is able to take advantage of.
\end{itemize}

All the baselines are trained with PPO \citep{schulman2017proximal} 
or SAC \citep{haarnoja18b}, 
including the VIPER teacher policy.
More training details can be found in~\mysecref{sec:hyperparameters}.

\section{Program Dataset Generation Details}
\label{sec:dataset details}

To learn a program embedding space for the proposed framework 
and its ablations,
we randomly generate 50k programs to 
form a dataset with 35k training programs and 
7.5k programs for validation and testing.
Simply generating programs by uniformly sampling all the tokens 
from the DSL would yield programs that mainly only contain action tokens 
since the chance to synthesize conditional statements with correct grammar is low. 
Therefore, to produce programs that are longer and deeply nested with conditional statements to induce more complex behaviors, we propose to sample programs using a probabilistic sampler. 

\begin{figure}[!h]
    \centering
    \includegraphics[width=0.49\textwidth]{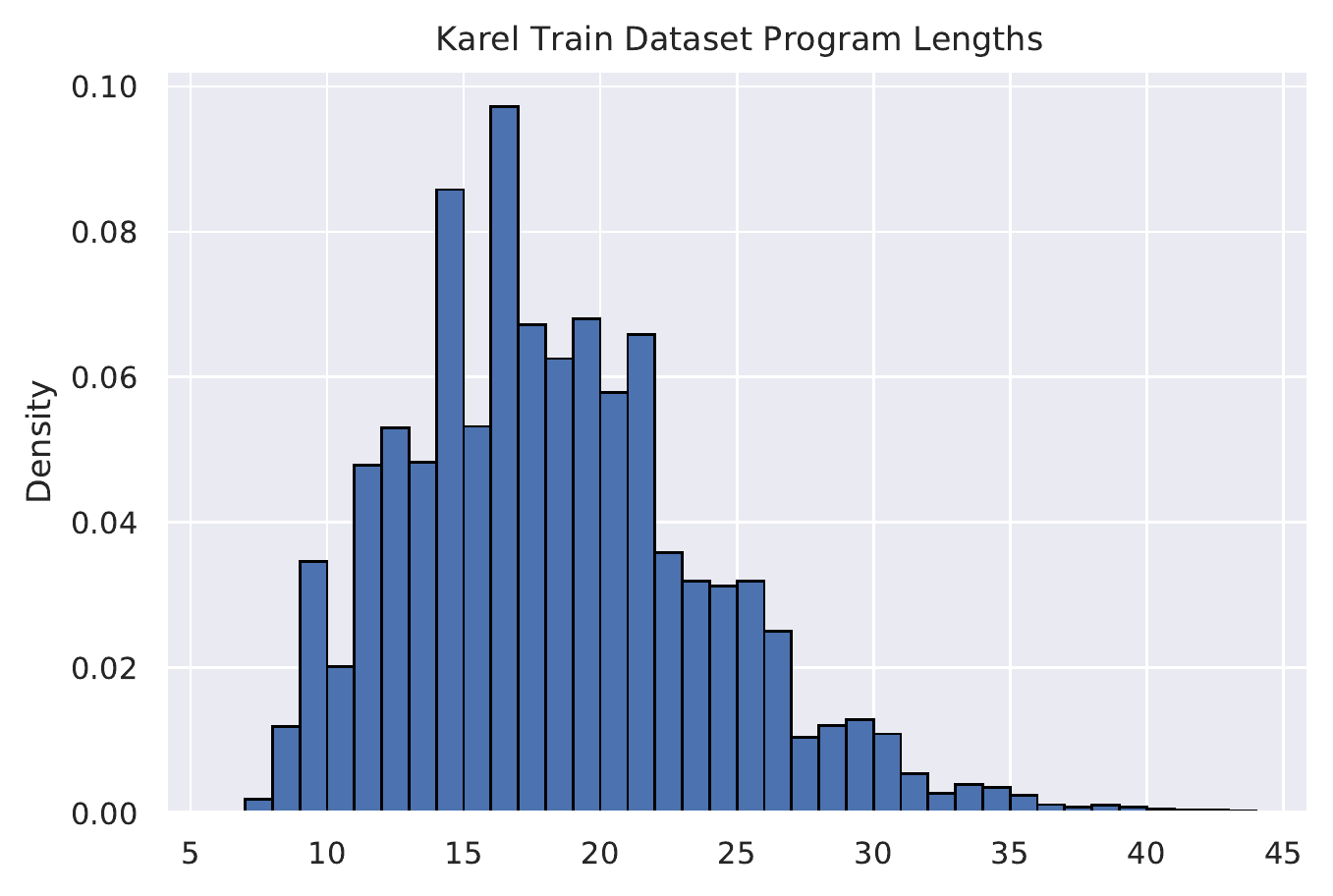}
    \includegraphics[width=0.49\textwidth]{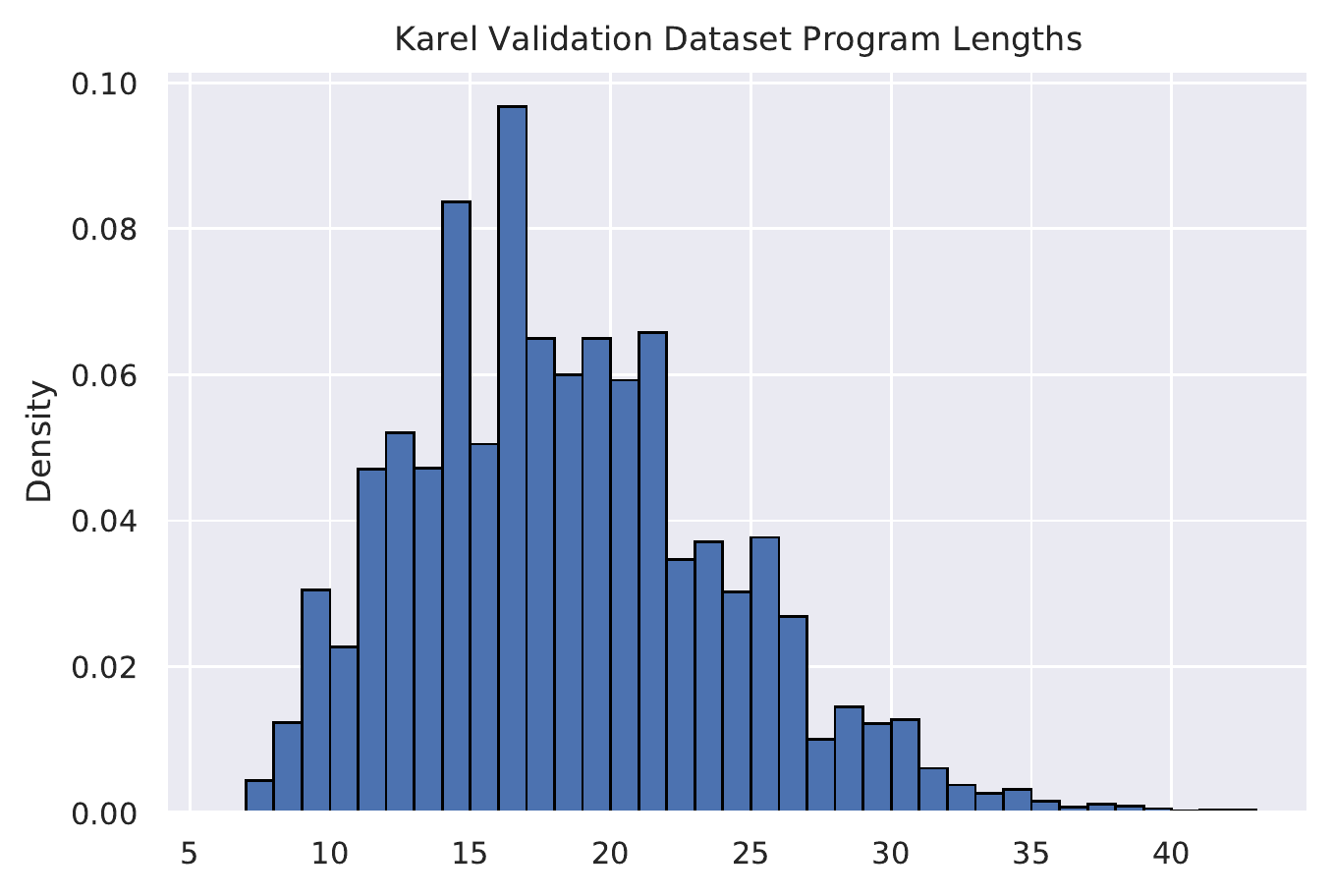}
    \caption[Program Length Histograms]{Histograms of the program length (\ie number of program tokens) in the training and validation datasets.}
    \label{fig:data histogram}
\end{figure}

To generate each program, 
we sample program tokens according to the probabilities listed
in Table~\ref{table:dataset generation probabilities}
at every step until we sample an ending token or 
when a maximum program length is reached.
When generating programs, 
we ensure that no program is identical to any other. 
Each token is generated sequentially, 
and length is effectively governed by the \texttt{STMT\_STMT} token detailed
in Table~\ref{table:dataset generation probabilities}'s caption.
There is
a maximum depth limit of 4 nested conditional/loop statements,
and a maximum statement depth limit of 6 (can't have more than
6 nested \texttt{STMT\_STMT} tokens). Note that 
this sampling procedure does not guarantee that the programs
generated will terminate, hence when executing them 
to obtain ground-truth interactions for training the
Program Behavior and Latent Behavior Reconstruction losses
we limit the max program execution length to 100 environment
timesteps.
This sampling procedure
results in the distribution of program lengths
seen in \myfig{fig:data histogram}. 

Intuitively, shorter lengths can bias synthesized
programs to compress the same behaviors into fewer tokens
through the use of loops, making program search easier.
Therefore, in our experiments, we have limited the maximum
output program length of \method\\ to 45 tokens (as the maximum in the dataset is 44).
As shown in the example programs generated by \method\\ in~\myfig{fig:karel program examples},
\method\\ successfully generates loops for our Karel tasks, 
which can be probably attributed to this bias of program length.
We further verify this intuition by rerunning \method\\
with the max program length set to 100 tokens on the Karel tasks.
We display generated programs in Table \ref{tab:100 program length}, 
where we see that some of the generated programs are indeed much longer and lack loop statements and structures. 

\begin{table*}[!h]
\centering
\small
\caption[Program Token Generation Probabilities]{The probability of sampling program tokens when generating the program dataset. Tokens are generated sequentially, 
and \textsc{STMT\_STMT} refers to 
breaking up the current token into two tokens, each of which
is selected according to the same probability distribution again. Thus it effectively controls how long programs will be.}
\begin{tabular}{@{}cccccccccccc@{}}\toprule
& \texttt{WHILE} & \texttt{REPEAT} & \texttt{STMT\_STMT} & \texttt{ACTION} & \texttt{IF} & \texttt{IFELSE} \\
\cmidrule{2-7}
Standard Dataset & 0.15 & 0.03 & 0.5 & 0.2 & 0.08 & 0.04 \\
\bottomrule
\end{tabular}
\label{table:dataset generation probabilities}
\end{table*}

\renewcommand{\arraystretch}{1.5}
\begin{table*}[]
    \centering
    \small
    \caption[\method\\ Length 100 Synthesized Karel Programs]{\textbf{\method\\ Length 100 Synthesized Karel Programs.} Line breaks are not shown here as the programs are very long.
    The examples picked are ones that 
    represent the programs generated by most seeds for each task. Without the 45 token 
    restriction on program lengths, programs for \textsc{TopOff}, \textsc{fourCorner}, and 
    \textsc{Harvester} are very long and have repetitive movements that 
    can easily be put into \texttt{REPEAT} or \texttt{WHILE} loops. The \textsc{CleanHouse}
    program also contains repeated, somewhat redundant \texttt{WHILE} loops. \textsc{Maze} 
    and \textsc{StairClimber} programs are mostly unaffected by the change in 
    maximum program length. These programs demonstrate that the bias induced by program length restriction is important for producing more complex programs in the program synthesis phase of \method\\. 
    }
    \begin{tabular}{cp{0.78\textwidth}}\toprule
        Karel Task & Program\\
        \cmidrule{1-2}
         \textsc{StairClimber} & {\begin{lstlisting}[frame=none, belowskip=-6.5pt, aboveskip=-6.5pt]
DEF run m( turnLeft turnRight turnLeft turnLeft turnRight WHILE c( noMarkersPresent c)} w( turnLeft move w) m)
         \end{lstlisting}}\\
         \textsc{TopOff} & {\begin{lstlisting}[frame=none, belowskip=-6.5pt, aboveskip=-6.5pt]
DEF run m( WHILE c( noMarkersPresent c) w( move w) turnRight turnRight turnRight turnRight turnRight} turnRight turnRight turnRight turnRight turnRight turnRight turnRight turnRight turnRight turnRight turnRight turnRight turnRight turnRight turnRight turnRight turnRight turnRight turnRight turnRight turnRight turnRight turnRight turnRight putMarker turnRight turnRight move turnRight move turnRight move turnRight move turnRight move turnRight move turnRight move turnRight move turnRight move turnRight move turnRight move turnRight move turnRight move turnRight move turnRight move turnRight move turnRight move turnRight move turnRight move m)
\end{lstlisting}}\\
         \textsc{CleanHouse} & {\begin{lstlisting}[frame=none, belowskip=-6.5pt, aboveskip=-6.5pt]
DEF run m( turnRight pickMarker turnLeft turnRight turnLeft pickMarker move turnLeft WHILE c( leftIsClear c) w( pickMarker move w) turnRight turnLeft pickMarker move turnLeft WHILE c( leftIsClear c) w( pickMarker move w) turnLeft pickMarker} WHILE c( leftIsClear c) w( pickMarker move turnLeft pickMarker w)} WHILE c( noMarkersPresent c) w( turnLeft move pickMarker w) turnLeft pickMarker turnLeft m)
\end{lstlisting}}\\
         \textsc{fourCorner} & {\begin{lstlisting}[frame=none, belowskip=-6.5pt, aboveskip=-6.5pt]
DEF run m( turnRight turnRight turnRight turnRight turnRight turnRight turnRight turnRight turnRight turnRight turnRight turnRight turnRight turnRight turnRight turnRight turnRight turnRight turnRight turnRight turnRight turnRight turnRight turnRight turnRight WHILE c( frontIsClear c) w( move w) turnRight WHILE c( frontIsClear c) w( move w) turnRight WHILE c( frontIsClear c) w( move w) turnRight putMarker WHILE c( frontIsClear c) w( move w) turnRight putMarker WHILE c( frontIsClear c) w( move w)} turnRight putMarker WHILE c( frontIsClear c) w( move w) turnRight putMarker m)
\end{lstlisting}}\\
         \textsc{Maze} & {\begin{lstlisting}[frame=none, belowskip=-6.5pt, aboveskip=-6.5pt]
DEF run m( WHILE c( noMarkersPresent c) w( REPEAT R=1 r( turnRight r) move w) turnLeft turnRight m)
         \end{lstlisting}}\\
         \textsc{Harvester} & {\begin{lstlisting}[frame=none, belowskip=-6.5pt, aboveskip=-6.5pt]
DEF run m( turnLeft turnRight pickMarker move pickMarker move turnRight move pickMarker move pickMarker move turnRight move pickMarker move pickMarker move pickMarker move turnRight move pickMarker move pickMarker move pickMarker move turnRight move pickMarker move pickMarker move pickMarker move pickMarker move turnRight move pickMarker move pickMarker move pickMarker move pickMarker move turnRight move pickMarker move pickMarker move pickMarker move pickMarker move pickMarker move turnRight move pickMarker move pickMarker move pickMarker move pickMarker move pickMarker move turnRight move m)         
         \end{lstlisting}}\\
    \bottomrule
    \end{tabular}
    \label{tab:100 program length}
\end{table*}

\section{Karel Task Details}
\label{sec:env details}
\begin{figure*}[h]
    \centering
    \begin{subfigure}[b]{0.45\textwidth}
    \centering
    \includegraphics[trim=25 16 25 55, clip,height=0.45\textwidth]{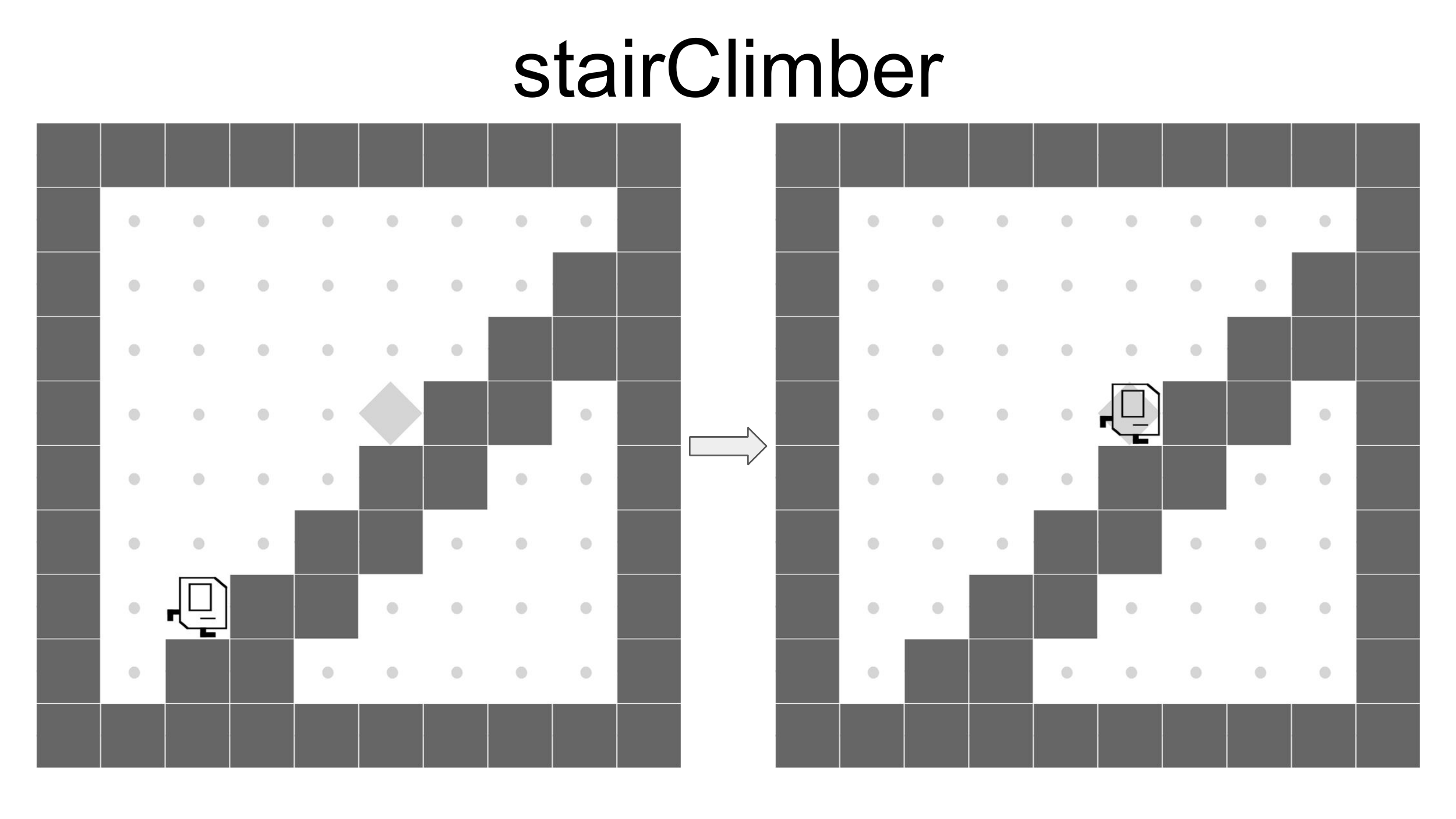}
    \caption{\textsc{StairClimber}}
    \end{subfigure}
    \hspace{0.5cm}
    \begin{subfigure}[b]{0.45\textwidth}
    \centering
    \includegraphics[trim=25 16 25 55, clip,height=0.45\textwidth]{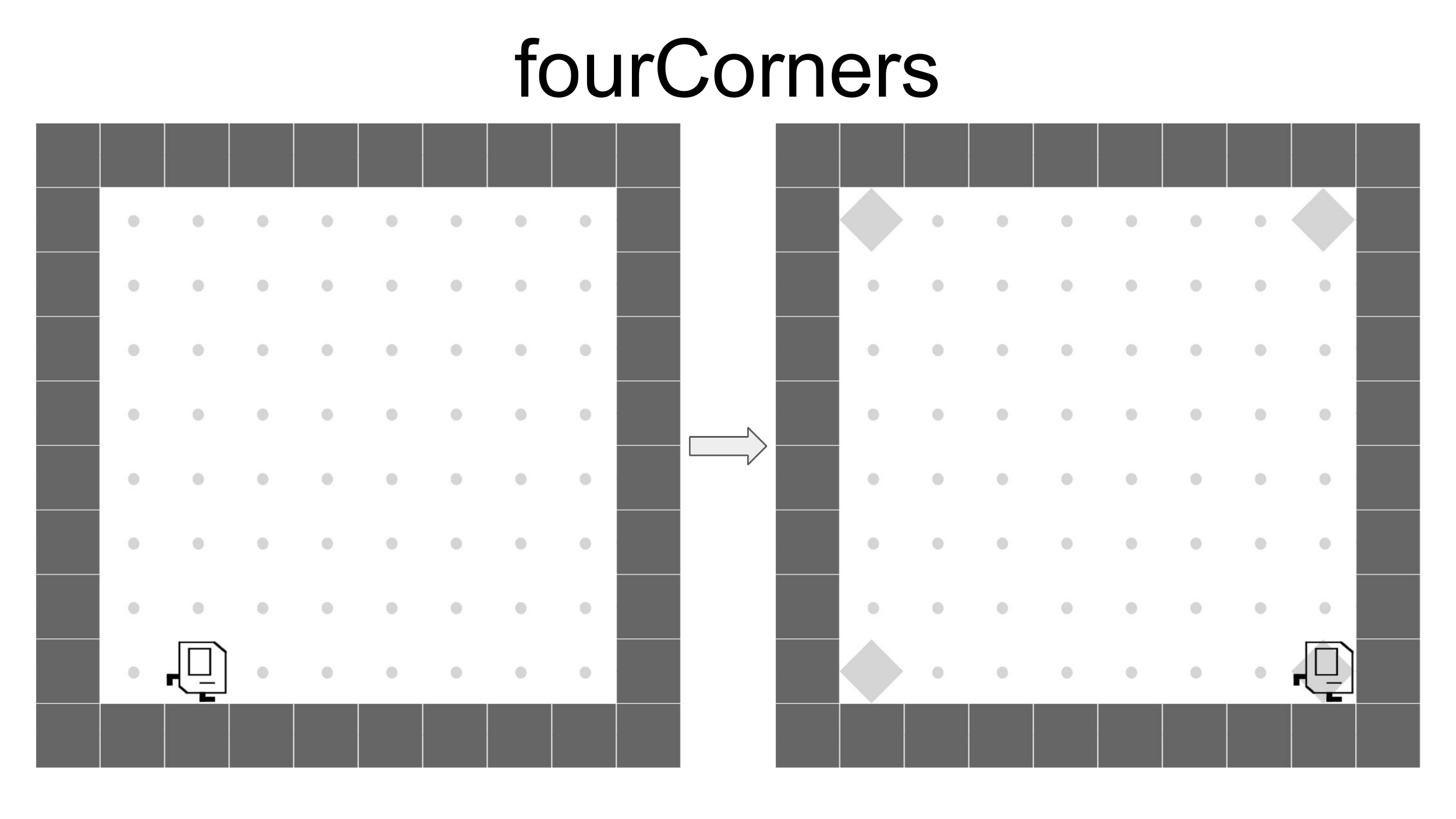}
    \caption{\textsc{fourCorner}}
    \end{subfigure}
    \hspace{0.5cm}
    \begin{subfigure}[b]{0.45\textwidth}
    \centering
    \includegraphics[trim=25 16 25 55, clip,height=0.45\textwidth]{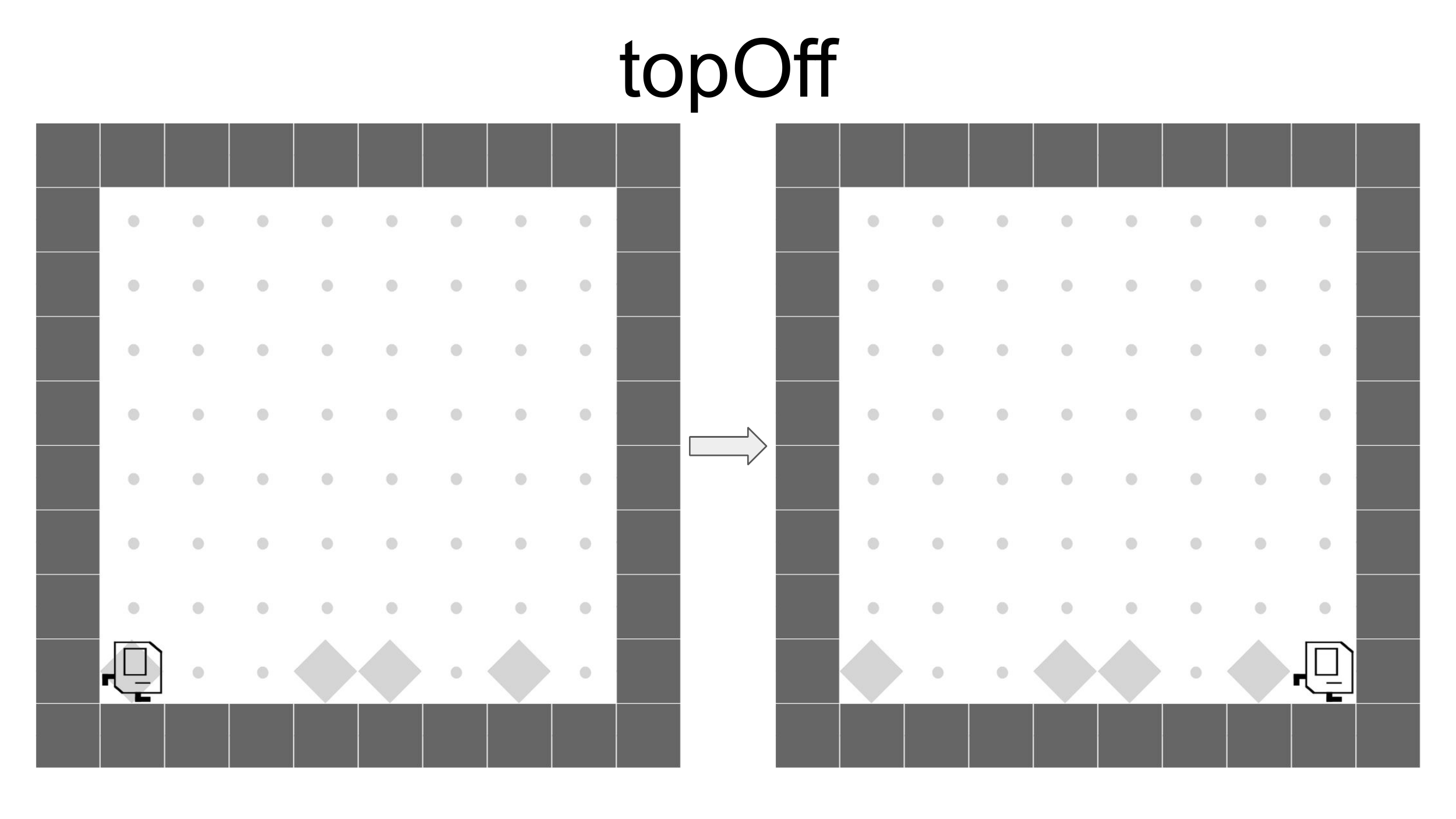}
    \caption{\textsc{TopOff}}
    \end{subfigure}
    \hspace{0.5cm}
    \begin{subfigure}[b]{0.45\textwidth}
    \centering
    \includegraphics[trim=25 16 25 50, clip,height=0.45\textwidth]{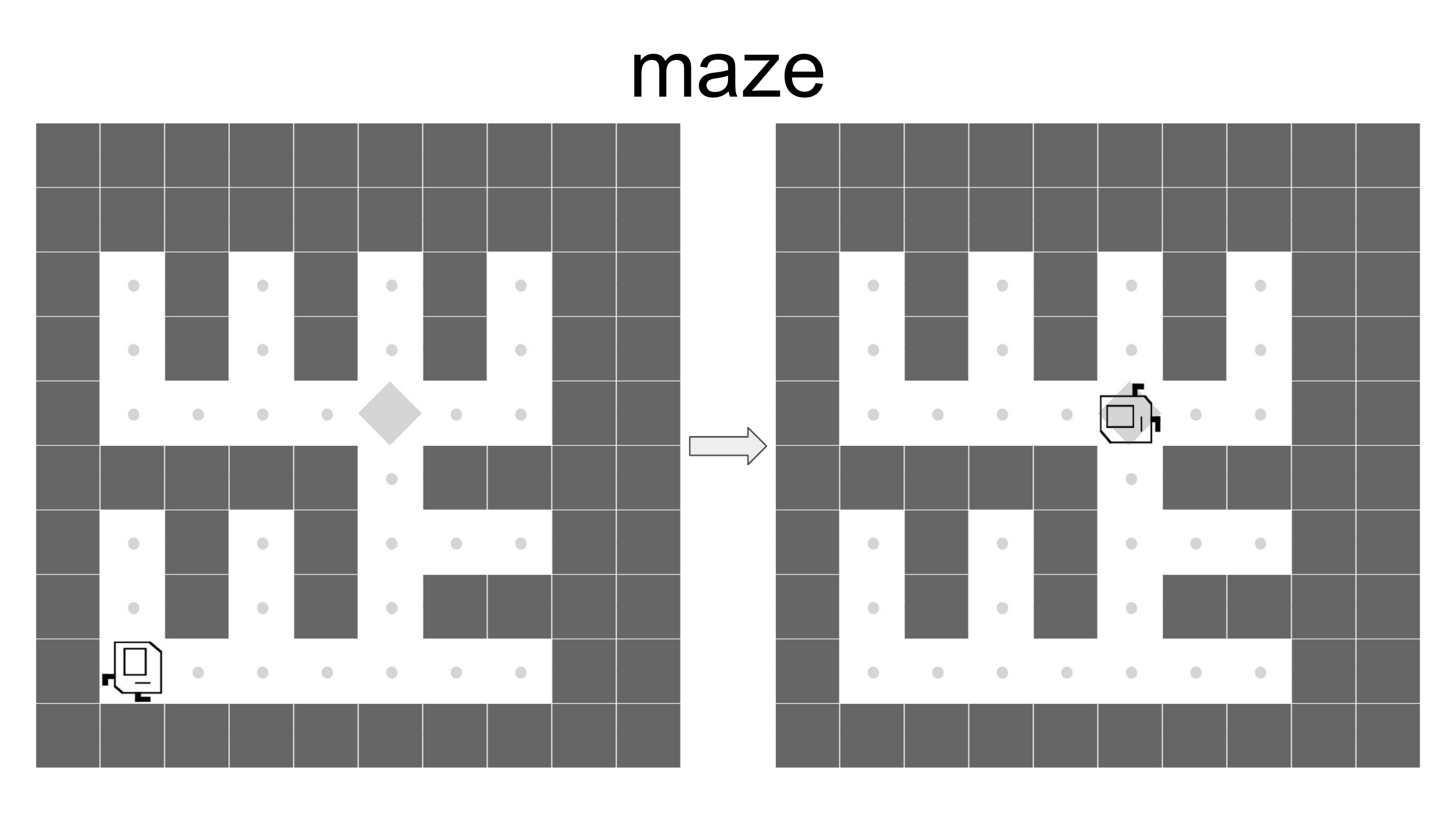}
    \caption{\textsc{Maze}}
    \end{subfigure}
    \hspace{0.5cm}
    \begin{subfigure}[b]{0.45\textwidth}
    \centering
    \includegraphics[trim=25 16 25 50, clip,height=0.45\textwidth]{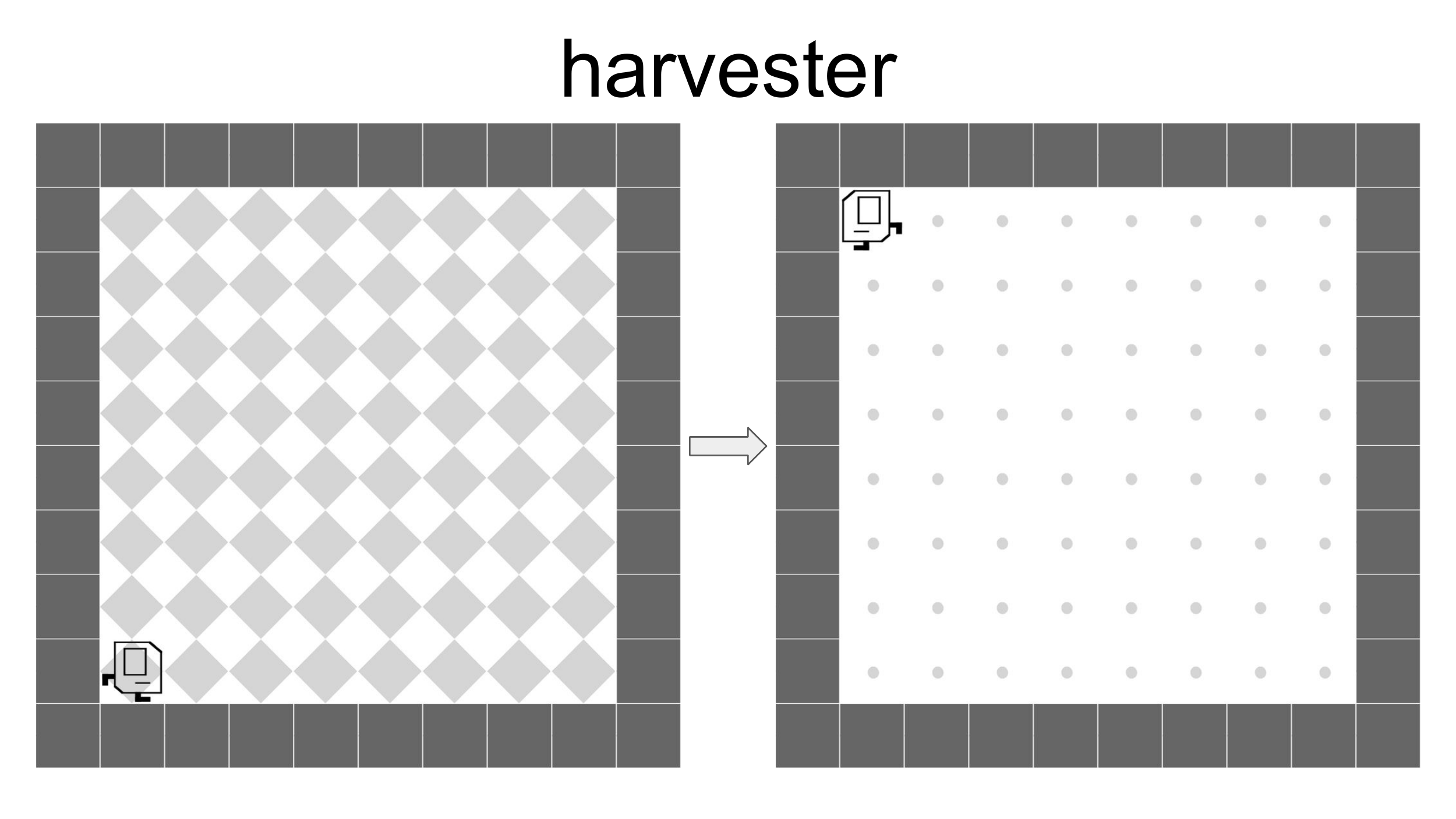}
    \caption{\textsc{Harvester}}
    \end{subfigure}
    \hspace{0.5cm}
    \begin{subfigure}[b]{0.9\textwidth}
    \centering
    \includegraphics[trim=25 0 25 30, clip,width=\textwidth]{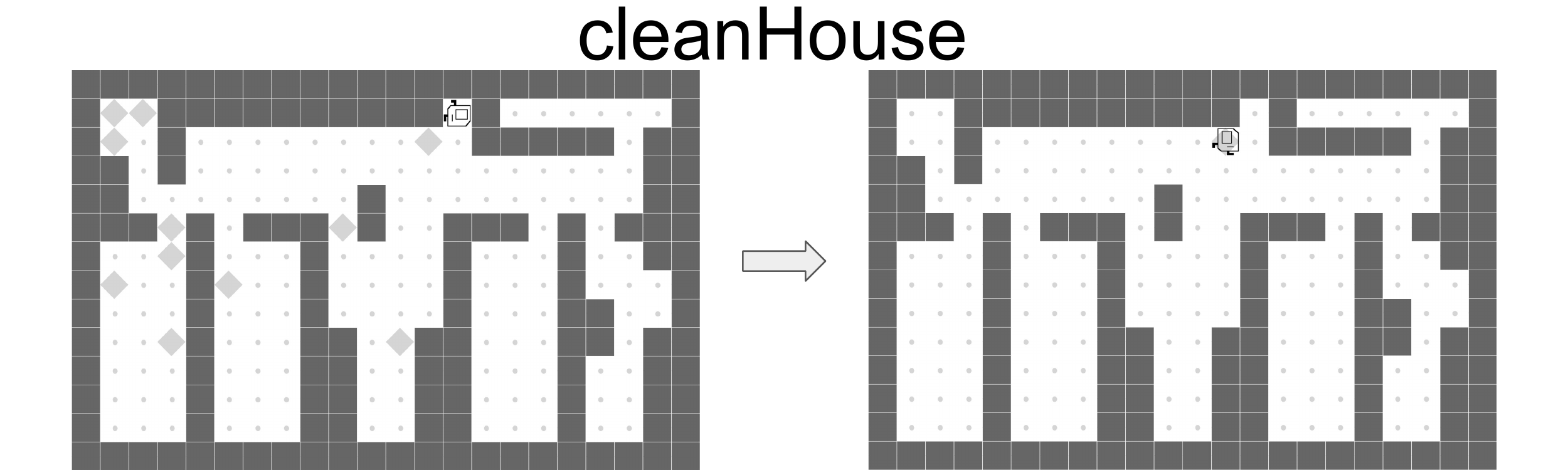}
    \caption{\textsc{CleanHouse}}
    \end{subfigure}
    \hspace{0.5cm}
    \caption[Karel Task Start/End State Depictions]{Example of initial configurations and their ideal end states of the Karel tasks. 
    Note that we show only one example of initial configuration and its ideal end sate pair for each task. 
    However, markers, walls and agent's position are randomized 
    in initial configurations depending upon task. 
    Please see section~\ref{sec:env details} for more details.
    }
    \label{fig:karel both envs}
\end{figure*}

\textbf{MDP Tasks} We utilize environment state based reward functions for the RL tasks \textsc{StairClimber}, \textsc{FourCorner}, \textsc{TopOff}, \textsc{Maze}, \textsc{Harvester}, and \textsc{CleanHouse}. For each task, we average performance of the policies on 10 random environment start configurations. For all tasks
with marker placing objectives, the final reward will be 0---regardless of the any other agent actions---if a marker is placed in the wrong location. This is done in order to discourage ``spamming'' marker placement on every grid location to exploit the reward functions. All rewards described below are then normalized
so that the return is between [0, 1.0] for tasks without penalties, and [-1.0, 1.0] for tasks with negative penalties, for easier learning for the DRL methods.
We visualize all tasks as well as their 
start and ideal end states in \myfig{fig:karel both envs} 
on a $10 \times 10$ grid for consistency in the visualizations (except \textsc{CleanHouse}).

\subsection{\textsc{StairClimber}}

The goal is to climb the stairs to reach where the marker is located. The reward is defined as a sparse reward: 1 if the agent reaches the goal in the environment rollout, -1 if the agent moves to a position off of the stairs during the rollout, and 0 otherwise. This is on a $12 \times 12$ grid, and the marker location and agent's initial location are randomized between rollouts.

\subsection{\textsc{FourCorner}}

The goal is to place a marker at each corner of the Karel environment grid. The reward is defined as sum of corners having a marker divided by four. If the Karel state has a marker placed in wrong location, the reward will be 0. This is on a $12 \times 12$ grid.

\subsection{\textsc{TopOff}}

The goal is to place a marker wherever there is already a marker in the last row of the environment, and end up in the rightmost square on the bottom row at the end of the rollout. The reward is defined as the number of consecutive places until the agent either forgets to place a marker where the marker is already present or places a marker at an empty location in last row, with a bonus for ending up on the last square. This is on a $12 \times 12$ grid, and the marker locations in the last row are randomized between rollouts.

\subsection{\textsc{Maze}}

The goal is to find a marker in randomly generated maze. The reward is defined as a sparse reward: 1 if the agent finds the marker in the environment rollout, 0 otherwise. This is on a $8 \times 8$ grid, and the marker location, agent's initial location, and
the maze configuration itself are randomized between rollouts.

\subsection{\textsc{CleanHouse}}

We design a complex $14 \times $22 Karel environment grid that resembles an apartment. The goal is to pick up the garbage (markers) placed at 10 different locations and reach the location where there is a dustbin (2 markers in 1 location). To make the task simpler, we place the markers adjacent to any wall in the environment. The reward is defined as total locations cleaned (markers picked) out of the total number of markers placed in initial Karel environment state (10). The agent's initial location is fixed but the marker locations are randomized between rollouts.

\subsection{\textsc{Harvester}}

The goal is to pickup a marker from each location in the Karel environment. The final reward is defined as the number of markers picked up divided the total markers present in the initial Karel environment state. This is on a $8 \times 8$ grid. We run both \textsc{Maze} and \textsc{Harvester} on smaller Karel environment grids to save time and compute resources because these are long horizon tasks.

\section{Hyperparameters and Training Details}
\label{sec:hyperparameters}
\subsection{DRL and DRL-abs}
RL training directly on the Karel environment is performed 
with the PPO algorithm \citep{schulman2017proximal} for 2M timesteps
using the ALF codebase\footnote{https://github.com/HorizonRobotics/alf/}.
We tried a discretized SAC \citep{haarnoja18b} implementation (by replacing Gaussian distributions with Categorical distributions), 
but it was outperformed by PPO on the Karel tasks on all environments.
We also tried tabular Q-learning from raw Karel grids (it wouldn't work well on abstract states as the state is partially observed), however it was also consistently outperformed by PPO.
For DRL, the 
policies and value networks are the same with a shared convolutional 
encoder that first processes the state (as the Karel state size is
$(H \times W \times 16)$ for $16$ possible agent direction or marker placement 
values that each state in the grid can take on at a time. The convolutional
encoder consists of two layers: the first with $32$ filters, kernel size $2$, and stride $1$,
the second with $32$ filters, kernel size $4$, and stride $1$.
For DRL-abs, the policy and value networks are both comprised of
an LSTM layer and a 2-layer fully connected network, all with hidden sizes of $100$.

For each task, we perform a comprehensive
hyperparameter grid search over the following parameters,
and report results from the run with the best averaged final reward over 5 seeds.

The hyperparameter grid is listed below, shared parameters are also listed:
\begin{itemize}
    \item Importance Ratio Clipping: \{0.05, 0.1, 0.2\}
    \item Advantage Normalization: \{True, False\}
    \item Entropy Regularization: \{0.1, 0.01, 0.001\}
    \item Number of updates per training iteration (This controls the ratio
    of gradient steps to environment steps): \{1, 4, 8, 16\}
    \item Number of environment steps per set of training iterations: 32
    \item Number of parallel actors: 10
    \item Optimizer: Adam
    \item Learning Rate: 0.001
    \item Batch Size: 128
\end{itemize}

Hyperparameters that performed best for each task are listed below.
\begin{center}
    
\scalebox{0.8}{\begin{tabular}{ccccc}
    \toprule
    DRL & Import Ratio Clip & Adv Norm & Entropy Reg & Updates per Train Iter \\
    \midrule
    \textsc{CleanHouse} & 0.1 & True & 0.01 & 4\\
    \textsc{FourCorner} & 0.2 & True & 0.01 & 16\\
    \textsc{Harvester} & 0.05 & True & 0.01 & 8  \\
    \textsc{Maze:} & 0.05 & True & 0.001 &  8\\
    \textsc{StairClimber} & 0.1 & True & 0.1 & 4\\
    \textsc{TopOff} & 0.05 & True & 0.001 & 4\\
    \bottomrule
\end{tabular}}

\scalebox{0.8}{\begin{tabular}{ccccc}
    \toprule
    DRL-abs & Import Ratio Clip & Adv Norm & Entropy Reg & Updates per Train Iter \\
    \midrule
    \textsc{CleanHouse} & 0.2 & True & 0.01 & 8\\
    \textsc{FourCorner} & 0.05 & True & 0.01 & 4\\
    \textsc{Harvester} & 0.2 & True & 0.01 & 4  \\
    \textsc{Maze:} & 0.2 & True & 0.001 &  4\\
    \textsc{StairClimber} & 0.05 & True & 0.1 & 16\\
    \textsc{TopOff} & 0.2 & True & 0.001 & 8\\
    \bottomrule
\end{tabular}}

\end{center}
\subsection{DRL-abs-t}
DRL-abs-t is limited to DRL-abs policies as the state spaces are different
for some of the Karel tasks.
For DRL-abs-t, we use the best hyperparameter configuration for 
each Karel task to train a policy to 1M timesteps. Then, we attempt 
direct policy transfer to each other task by training for another 1M
timesteps on the new task with the same hyperparameters (excluding
transferring to the same task). Numbers
reported are from the task transfer that achieved the highest reward.
The tasks that we transfer from for each task are listed below:
\begin{center}
    
\scalebox{0.8}{\begin{tabular}{cc}
    \toprule
    DRL-abs-t & Transferred from\\
    \midrule
    \textsc{CleanHouse} & \textsc{Harvester}\\
    \textsc{FourCorner} & \textsc{TopOff}\\
    \textsc{Harvester} & \textsc{Maze}\\
    \textsc{Maze} & \textsc{StairClimber}\\
    \textsc{StairClimber} & \textsc{Harvester}\\
    \textsc{TopOff} & \textsc{Harvester}\\
    \bottomrule
\end{tabular}}

\end{center}

\subsection{HRL}

\textbf{Pretraining stage:}
We first train a VAE to reconstruct action trajectories generated from our program dataset.
For each program, we generate 10 rollouts in randomly configured Karel environments to produce the HRL dataset, giving this baseline
the same data as \method\\. These variable-length action sequences are 
encoded via an LSTM encoder into a 10-dimensional, continuous latent space 
and decoded by an LSTM decoder into the original action trajectories. 
We chose 10-dimensional so as to not make downstream RL too difficult.
We tune the KL divergence weight ($\beta$) of this network such that it's as high as possible
while being able to reconstruct the trajectories well.
Network/training details below:
\begin{itemize}
    \item $\beta$: 1.0
    \item Optimizer: Adam (All optimizers)
    \item Learning Rates: $0.0003$
    \item Hidden layer size: 128
    \item \# LSTM layers (both encoder/decoder): 2
    \item Latent embedding size: 10 
    \item Nonlinearity: ReLU
    \item Batch Size: 128
\end{itemize}

\textbf{Downstream (Hierarchical) RL}
On our Karel tasks, we use the VAE's decoder to decode latent vectors
(actions for the RL agent) into varied-length action sequences for all
Karel tasks. The decoder parameters are frozen and used for all environments.
The RL agent is retrained from scratch for each task, in the same manner as 
the standard RL baselines DRL-abs and DRL.
We use Soft-Actor Critic (SAC, \citet{haarnoja18b}) as the RL algorithm as
it is state of the art in many continuous action space environments.
SAC grid search parameters for all environments follow below:
\begin{itemize}
    \item Number of updates per training iteration: \{1, 8\}
    \item Number of environment steps per set of training iterations: 8 (multiplied by the number of steps taken by the decoder in the environment)
    \item Polyak Averaging Coefficient: \{0.95, 0.9\}
    \item Number of parallel actors: 1 
    \item Batch size: 128
    \item Replay buffer size: 1M
\end{itemize}
The best hyperparameters follow:

\begin{center}
\scalebox{0.8}{\begin{tabular}{ccc}
    \toprule
    HRL-abs & Updates per Train Iter & Polyak Coefficient \\
    \midrule
    \textsc{CleanHouse} & 1 & 0.95 \\
    \textsc{FourCorner} & 8 & 0.9 \\
    \textsc{Harvester} & 8 & 0.95 \\
    \textsc{Maze} & 1 & 0.95\\
    \textsc{StairClimber} & 1 & 0.9 \\
    \textsc{TopOff} & 1 & 0.9 \\
    \bottomrule
\end{tabular}}

\scalebox{0.8}{\begin{tabular}{ccc}
    \toprule
    HRL & Updates per Train Iter & Polyak Coefficient \\
    \midrule
    \textsc{CleanHouse} & 1 & 0.9 \\
    \textsc{FourCorner} & 1 & 0.95 \\
    \textsc{Harvester} & 1 & 0.95 \\
    \textsc{Maze} & 8 & 0.9\\
    \textsc{StairClimber} & 8 & 0.95 \\
    \textsc{TopOff} & 8 & 0.95 \\
    \bottomrule
\end{tabular}}
\end{center}

\subsection{Na\"{i}ve}
\label{sec:RL-rho appendix}
The na\"{i}ve program synthesis baseline
takes an initial token as input and outputs an entire program at each timestep to learn a recurrent policy guided by the rewards of these programs. We execute these generated programs on 10 random environment start configurations in Karel to get the reward. We run PPO for 2M Karel environment timesteps. The policy network is comprised of one shared GRU layer, followed by two fully connected layers, for both the policy and value networks. For evaluation, we generate 64 programs from the learned policy, and choose the program with the maximum reward on 10 demonstrations.
For each task, we perform a hyperparameter grid search over the following parameters, and report results from the run with the best averaged final reward over 5 seeds. We exponentially decay the entropy loss coefficient in PPO from the initial to final entropy coefficient to avoid local minima during the initial training steps.
\begin{itemize}
    \item Learning Rate: $0.0005$
    \item Batch Size (B): \{64, 128, 256\}
    \item initial entropy coefficient ($E_i$): \{1.0, 0.1\}
    \item final entropy coefficient: \{0.01\}
    \item Hidden Layer Size: $64$
\end{itemize}
Hyperparameters that performed best for each task are listed below.
\begin{center}
\scalebox{0.8}{\begin{tabular}{ccc}
    \toprule
    Na\"{i}ve & B & $E_i$ \\
    \midrule
    \textsc{WHILE} & 128 & 0.1 \\
    \textsc{IFELSE+WHILE} & 256 & 1.0 \\
    \textsc{2IF+IFELSE} & 256 & 0.1 \\
    \textsc{WHILE+2IF+IFELSE} & 128 & 0.1 \\ 
    \bottomrule
\end{tabular}}

\scalebox{0.8}{\begin{tabular}{ccc}
    \toprule
    Na\"{i}ve & B & $E_i$ \\
    \midrule
    \textsc{CleanHouse} & 128 & 0.1 \\
    \textsc{FourCorner} & 128 & 1.0 \\
    \textsc{Harvester} & 128 & 1.0 \\
    \textsc{Maze} & 256 & 1.0 \\
    \textsc{StairClimber} & 128 & 1.0 \\
    \textsc{TopOff} & 128 & 1.0 \\
    \bottomrule
\end{tabular}}

\end{center}

\subsection{VIPER}
VIPER \citep{bastani2018verifiable} builds a decision tree programmatic policy by imitating a given teacher policy. 
We use the best DRL policies as teachers instead of the DQN \citep{mnih2015human} teacher policy used in \citet{bastani2018verifiable}. 
We did this in order to give the teacher the best performance possible for maximum fairness in comparison against VIPER, as we empirically found the PPO policy to perform much better on our tasks than a DQN policy.

We perform a grid search over VIPER hyperparameters, listed below:
\begin{itemize}
    \item Max depth of decision tree: \{6, 12, 15\}
    \item Max number of samples for tree policy: \{100k, 200k, 400k\}
    \item Sample reweighting: \{True, False\}
\end{itemize}

The best hyperparameters found for each task are listed below:

\begin{center}
\scalebox{0.8}{\begin{tabular}{cccc}
    \toprule
    VIPER & Max Depth & Max Num Samples & Sample Reweighting \\
    \midrule
    \textsc{CleanHouse} & 6 & 100k & False \\
    \textsc{FourCorner} & 12 & 100k & False \\
    \textsc{Harvester} & 12 & 400k & True\\
    \textsc{Maze} & 12 & 100k & True\\
    \textsc{StairClimber} & 12 & 400k & True\\
    \textsc{TopOff} & 15 & 100k & False \\
    \bottomrule
\end{tabular}}
\end{center}

\subsection{Program Embedding Space VAE Model}
\label{sec:training}
\noindent \textbf{Encoder-Decoder Architecture.}
The encoder and decoder are both recurrent networks.
The encoder structure consists of a PyTorch token embedding layer, then a recurrent GRU cell, and two linear layers that produce $\mu$ and $\log \sigma$ vectors to sample the program embedding.

The decoder consists of a recurrent GRU cell which takes in the embedding of the previous token generated and then a linear token output layer which models the log probabilities of all discrete tokens.
Since we have access to DSL grammar during program synthesis, we
utilize a syntax checker 
based on the Karel DSL grammar from
\citet{bunel2018leveraging} at the output of the decoder to limit predictions
to syntactically valid tokens. 
We restrict our decoder from predicting syntactically invalid programs by masking out tokens that make a program syntactically invalid at each timestep. 
This syntax checker is designed as a state machine that keeps track of a set of valid next tokens based on the current token, open code blocks (\eg \texttt{while, if, ifelse}) in the given partial program, and the grammar rules of our DSL. 
Since we generate a program as a sequence of tokens, the syntax checker outputs at each timestep a mask 
$M$, where $M \in \{-\infty, 0\}^{\text{number of DSL tokens}}$, and 
\begin{equation*}
    M_j = 
        \begin{cases}
        -\infty & \text{if the j-th token is not valid in the current context}\\
        0 & \text{otherwise}
        \end{cases}
\end{equation*}
This mask is added to the output of the last layer of the decoder, just before the Softmax operation that normalizes the output to a probability over the tokens.

\noindent \textbf{$\mathbf{\pi}$ Architecture.}
The program-embedding conditioned policy $\pi$ consists of a GRU layer that operates on the inputs and three MLP layers that output the log probability of environment actions. Specifically, it takes a latent program vector, current environment state, and previous action as input and outputs the predicted environment action for each timestep.

To evaluate how close the predicted neural execution traces are to the execution traces of the ground-truth programs, we consider the following metrics: 
\begin{itemize}
    \item Action token accuracy: the percentage of matching actions in the predicted execution traces and the ground-truth execution traces. 
    \item Action sequence accuracy: the percentage of matching action sequences in the predicted execution traces and the ground-truth execution traces. It requires that a predicted execution trace entirely matches the ground-truth execution trace.
\end{itemize}   

After convergence, our model achieves an action token accuracy of 96.5\% and an action sequence accuracy of 91.3\%. 

\noindent \textbf{Training.}
The reinforcement learning algorithm used for the program behavior reconstruction 
$\mathcal{L}^{\text{R}}$ is REINFORCE \citep{williams1992simple}.

When training \method\\ with all losses, we first train with the Program Reconstruction ($\mathcal{L}^{\text{P}}$) and Latent Behavior Reconstruction ($\mathcal{L}^{\text{L}}$) losses, essentially setting $\lambda_1 = \lambda_3 = 1$ and $\lambda_2 = 0$ of our full objective, reproduced below:

\begin{equation}
    \min_{\theta, \phi, \pi} 
    \lambda_1 \mathcal{L}_{\theta, \phi}^{\text{P}}(\rho) + 
    \lambda_2 \mathcal{L}_{\theta, \phi}^{\text{R}}(\rho) + 
    \lambda_3 \mathcal{L}_\pi^{\text{L}}(\rho, \pi),
\end{equation}

Once this model is trained for one epoch, we then train exclusively with the Program
Behavior Reconstruction loss ($\mathcal{L}^{\text{R}}$), setting
$\lambda_2 = 1$ and $\lambda_1 = \lambda_3 = 0$, with equal number of updates. These two update steps are repeated alternatively till convergence is achieved. This is done to avoid
potential issues of updating with supervised and reinforcement learning gradients at the same time. We did not attempt to train these 3 losses jointly.

All other shared hyperparameters and training details are listed below:
\begin{itemize}
    \item $\beta$: 0.1
    \item Optimizer: Adam (All optimizers)
    \item Supervised Learning Rate: $0.001$
    \item RL Learning Rate: $0.0005$
    \item Batch Size: $256$
    \item Hidden Layer Size: $256$
    \item Latent Embedding Size: $256$
    \item Nonlinearity: $Tanh()$
\end{itemize}

\subsection{Cross-Entropy Method (CEM)}
CEM search works as follows: we sample an initial latent program vector from the initial distribution $D_I$, and generate population of latent program vectors from a $\mathcal{N}(0, \sigma I_d)$ distribution, where $I_d$ is the identity matrix of dimension $d$. The samples are added to the initial latent program vector to obtain the population of latent program vectors which are decoded into programs to obtain their rewards. The population is then sorted based on rewards obtained, and a set of `elites' with the highest reward are reduced using weighted mean to one latent program vector for the next iteration of sampling. This process repeats for all CEM iterations.

We include the following sets of hyperparameters when searching over the program embedding space to maximize $R_{\text{mat}}$ to reproduce ground-truth program behavior or to maximize $R_{\text{mat}}$ in the Karel task MDP.

\begin{itemize}
    \item Population Size ($S$): \{8, 16, 32, 64\}
    \item $\mu$: \{$0.0$\}
    \item $\sigma$: \{$0.1, 0.25, 0.5$\}
    \item $\%$ of population elites (this refers to the percent of the population considered `elites'): \{$0.05, 0.1, 0.2$\}
    \item Exponential $\sigma$ decay\footnote{Over the first 500 epochs, we exponentially decay $\sigma$ to $0.1$, and then we keep it at $0.1$ for the rest of the epochs if True.}: \{True, False\}
    \item Initial distribution $D_I$: $\{\mathcal{N}(1,\mathbf{0}), \mathcal{N}(0,I_d), \mathcal{N}(0,0.1I_d) \}$
\end{itemize}

Since a comprehensive grid search over the hyperparameter space 
would be too computationally expensive, we choose parameters
heuristically. 
We report results from the run with the best averaged reward over 5 seeds. 
Hyperparameters that performed best for each task are listed below.

\textbf{Ground-Truth Program Reconstruction}
We include the following sets of hyperparameters when searching over the program
embedding space to maximize $R_{\text{mat}}$ to reproduce ground-truth program behavior.
We allow the search to run for $1000$ CEM iterations, counting the search as a success when it achieves
10 consecutive CEM iterations with matching the ground-truth program behaviors exactly in the environment
across 10 random environment start configurations. We use same hyperparameter set to compare \method\\-P, \method\\-P+R, \method\\-P+L, and \method\\.

\begin{center}
\scalebox{0.8}{\begin{tabular}{cccccc}
    \toprule
    CEM & $S$ & $\sigma$ & \# Elites & Exp Decay & $D_I$\\
    \midrule
    \textsc{WHILE} & 32 & 0.25 & 0.1 & False & $\mathcal{N}(0,0.1I_d)$ \\
    \textsc{IFELSE+WHILE} & 32 & 0.25 & 0.1 & True & $\mathcal{N}(0,0.1I_d)$\\
    \textsc{2IF+IFELSE} & 16 & 0.25 & 0.2 & True & $\mathcal{N}(0,0.1I_d)$\\
    \textsc{WHILE+2IF+IFELSE} & 32 & 0.25 & 0.2 & False & $\mathcal{N}(0,0.1I_d)$\\
    \bottomrule
\end{tabular}}
\end{center}
\textbf{MDP Task Performance}
We include the following sets of hyperparameters when searching over the \method\\ program
embedding space to maximize rewards in the MDP.
We allow the search to run for $1000$ CEM iterations, counting the search as a success when it achieves
10 consecutive CEM iterations of maximizing environment reward (solving the task) across 10 random environment
start configurations.

\begin{center}
\scalebox{0.8}{\begin{tabular}{cccccc}
    \toprule
    CEM & $S$ & $\sigma$ & \# Elites & Exp Decay & $D_I$\\
    \midrule
    \textsc{CleanHouse} & 32 & 0.25 & 0.05 & True & $\mathcal{N}(1,\bf{0})$\\
    \textsc{FourCorner} & 64 & 0.5 & 0.2 & False & $\mathcal{N}(0,0.1I_d)$\\
    \textsc{Harvester} & 32 & 0.5 & 0.1 & True & $\mathcal{N}(0,I_d)$\\
    \textsc{Maze} & 16 & 0.1 & 0.1 & False & $\mathcal{N}(1,\bf{0})$\\
    \textsc{StairClimber} & 32 & 0.25 & 0.05 & True & $\mathcal{N}(0,0.1I_d)$\\
    \textsc{TopOff} & 64 & 0.25 & 0.05 & False & $\mathcal{N}(0,0.1I_d)$\\
    \bottomrule
\end{tabular}}
\end{center}

\subsection{Random Search \method\\ Ablation}
The random search \method\\ ablations (\method\\-rand-8 and \method\\-rand-64) replace
the CEM search method for latent program synthesis with a simple random search method.
Both use the full \method\\ model trained with all learning objectives. 
We sample an initial vector from an initial distribution 
$D_I$ and add it to either 8 or 64 latent vector samples from a $\mathcal{N}(0, \sigma I_d)$ distribution. We then decode those vectors
into programs and evaluate their rewards, and then report the rewards of the best-performing latent program
from that population.

As such, the only parameters that we require are the initial sampling distribution and $\sigma$.
We perform a grid search over the following for both \method\\-rand-8 and \method\\-rand-64.

\begin{itemize}
    \item $\sigma$: \{0.1, 0.25, 0.5\}
    \item Initial distribution $D_I$: $\{\mathcal{N}(0,I_d), \mathcal{N}(0,0.1I_d) \}$
\end{itemize}

\textbf{Ground-Truth Program Reconstruction}
We report hyperparameters below for both random search methods on program reconstruction tasks.
\begin{center}
\scalebox{0.8}{\begin{tabular}{cccccc}
    \toprule
    \method\\-rand-8 & $\sigma$ & $D_I$\\
    \midrule
    \textsc{WHILE} & 0.1 & $\mathcal{N}(0,0.1I_d)$ \\
    \textsc{IFELSE+WHILE} & 0.5 & $\mathcal{N}(0,0.1I_d)$\\
    \textsc{2IF+IFELSE} & 0.5 & $\mathcal{N}(0,0.1I_d)$\\
    \textsc{WHILE+2IF+IFELSE} & 0.5 & $\mathcal{N}(0,0.1I_d)$\\
    
    \bottomrule
\end{tabular}}
\end{center}

\begin{center}
\scalebox{0.8}{\begin{tabular}{cccccc}
    \toprule
    \method\\-rand-64 & $\sigma$ & $D_I$\\
    \midrule
    \textsc{WHILE} & 0.5 & $\mathcal{N}(0,0.1I_d)$ \\
    \textsc{IFELSE+WHILE} & 0.5 & $\mathcal{N}(0,0.1I_d)$\\
    \textsc{2IF+IFELSE} & 0.5 & $\mathcal{N}(0,0.1I_d)$\\
    \textsc{WHILE+2IF+IFELSE} & 0.5 & $\mathcal{N}(0,0.1I_d)$\\
    \bottomrule
\end{tabular}}
\end{center}

\textbf{MDP Task Performance}
We report hyperparameters below for both random search methods on Karel tasks.
\begin{center}
\scalebox{0.8}{\begin{tabular}{cccccc}
    \toprule
    \method\\-rand-8 & $\sigma$ & $D_I$\\
    \midrule
    \textsc{CleanHouse} & 0.5 & $\mathcal{N}(0,0.1I_d)$\\
    \textsc{FourCorner} & 0.5 & $\mathcal{N}(0,0.1I_d)$\\
    \textsc{Harvester} & 0.5 & $\mathcal{N}(0,0.1I_d)$\\
    \textsc{Maze} & 0.25 & $\mathcal{N}(0,0.1I_d)$\\
    \textsc{StairClimber} & 0.5 & $\mathcal{N}(0,I_d)$\\
    \textsc{TopOff} & 0.25 & $\mathcal{N}(0,0.1I_d)$\\
    \bottomrule
\end{tabular}}
\end{center}

\begin{center}
\scalebox{0.8}{\begin{tabular}{cccccc}
    \toprule
    \method\\-rand-64 & $\sigma$ & $D_I$\\
    \midrule
    \textsc{CleanHouse} & 0.5 & $\mathcal{N}(0,0.1I_d)$\\
    \textsc{FourCorner} & 0.25 & $\mathcal{N}(0,0.1I_d)$\\
    \textsc{Harvester} & 0.5 & $\mathcal{N}(0,0.1I_d)$\\
    \textsc{Maze} & 0.1 & $\mathcal{N}(0, 0.1I_d)$\\
    \textsc{StairClimber} & 0.25 & $\mathcal{N}(0,0.1I_d)$\\
    \textsc{TopOff} & 0.5 & $\mathcal{N}(0,0.1I_d)$\\
    \bottomrule
\end{tabular}}
\end{center}

\section{Computational Resources}
\label{sec:computation}
For our experiments, we used both internal and cloud provider machines. Our internal machines
are:
\begin{itemize}
    \item M1: 40-vCPU Intel Xeon with 4 GTX Titan Xp GPUs
    \item M2: 72-vCPU Intel Xeon with 4 RTX 2080 Ti GPUs
\end{itemize}

The cloud instances that we used are either 
128-thread AMD Epyc or 96-thread Intel Xeon based cloud instances with 4-8 NVIDIA Tesla T4 GPUs. 
Experiments were run in parallel across many CPUs whenever possible, 
thus requiring the high vCPU count machines.

The experiment costs (GPU memory/time) are as follows:

Learning Program Embedding Stage:
\begin{itemize}
    \item LEAPS-P: 4.2GB/13hrs on either M1 or M2
    \item LEAPS-P+R: 4.2GB/44-54hrs on M2
    \item LEAPS-P+L: 8.7GB/26hrs on either M1 or M2
    \item LEAPS: 8.8GB/104hrs on M1, 8.8GB/58hrs on M2
\end{itemize}

Policy Learning Stage:
\begin{itemize}
    \item CEM search: 0.8GB/4-10min (depends on the CEM population size and the number of iterations until convergence)
    \item DRL/DRL-abs/DRL-abs-t: 0.7-2GB/1hr per run with parallelization across 10 processes
    \item HRL/HRL-abs: 1-2GB/2.5hrs per run
    \item VIPER: 0.7GB/20-30 minutes (excluding the time for learning its teacher policy)
\end{itemize}
\begin{figure*}
\centering
\begin{subfigure}[b]{0.94\textwidth}
\centering
\includegraphics[trim=5 5 5 5, clip,width=\textwidth]{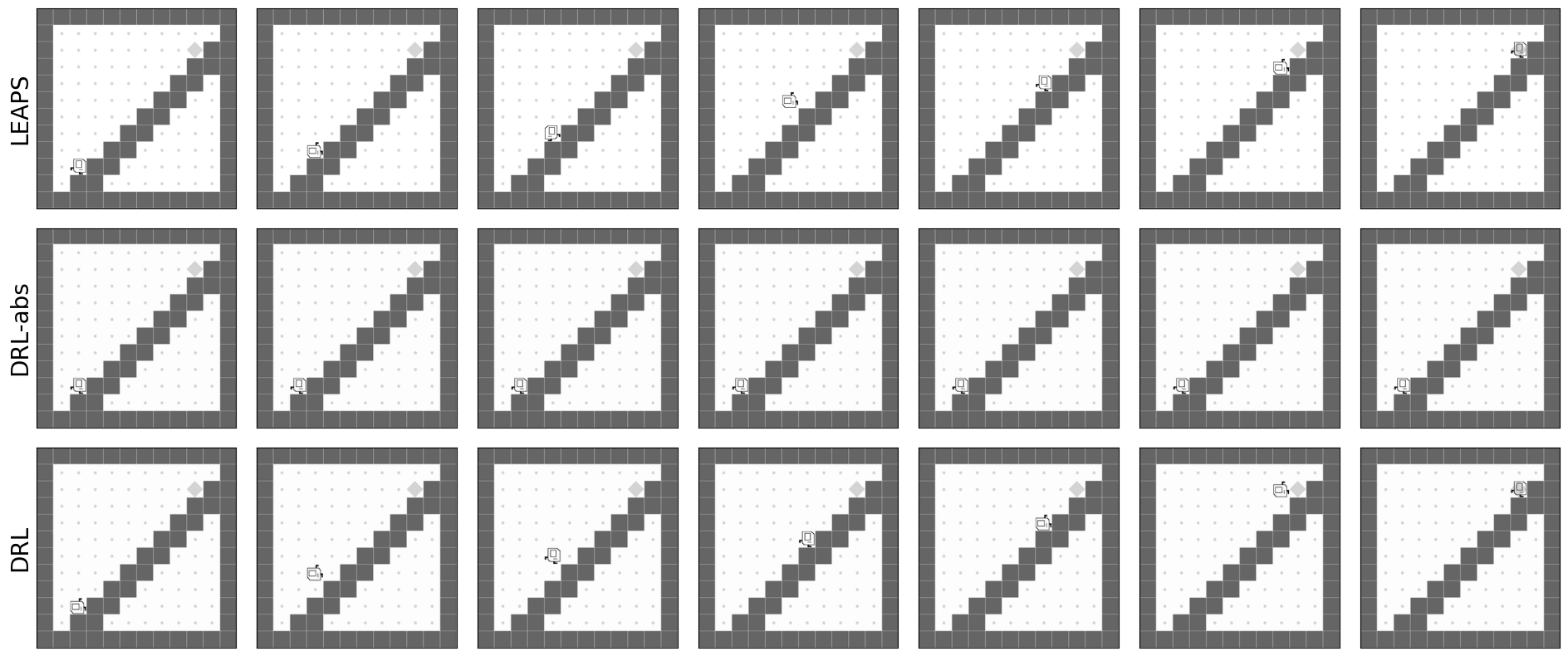}
\caption{\textsc{StairClimber}: \method\\ and DRL are able to climb the stairs,
DRL-abs is unable to do so.}
\end{subfigure}
\begin{subfigure}[b]{0.94\textwidth}
\includegraphics[trim=5 5 5 5, clip,width=\textwidth]{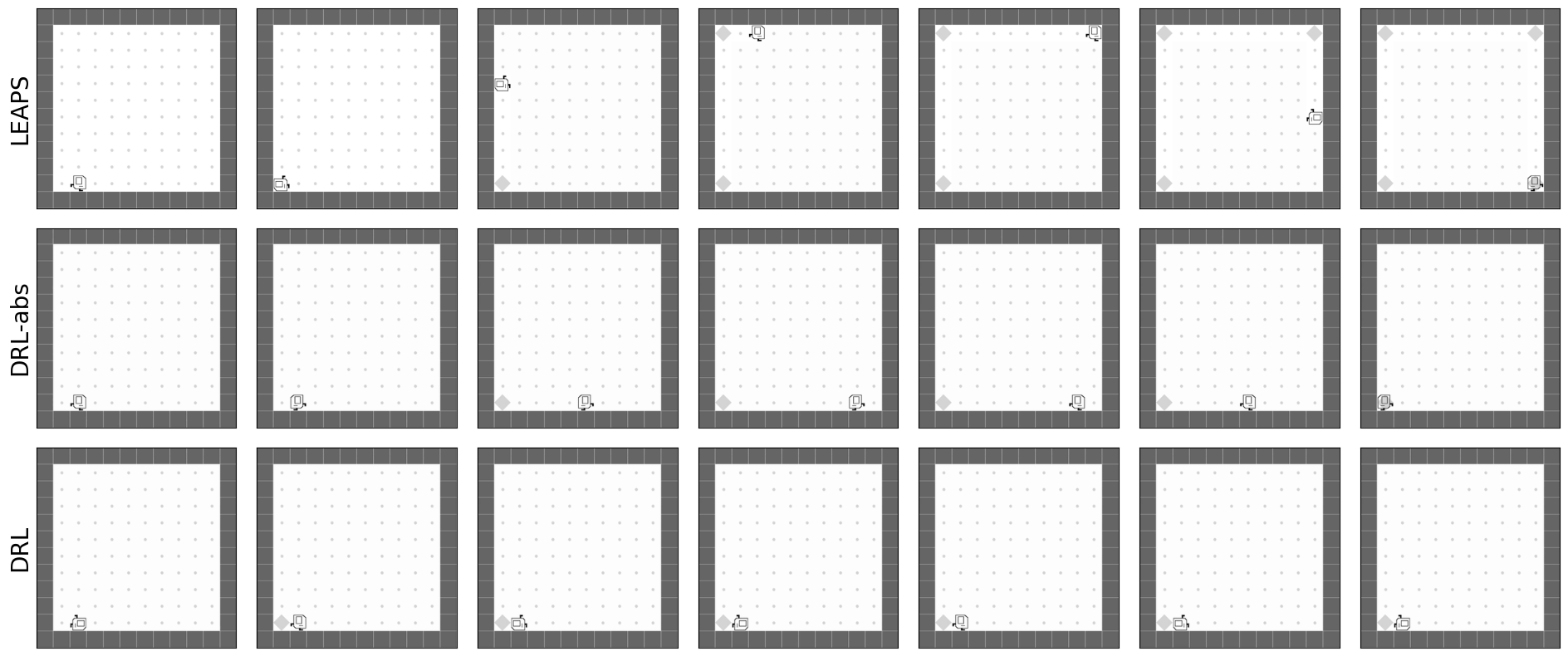}
\caption{\textsc{fourCorner}: In this example, \method\\ generates
a program which is able to completely solve the task. Both DRL methods learn 
to only place one single marker in the left bottom corner.}
\end{subfigure}
\begin{subfigure}[b]{0.94\textwidth}
\includegraphics[trim=5 5 5 5, clip,width=\textwidth]{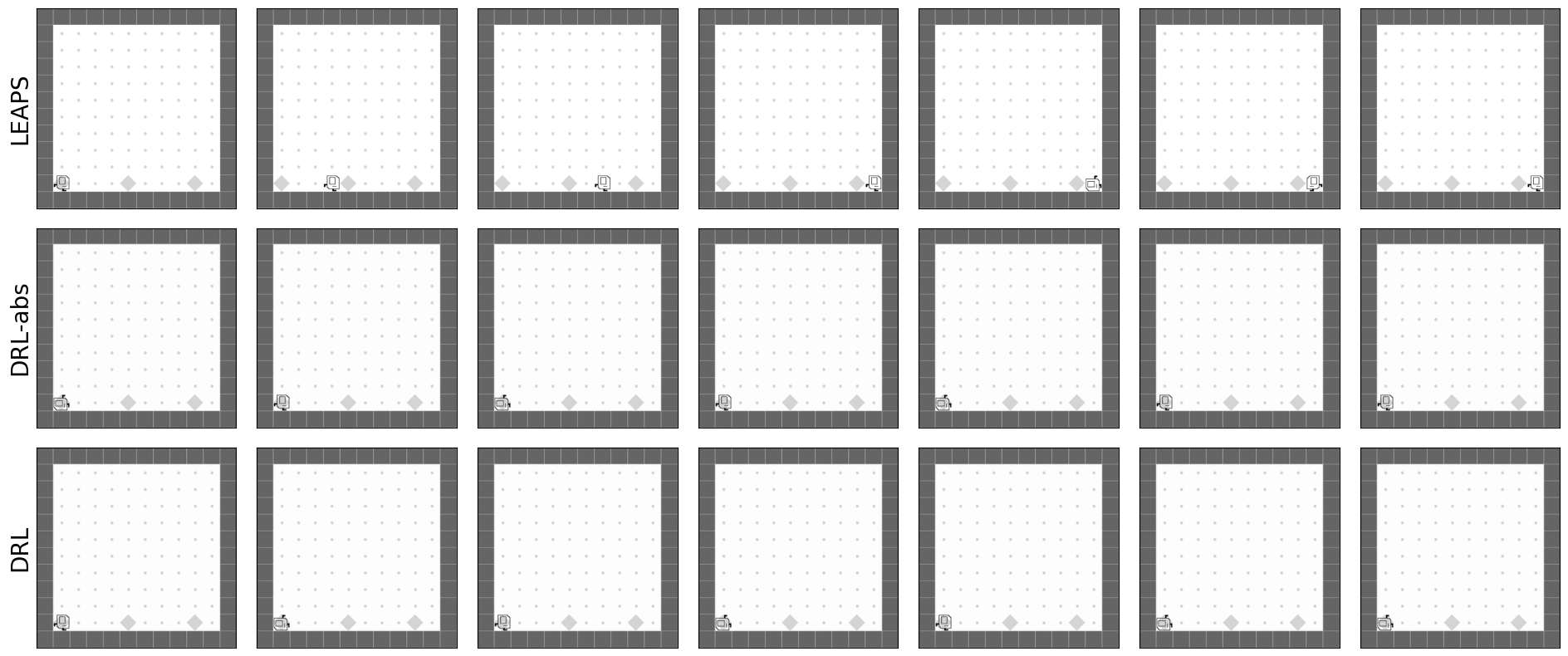}
\caption{\textsc{TopOff}: Here, \method\\ generates a program that 
solves the task by ``topping off'' each marker. Both DRL methods
only learn to top off the initial marker.}
\end{subfigure}
\end{figure*}

\begin{figure*}
\centering
\ContinuedFloat
\begin{subfigure}[b]{0.94\textwidth}
\includegraphics[trim=5 5 5 5, clip,width=\textwidth]{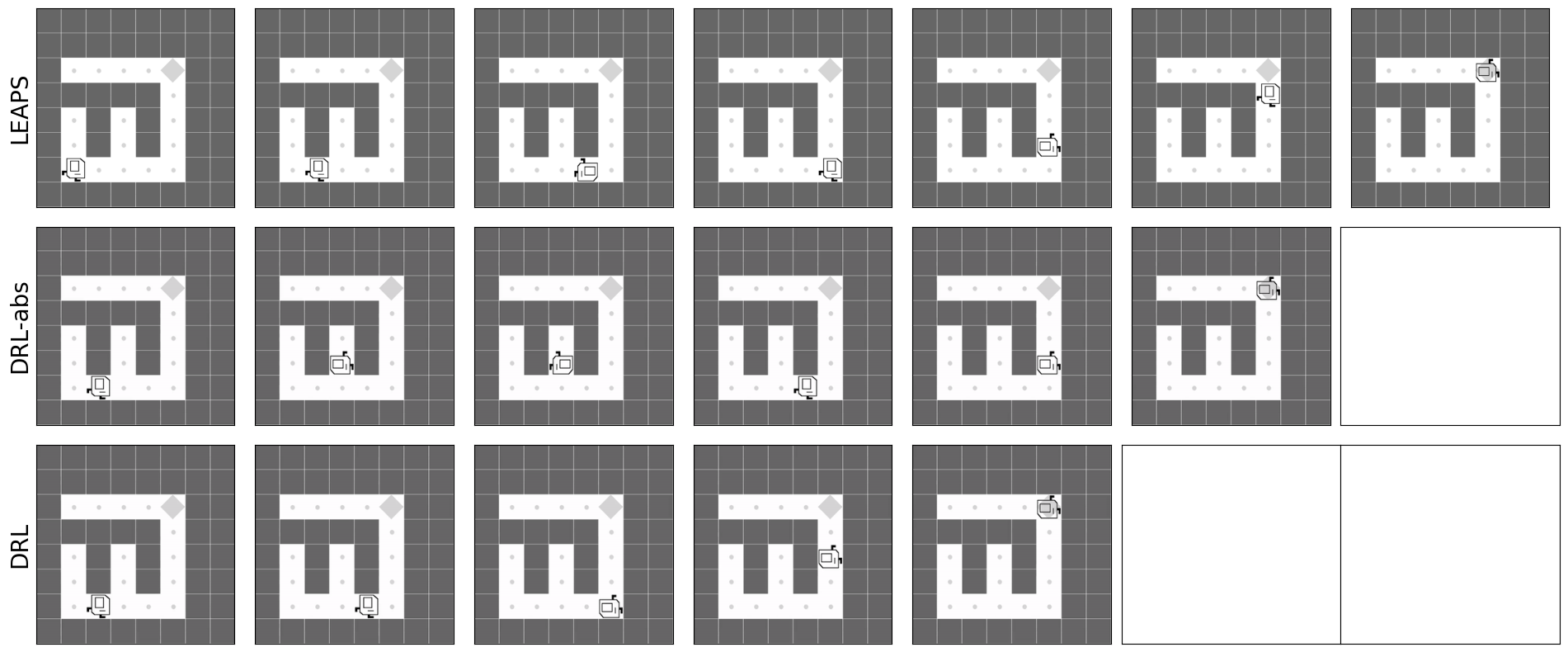}
\caption{\textsc{Maze}: All three methods are able to solve the task.}
\end{subfigure}
\begin{subfigure}[b]{0.94\textwidth}
\includegraphics[trim=5 5 5 5, clip,width=\textwidth]{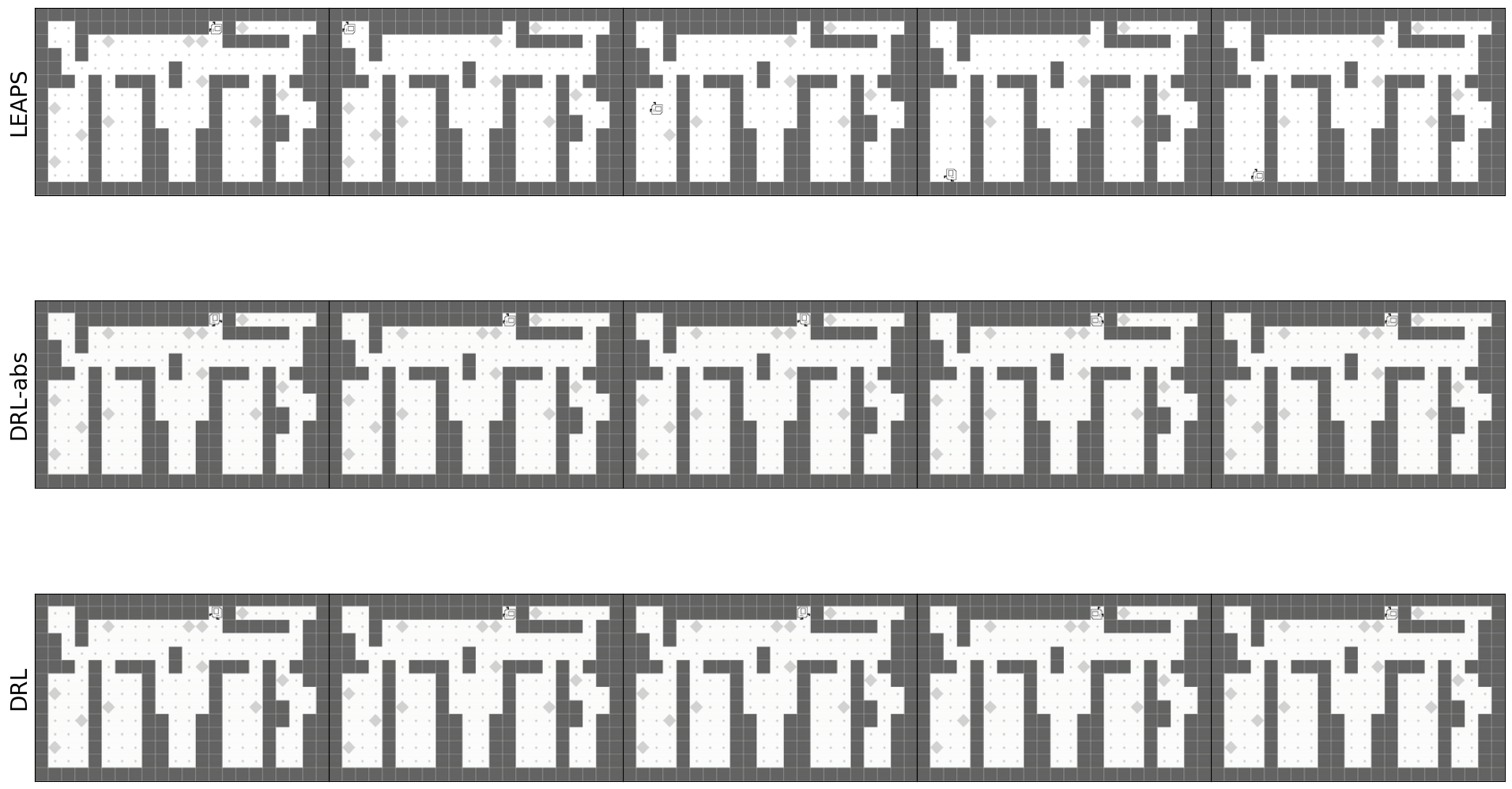}
\caption{\textsc{CleanHouse}: While both DRL methods
learn no meaningful behaviors (generally just spinning around
in place), \method\\ generates a program that is able to navigate
to and clean the leftmost room.}
\end{subfigure}
\begin{subfigure}[b]{0.94\textwidth}
\includegraphics[trim=5 5 5 5, clip,width=\textwidth]{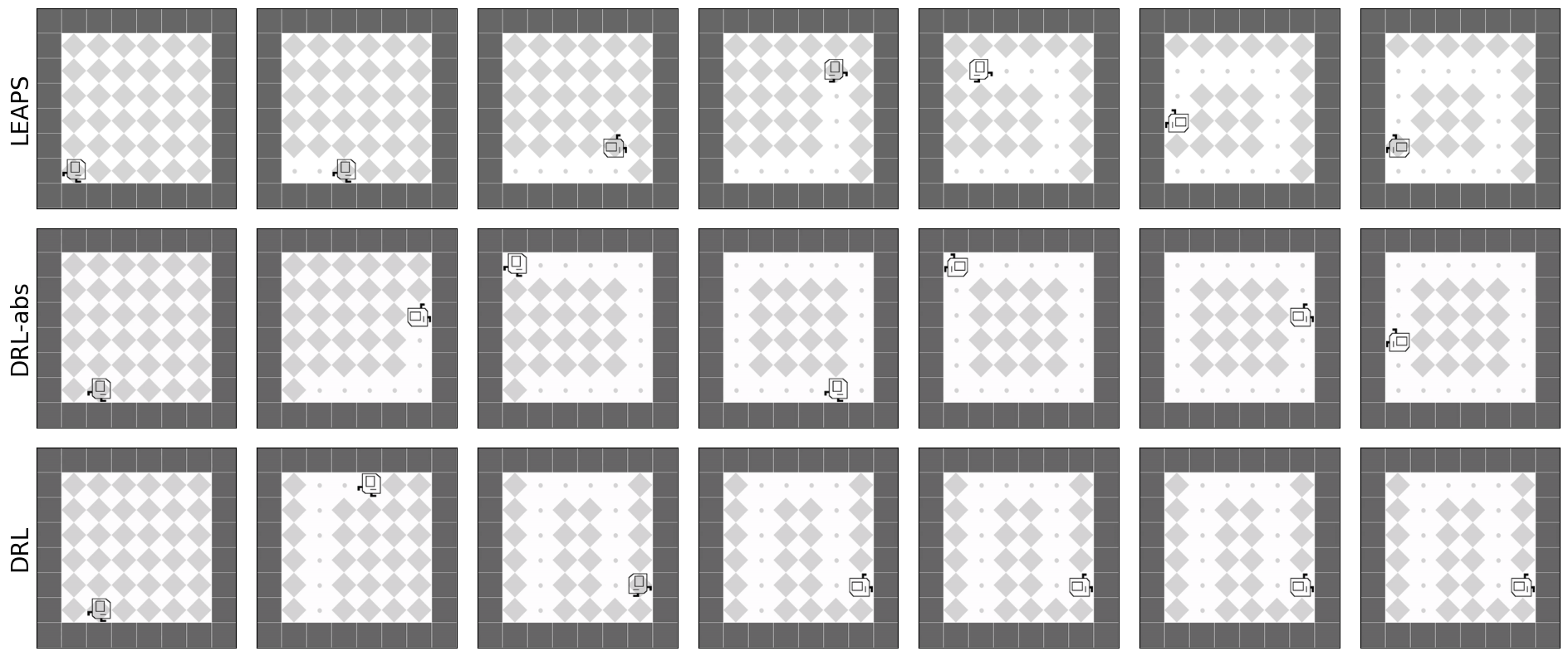}
\caption{\textsc{Harvester}: All three methods make partial progress
on \textsc{Harvester}.}
\end{subfigure}
\caption[Karel Rollout Visualizations]{\textbf{Karel Rollout Visualizations.} Example rollouts for \method\\,
DRL-abs, and DRL for each task.}
\label{fig:example rollouts}
\end{figure*}

\end{document}